%% file: paper.tex
\documentclass[]{fairmeta}
\usepackage[table]{xcolor}
\usepackage{pifont}
\usepackage{colortbl}
\usepackage{soul}   
\usepackage{enumitem}

\usepackage{amsmath,amssymb}
\usepackage[table,svgnames]{xcolor}
\usepackage{algorithm}
\usepackage{algpseudocode}
\usepackage{framed}
\usepackage{wrapfig}
\usepackage{bbm}

\definecolor{darkgreen}{rgb}{0,0.5,0}
\definecolor{light-gray}{gray}{0.9}
\definecolor{comm}{gray}{0.5}
\definecolor{denim}{rgb}{0.08, 0.38, 0.74}
\definecolor{mint}{RGB}{220,245,230}
\definecolor{pistachio}{RGB}{236,246,228} 
\hypersetup{
  colorlinks   = true,
  urlcolor     = denim,
  linkcolor    = denim,
  citecolor    = denim,
}

\usepackage{booktabs,multirow,makecell}
\newcommand{\bench}[2]{\makecell[l]{#1\\#2}}
\newcommand{\benchcenter}[2]{\makecell[c]{#1\\#2}}

\newcommand{\methodrm}[0]{\texttt{J1}}

\usepackage[most]{tcolorbox}
\usepackage{listings}

\usepackage{color-edits}
\addauthor[1]{gn}{magenta}

\newtcolorbox{prompt}[1]{
    enhanced,
    left=4mm,
    right=4mm,
    top=2mm,
    bottom=2mm,
    boxsep=0mm,
    rounded corners,
    title=#1,
    fontupper=\footnotesize\linespread{0.9}\fontfamily{lmr}\selectfont,
}

\newtcblisting{promptbox}[1][]{
  listing only,
  breakable,
  colback=gray!5,
  colframe=black!70,
  coltitle=white,
  fonttitle=\bfseries,
  title={#1},  
  left=3mm,
  right=3mm,
  top=3mm,
  bottom=3mm,
  listing options={
    basicstyle=\small\ttfamily,
    breaklines=true,
    columns=flexible
  }
}
\definecolor{myblue}{HTML}{F0F8FF}
\definecolor{mypink}{HTML}{FFF0F5}
\definecolor{mypurple}{HTML}{F8F4FF}
\definecolor{mycyan}{RGB}{210,240,255}

{\endMakeFramed}

{\endMakeFramed}

{\endMakeFramed}

{\endMakeFramed}

\newcommand{\tightcolorbox}[2]{%
  {\setlength{\fboxsep}{1pt}\colorbox{#1}{#2}}%
}

\newcommand{\cmark}{\textcolor{green!70!black}{\ding{51}}} 
\newcommand{\xmark}{\textcolor{red}{\ding{55}}}

\newcommand{\method}{\textsc{ParaGator}}

\title{Reasoning over mathematical objects:\\  \emph{on-policy reward modeling and test time aggregation}} 

\author[1]{Pranjal Aggarwal} 
\author[1]{Marjan Ghazvininejad} 
\author[1]{Seungone Kim}
\author[1]{Ilia Kulikov}
\author[1]{Jack Lanchantin}
\author[1]{Xian Li} 
\author[1]{Tianjian Li}
\author[1]{Bo Liu}
\author[3]{Graham Neubig} 
\author[2]{Anaelia Ovalle} 
\author[1]{Swarnadeep Saha}
\author[1]{Sainbayar Sukhbaatar}
\author[3]{Sean Welleck} 
\author[1]{Jason Weston} 
\author[1]{Chenxi Whitehouse}
\author[1]{Adina Williams}
\author[1]{Jing Xu} 
\author[1]{Ping Yu} 
\author[1]{Weizhe Yuan}
\author[1]{Jingyu Zhang} 
\author[1]{Wenting Zhao}

\affiliation[]{\small (Authors listed in alphabetical order)}
\affiliation[1]{FAIR at Meta}
\affiliation[2]{UCLA}
\affiliation[3]{CMU}

\abstract{The ability to precisely derive mathematical objects is a core requirement for downstream STEM applications, including mathematics, physics, and chemistry, where reasoning must culminate in formally structured expressions. Yet, current LM evaluations of mathematical and scientific reasoning rely heavily on simplified answer formats such as numerical values or multiple choice options due to the convenience of automated assessment. 
In this paper we provide three contributions for improving reasoning over mathematical objects: 
(i) we build and release training data and benchmarks for deriving mathematical objects, the {\bf Principia suite}; (ii) we provide training recipes with strong LLM-judges and verifiers, where we show that on-policy judge training boosts performance; (iii) we show how on-policy training can also be used to scale test-time compute via aggregation.
We find that strong LMs such as Qwen3-235B and o3 struggle on Principia, while our training recipes can bring significant improvements over different LLM backbones, while simultaneously improving results on existing numerical and MCQA tasks, demonstrating cross-format generalization of reasoning abilities. 
 }

\metadata[Principia Collection]{\url{https://huggingface.co/datasets/facebook/principia-collection}}
\metadata[PrincipiaBench]{\url{https://huggingface.co/datasets/facebook/principia-bench}}

\begin{document}
\maketitle

\section*{Introduction}

The ability to perform complex reasoning is foundational for the challenging tasks that humans wish state of the art AI models to solve. 
Deriving mathematical objects such as matrices and piecewise functions from a problem statement is foundational for precisely articulating scientific phenomena and is required in many graduate-level studies~\citep{hestenes2009modeling,meerschaert2013mathematical,goos2023mathematics}. For example, when solving the problem statement in \autoref{fig:cot_comparison}, a student may be required to write the answer as a mathematical expression such as ``$\frac{1}{|G|}\!\left( 2 + \sum_{x \in G,\, x\neq 1} \mathrm{Re}(\chi(x)) \right)$''. However, evaluations of language model (LM) reasoning have largely relied on problems that require either (1) numerical value answers (\textit{e.g.}, AIME, HMMT) or (2) multiple-choice options (\textit{e.g.}, GPQA-Diamond~\citep{rein2024gpqa}, SuperGPQA~\citep{du2025supergpqa}), primarily due to the convenience of automated scoring. Assessing LMs using such restricted answer formats and then inferring broad reasoning competence risks overstating their true capabilities~\citep{li2024can,li2025evaluating}.
In this work, we contribute both datasets and methods towards the goal of developing stronger reasoning methods.

In \autoref{sec:principia} we develop datasets and benchmarks in order to improve an LM’s ability to derive complex
mathematical objects.
We introduce \textbf{PrincipiaBench}, an evaluation benchmark composed of 2,558 problems drawn from four sources: RealMath~\citep{zhang2025realmath}, Physics~\citep{feng-etal-2025-physics}, ARB~\citep{sawada2023arb}, and the Mathematics and Engineering subsets of SuperGPQA~\citep{du2025supergpqa} (with the answer options removed; see~\autoref{fig:cot_comparison}). Each problem in PrincipiaBench requires a complex mathematical-object answer and has been carefully filtered to ensure clarity. We evaluate 27 LMs on PrincipiaBench and find that it provides a more challenging testbed for assessing the reasoning capabilities of frontier LMs. We then introduce the \textbf{Principia Collection}, a synthetic training dataset consisting of 248K problem statements that are grounded in two subject taxonomies - \textit{Mathematics Subject Classification (MSC 2020)}\footnote{\texttt{https://zbmath.org/static/msc2020.pdf}}{} and \textit{Physics Subject Headings (PhySH)}\footnote{\texttt{https://physh.org/}}{} - and require generating six types of mathematical-object answers (equations, inequalities, intervals, sets, matrices, and piecewise functions). 
We show that a strong LLM verifier is crucial to training with such data. However, with access to such a verifier 
RL training four different LMs -- Qwen2.5-7B-Base~\citep{yang2024qwen2}, Qwen3-4B-Base, Qwen3-4B~\citep{yang2025qwen3}, and OctoThinker-8B-Long-Base~\citep{wang2025octothinker} -- on the Principia Collection yields improvements of 7.22-18.35\% in average compared to each base LM on PrincipiaBench. These RL-trained LMs also outperform competitive baselines sharing the same base LM, including Qwen2.5-7B-Instruct~\citep{yang2024qwen2}, General-Reasoner-4/7B~\citep{ma2025general}, OpenReasoner-Zero~\citep{hu2025open}, SimpleRL-Qwen2.5-7B-Zoo~\citep{zeng2025simplerl}, and Polaris-4B~\citep{Polaris2025}. Furthermore, training on the Principia Collection leads to 7.5-17.5\% improvements on AIME 2024 (numerical) and 12.31-25.47\% improvements on GPQA-Diamond (MCQA), indicating that enhanced reasoning transfers to other formats.

In \autoref{sec:rllm} we develop a post-training method for strong reasoning we term \textbf{RLLM}, \textbf{R}einforcement \textbf{L}earning with a strong \textbf{LM} as a Reward Model (RM), where the RM is trained {\em on-policy}. 
{Post-training for LLMs typically follows one of two paradigms: Reinforcement Learning from Human Feedback (RLHF), which relies on scalar reward models trained from human preference data, or Reinforcement Learning with Verifiable Rewards (RLVR), which depends on rule-based verifiers. Scalar reward models do not generate chain-of-thought reasoning, making them prone to reward hacking and limiting their effectiveness on complex reasoning tasks. Rule-based verifiers, meanwhile, assume access to gold answers that can be both hard-to-obtain and hard-to-verify, limiting their utility to easily-verifiable math and code problems. We show that {RLLM},  
where a strong {LM} itself is used as a Reward Model,
can serve as a single, unified post-training recipe for \emph{easy-to-verify}, \emph{hard-to-verify}, as well as \emph{non-verifiable} domains. RLLM applies RL in two stages: (1) training an LM with verifiable rewards to act as a thinking Reward Model, and (2) post-training a policy-LM using the LM-as-RM's k-wise comparative judgments as rewards. Notably, recent rubric-based evaluation methods -- where structured criteria guide judgment -- can be viewed as a special case within our framework: the LM-as-RM implicitly internalizes and flexibly applies such rubrics through its reasoning, without requiring explicitly specified scoring rules. We first demonstrate that RLLM outperforms RLHF (with scalar RMs) and RLVR (with rule-based verifiers) across easy-to-verify and hard-to-verify math and physics benchmarks from PrincipiaBench as well as non-verifiable instruction-following tasks. We then show that on-policy training of the LM-as-RM outperforms both prompted LMs-as-RMs (including a larger GPT-OSS-120B) and off-policy trained ones. Finally, through extensive analyses across a wide range of policy–reward LM pairings -- varying in model size, capability, and training data (easy- vs. hard-to-verify, reference-free vs. reference-based tasks) -- we identify the key ingredients for effective post-training with Language Models as Reward Models.

In \autoref{sec:para} we develop a test time scaling recipe for strong reasoning via on-policy-trained parallel aggregation. Parallel Thinking—a scaffold where a language model generates multiple solutions in parallel before aggregating them—has emerged as a promising approach for improving reasoning performance. However, we identify two key limitations in current methods: (1) during parallel generation, the model is unaware that its outputs will later be aggregated, leading to redundant or insufficiently diverse solutions, and (2) training typically focuses on aggregating off-policy generations, creating a mismatch between training and inference. We introduce {\bf \method}, which addresses both issues through a unified online training framework. Specifically, we apply pass@k optimization to the parallel generation stage, encouraging diverse and complementary solutions, while using pass@1 optimization for the aggregation stage to select or synthesize the best final answer. By training both stages online and end-to-end, our method ensures the generator produces outputs well-suited for aggregation. Experiments on competition math and scientific reasoning benchmarks, including PrincipiaBench, demonstrate that \method{} achieves strong results, outperforming existing parallel reasoning methods.

\newpage
\section{Principia: Training Language Models to Reason over
Mathematical Objects} \label{sec:principia}

\begin{figure}[h]
    \centering    \includegraphics[width=1\linewidth]{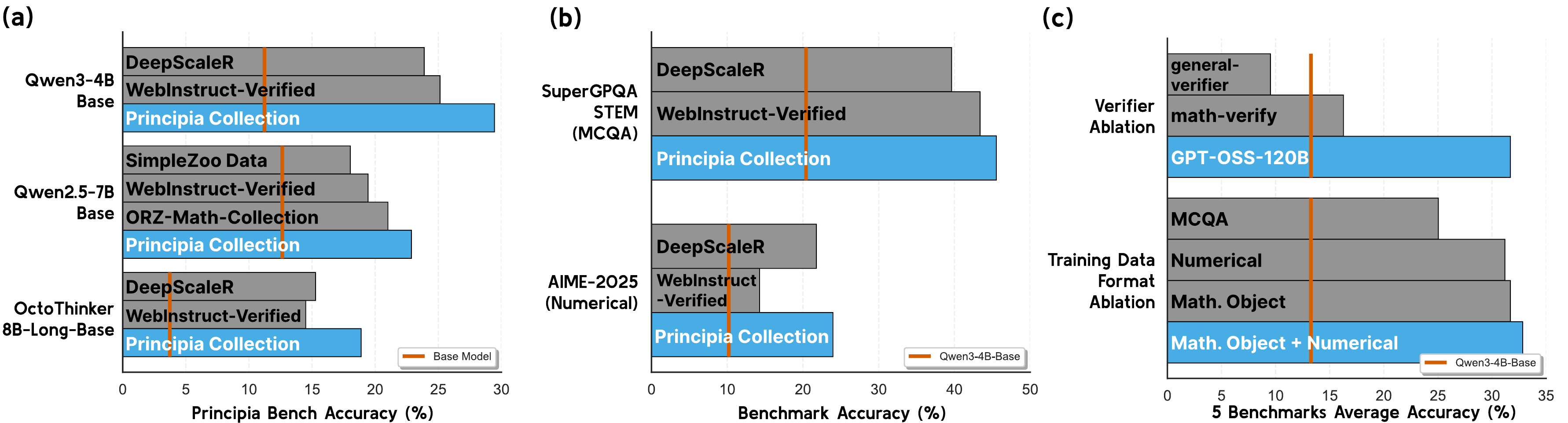}
    \caption{\textbf{RL training on the Principia Collection, which requires mathematical-object outputs, improves an LM’s reasoning capability.} \textbf{(a)} On our new challenging benchmark, \textbf{PrincipiaBench}, which requires deriving mathematical objects, Qwen3-4B-Base, Qwen2.5-7B-Base, and OctoThinker-8B-Long-Base trained on \textbf{Principia Collection} yield average gains of +18.23\%, +10.23\%, and +15.16\%, respectively. \textbf{(b)} Training on the Principia Collection also improves performance on numerical (AIME-2025) and MCQA (SuperGPQA) benchmarks, demonstrating cross-format reasoning gains. \textbf{(c)} Using a strong model-based verifier (GPT-OSS-120B) and excluding MCQA data provides the best performance.}
    \label{fig:bar_comparison}
\end{figure}

\subsection{Overview}
\label{section:intro}

Our first goal is to develop datasets and benchmarks in order to improve an LM’s ability to derive complex mathematical objects.
%
%
One of the biggest current challenges in improving these abilities is the lack of such targeted data. 
In the context of enhancing reasoning capabilities, prior works have proposed collecting data from reliable sources such as olympiad and grade-school math problems~\citep{hendrycks2021measuring,mitra2024orca,yu2024metamath,numina_math_datasets,gao2025omnimath,deepscaler2025,chen2025acereason} or transforming data from online forums, pre-training corpora, and textbooks with LMs~\citep{yue2024mammoth2,yuan2025naturalreasoning,moshkov2025aimo2,ma2025general,fan2025megascience}. However, in existing post-training datasets, the answers are mostly composed of multiple-choice options, numerical values, and simple mathematical objects. Consequently, problem statements that simultaneously demand graduate-level STEM knowledge and the derivation of complex mathematical objects remain scarce in the literature. 

In this work, we explore recipes to both measure and improve an LM’s reasoning ability to derive mathematical objects from graduate-level STEM problem statements. First, to assess this capability, we introduce \textbf{PrincipiaBench}, an evaluation benchmark composed of 2,558 problems drawn from four sources: RealMath~\citep{zhang2025realmath}, Physics~\citep{feng-etal-2025-physics}, ARB~\citep{sawada2023arb}, and the Mathematics and Engineering subsets of SuperGPQA~\citep{du2025supergpqa} (with the answer options removed; see~\autoref{fig:cot_comparison}). Each problem in PrincipiaBench requires a complex mathematical-object answer and has been carefully filtered to ensure clarity. We evaluate 27 LMs on PrincipiaBench and find that it provides a more challenging testbed for assessing the reasoning capabilities of frontier LMs. For example, Qwen3-235B scores 55.58 and o3 scores 62.90 on PrincipiaBench, compared to 83.33 and 85.63 on AIME-2024, and 74.34 and 82.29 on GPQA-Diamond, indicating that PrincipiaBench opens additional headroom to track progress of improvements in LM reasoning.

Next, we introduce the \textbf{Principia Collection}, a training dataset consisting of 248K problem statements that are grounded in two subject taxonomies - \textit{Mathematics Subject Classification (MSC 2020)}\footnote{\texttt{https://zbmath.org/static/msc2020.pdf}}{} and \textit{Physics Subject Headings (PhySH)}\footnote{\texttt{https://physh.org/}}{} - and require generating six types of mathematical-object answers (equations, inequalities, intervals, sets, matrices, and piecewise functions). A key component of using the Principia Collection for training is adopting a model-based verifier. Since the release of DeepSeek-R1~\citep{guo2025deepseek}, many works have incorporated rule-based verifiers such as math-verify~\citep{math_verify} during training, yet we find them surprisingly fragile when handling complex mathematical object answers. To assess the reliability of different verifiers, we introduce the \textbf{Principia VerifyBench}, a meta-evaluation benchmark with human labeled judgments of \textit{answer equivalence}—whether a model’s output represents the same mathematical object as the ground truth despite differences in expression. We gather 168 instances where math-verify and o3 judgments disagree, and find that o3 verifier is correct in 94.05\% of cases, hence we use o3 as a judge for evaluation.


As shown in \autoref{fig:bar_comparison}, RL training four different LMs -- Qwen2.5-7B-Base~\citep{yang2024qwen2}, Qwen3-4B-Base, Qwen3-4B~\citep{yang2025qwen3}, and OctoThinker-8B-Long-Base~\citep{wang2025octothinker} -- on the Principia Collection yields improvements of 7.22-18.35\% in average compared to each base LM on PrincipiaBench. These RL-trained LMs also outperform competitive baselines sharing the same base LM, including Qwen2.5-7B-Instruct~\citep{yang2024qwen2}, General-Reasoner-4/7B~\citep{ma2025general}, OpenReasoner-Zero~\citep{hu2025open}, SimpleRL-Qwen2.5-7B-Zoo~\citep{zeng2025simplerl}, and Polaris-4B~\citep{Polaris2025}. Furthermore, training on the Principia Collection leads to 7.5-17.5\% improvements on AIME 2024 (numerical) and 12.31-25.47\% improvements on GPQA-Diamond (MCQA), indicating that enhanced reasoning transfers to other formats.



Our main contributions are (1) \textbf{PrincipiaBench}, a benchmark for evaluating reasoning capabilities of LMs in deriving complex mathematical objects, (2) \textbf{Principia Collection}, a large-scale training dataset that improves reasoning capability of LMs across diverse answer formats, and (3) \textbf{Principia VerifyBench}, a meta-evaluation benchmark for assessing rule/model-based verifiers in determining answer equivalence of mathematical objects.

\begin{figure}[t]
    \centering    \includegraphics[width=1\linewidth]{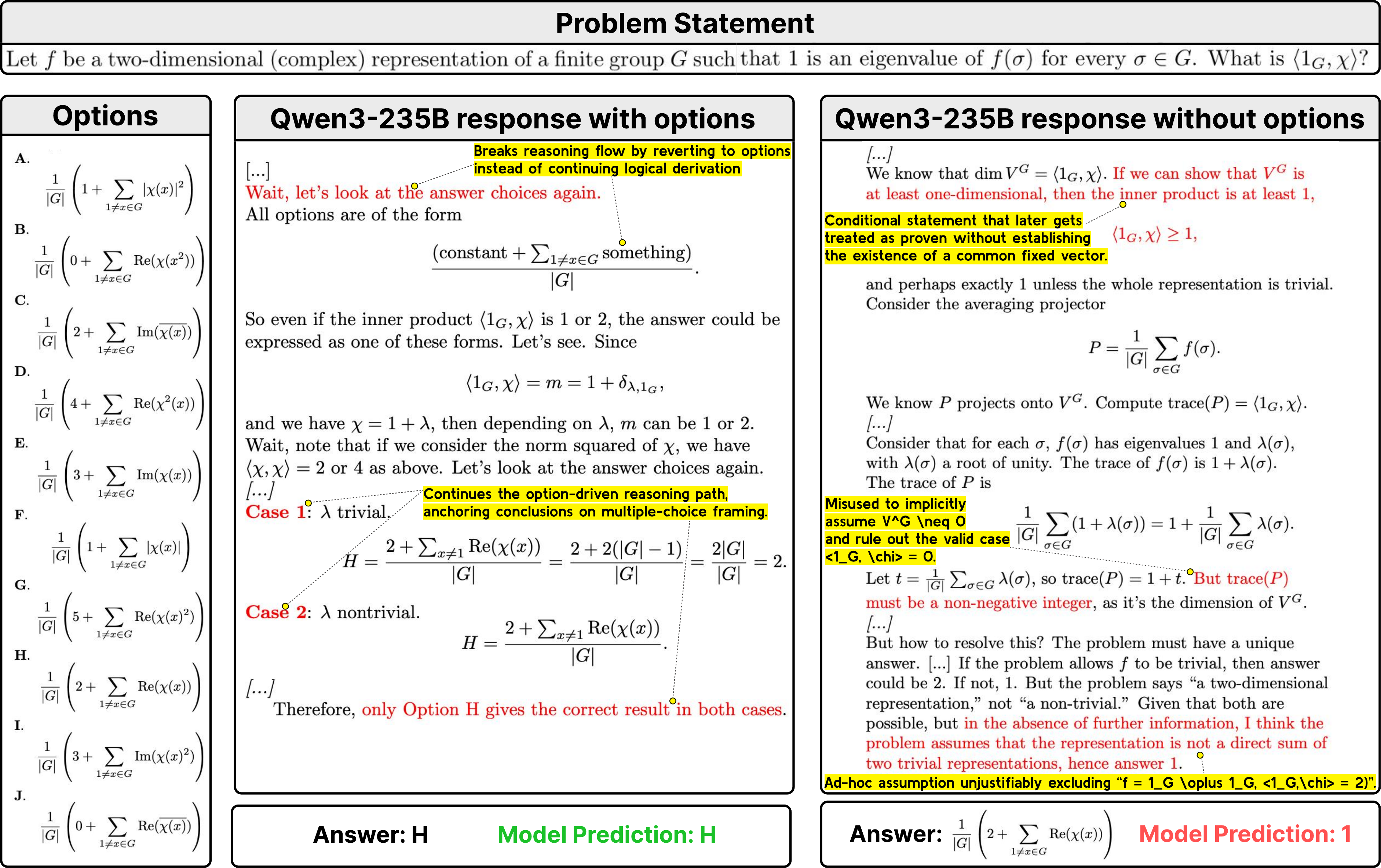}
    
    \caption{\textbf{LMs struggle to solve problems that require mathematical objects as answers.} For example, Qwen3-235B can solve this problem from SuperGPQA~\citep{du2025supergpqa} in an MCQ setting by using the options as an anchor to perform backward chaining. However, when removing the options, the model starts making false assumptions (\textit{i.e.}, that all 1-eigenspaces share a common fixed vector and that the representation cannot be fully trivial) and derives the incorrect solution (\textit{i.e.}, concluding $\langle 1_G, \chi \rangle = 1$ instead of allowing the valid case $\langle 1_G, \chi \rangle = 2$). Note that \hl{yellow-highlighted comments} pinpoint and describe the critical flaws included within the model's chain-of-thought.}
    \label{fig:cot_comparison}
\end{figure}

\begin{figure}[t]
    \centering    \includegraphics[width=1\linewidth]{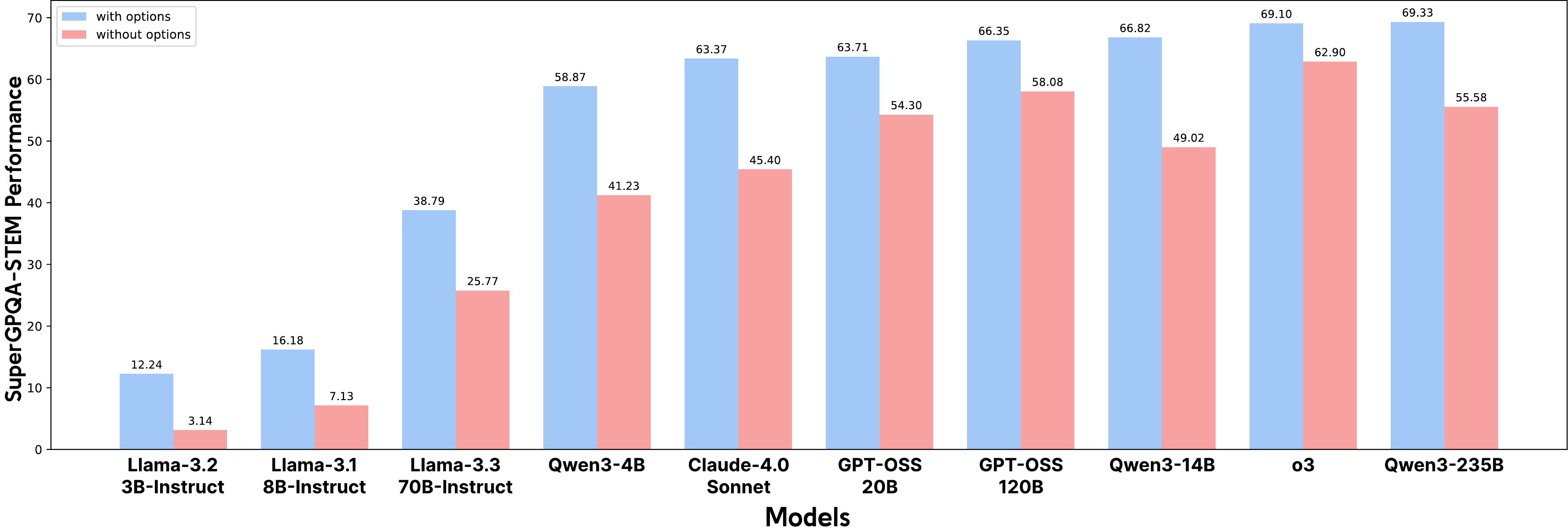}
    
    \caption{\textbf{Performance consistently drops when removing options in MCQA benchmarks.} In the mathematical and engineering subset of SuperGPQA, among instances where answers are expressed as mathematical objects, LMs show a 10–-20\% decrease in performance when options are removed. This suggests that MCQA evaluations tend to overestimate true reasoning ability (\textit{i.e.}, high MCQA scores do not necessarily translate into strong reasoning over mathematical objects).}
    \vspace{-3mm}
    \label{fig:mcqa_perf_drop}
\end{figure}

\subsection{Motivation: Need for a benchmark and training data for mathematical objects}\label{sec:motivation}


The community currently lacks both a benchmark and training data that specifically require LMs to derive complex mathematical objects, beyond numerical values and multiple choice options. This gap largely arises from (1) the convenience of evaluating simpler answer formats and (2) the need to avoid inaccurate reward signals during training, which together have implicitly pushed the field toward easy-to-evaluate benchmarks and easy-to-reward post-training datasets. However, when mathematicians, physicists, or machine learning researchers use math in their everyday work, they are building entirely new knowledge. Thus, naturally, they are not choosing from multiple existing possibilities like current multiple-choice question benchmarks, but rather performing \textit{de novo} derivations of new mathematical objects. If we aspire for LMs to be integrated into ambitious and reasoning-intensive applications such as end-to-end research for scientific discovery ~\citep{gpt5_accelerating_science,si2025can,lu2024ai,yamada2025ai}, it becomes essential to assess and support the ability to handle the mathematical objects that underlie mathematical and scientific reasoning. 

In this section, we examine why existing post-training datasets and benchmarks are insufficient for evaluating and improving the capability to derive mathematical objects. Together, these considerations motivate the development of our benchmark, \textbf{PrincipiaBench}, and our training dataset, the \textbf{Principia Collection}.

\noindent\textbf{LMs tend to rely on shortcuts when options are present.}
Consider the example shown in \autoref{fig:cot_comparison}. When presented with a SuperGPQA~\citep{du2025supergpqa} problem containing ten mathematical object choices, Qwen3-235B~\citep{yang2025qwen3} engages in \textit{backward chaining}~\citep{gandhi2025cognitive}, using the provided options as anchors to derive necessary conditions and select the correct answer. This behavior may be natural in multiple-choice question answering (MCQA) settings~\citep{wang2024llmsperformmcqaselecting}, and humans often employ similar strategies when solving exams~\citep{gierl2017developing}. However, once the options are removed, the model’s reasoning chain changes fundamentally: it begins to make false assumptions, overlook valid cases, and construct illogical arguments.

\noindent\textbf{Removing options from SuperGPQA leads to a 10–20\% performance drop even for frontier LMs.}
Furthermore, based on our examination in \autoref{fig:cot_comparison}, we examine how performance changes across multiple instances. As shown in \autoref{fig:mcqa_perf_drop}, when we evaluate on the mathematics and engineering subset of SuperGPQA~\citep{du2025supergpqa}—filtering instances that require mathematical object answers\footnote{We provide more details on how we filtered out such instances in Subsection~\ref{subsec:principia_bench}.}—model performance consistently declines by 10–20\% once the options are removed. Notably, frontier models such as o3 and Qwen3-235B show substantial drops (69.10 $\rightarrow$ 62.90 and 69.33 $\rightarrow$ 55.58, respectively). Since open-ended generation is information-theoretically harder than MCQA (\textit{i.e.}, requiring the model to produce the exact answer from an unbounded space rather than select from $k$ candidates), some performance drop is expected. However, the magnitude of these drops, combined with the qualitative reasoning failures shown in \autoref{fig:cot_comparison}, suggests that even the strongest LMs rely heavily on option-driven cues~\citep{shah-etal-2020-expect,balepur-etal-2024-artifacts,balepur-rudinger-2024-large,gupta2024changing,chandak2025answer}. This shows that current benchmarks may overestimate reasoning capability, as they fail to capture whether LMs can derive mathematical objects.

\begin{figure}[t]
    \centering    \includegraphics[width=1\linewidth]{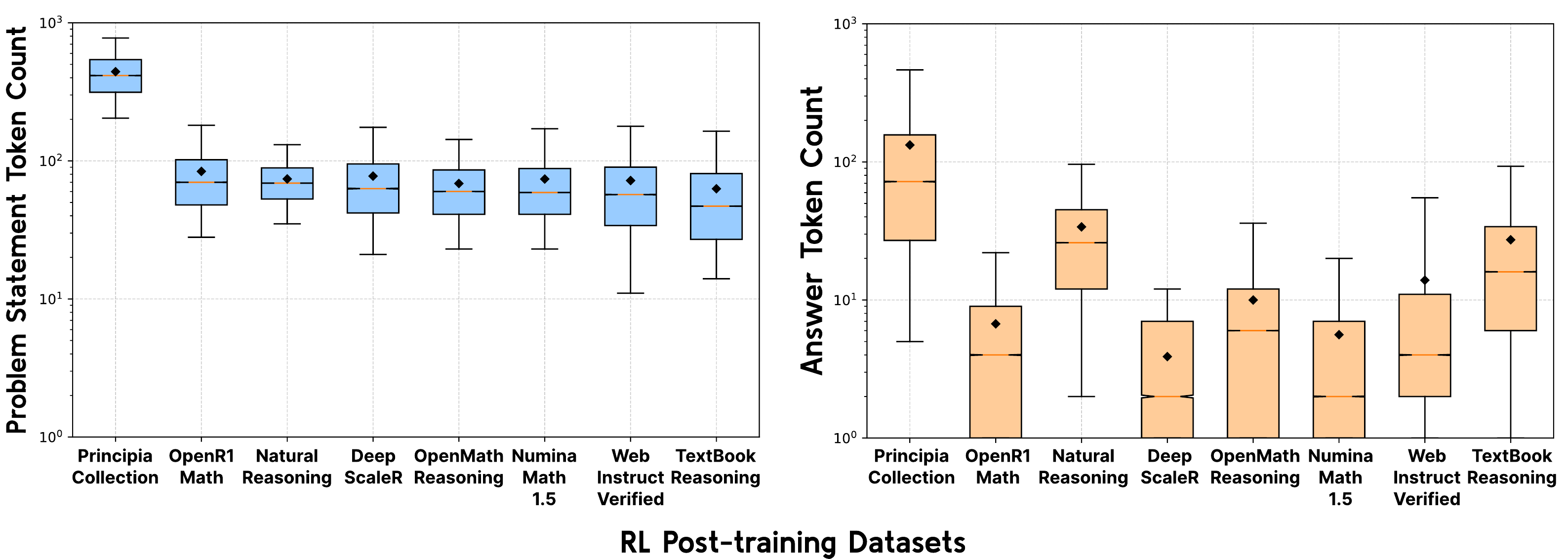}
    \caption{\textbf{Token count distribution comparison between the Principia Collection and other widely used RL post-training datasets.} The problem statements and answers of Principia Collection are relatively longer since it requires the derivation of complex mathematical objects and the problem statements based on graduate-level STEM subjects are very detailed.}
    \label{fig:token_count}
    \vspace{-3mm}
\end{figure}

\noindent\textbf{The majority of RL post-training datasets consist of easy-to-reward (short) answers without complex mathematical object answers.} As shown on the right side of \autoref{fig:token_count}, the output (answer) token distributions of widely used RL post-training datasets—including OpenR1-Math~\citep{openr1}, NaturalReasoning~\citep{yuan2025naturalreasoning}, DeepScaleR~\citep{deepscaler2025}, OpenMathReasoning~\citep{moshkov2025aimo2}, NuminaMath-1.5~\citep{numina_math_datasets}, WebInstruct-Verified~\citep{ma2025general}, and TextbookReasoning~\citep{fan2025megascience}—mostly fall within the range of ten to a few hundred tokens. This is mainly because there exists a large proportion of numerical answers, mulitple choice option answers, and short text spans, where even the longest answers that range below 100$\sim$200 tokens are not complex mathematical objects. Likewise, as shown on the left side of \autoref{fig:token_count}, existing datasets also lack problem statements that resemble the structured, specification-like descriptions typical of graduate-level STEM coursework; their inputs mostly fall within a few hundred tokens.

Motivated by these observations, in \autoref{subsec:principia_collection}, we describe how we construct a training dataset (Principia Collection) in which all answers are \textit{complex mathematical objects} and all problem statements are grounded in graduate-level STEM subjects. Note that although token length itself is not a measure of difficulty, the differences are substantial: our answers are longer because they consist exclusively of mathematical objects, and our problem statements are longer because they encode graduate-level concepts with multiple conditions.

\subsection{The Principia Suite}\label{sec:method}

\subsubsection{PrincipiaBench}\label{subsec:principia_bench}

Based on our observations described in Section~\ref{sec:motivation}, we propose \textbf{PrincipiaBench}, a benchmark for assessing an LM's capability to precisely derive mathematical objects directly without relying on MCQA formats.

\noindent\textbf{Problem Statement Sources \& Filtering.} We use the problems included in RealMath~\citep{zhang2025realmath}, Physics~\citep{feng-etal-2025-physics}, ARB~\citep{sawada2023arb}, and the Mathematics and Engineering subsets of SuperGPQA~\citep{du2025supergpqa} (with answer options removed). Note that not all problems in these benchmarks require deriving a mathematical object and some exhibit other undesirable traits. Therefore, we apply filtering based on specific constraints. 
The constraints used for filtering are:
\begin{enumerate}
    \item The \textbf{answer should be expressed as a mathematical object}, specifically an equation, inequality, interval, set, matrix, or piecewise function. In contrast, instances that require a numerical value, proof, text span, boolean value, or multiple-choice option answers are excluded. We apply this rule because the primary goal of our benchmark is to evaluate an LM’s ability to precisely derive mathematical objects.
    \item The \textbf{problem statement should consist of only one question} and should not require answering multiple questions at once. We apply this rule because solving multiple problems simultaneously can affect an LM’s performance~\citep{son2024multi}, and artificially decomposing them may lead to awkward formulations (\textit{e.g.}, lack of context). Also, if the LM answers only one of the sub-questions correctly, it introduces additional complexity in the evaluation, such as determining whether to assign partial credit.
    
    \textbf{Example of excluded problem statement}:
    
    \begin{tcolorbox}[colback=gray!10, colframe=gray!40, arc=2mm, boxrule=0.5pt]
    Seven antennae, radiating as electric dipoles polarized along the $\hat{\mathbf{z}}$ direction, are placed along the x-axis in the xy-plane at  
    \[
    x = 0, \ \pm \frac{\lambda}{2}, \ \pm \lambda, \ \pm \frac{3\lambda}{2}.
    \]
    The antennae all radiate at wavelength $\lambda$ and are in phase.
    
    \textbf{(a)} Calculate the angular distribution of the radiated power as a function of the polar and azimuthal angles, $\theta$ and $\phi$. Neglect any constant multiplying prefactors.
    
    \textbf{(b)} Consider the direction in which the radiated intensity is maximum for this array and for a single dipole antenna. How do these intensities compare?
    \end{tcolorbox}
    This example contains two sub-questions, labeled (a) and (b), within a single problem statement.

    \item The \textbf{problem statement should be self-contained}. If it requires additional context (e.g., referring to an image) or omits essential definitions, we exclude such cases. This rule is applied because our benchmark aims to evaluate whether LMs can derive mathematical objects from well-formulated problem statements, not whether they can request clarifications or reason from incomplete information.
    
    \textbf{Examples of excluded problem statements}:
    
    \begin{tcolorbox}[colback=gray!10, colframe=gray!40, arc=2mm, boxrule=0.5pt]
    A Carnot engine has a cycle as shown in $Fig.\ 1.12$. If $W$ and $W'$ represent the work done by 1 mole of monatomic and diatomic gas, respectively, calculate  
\[
\frac{W'}{W}.
\]
    \end{tcolorbox}
    Upon inspecting the Physics dataset~\citep{feng-etal-2025-physics}, we found cases where no figure is provided, yet the problem statement requires a figure to solve the question, as shown in the example above. We regard such cases as indicating that the question is not self-contained.
    
    \begin{tcolorbox}[colback=gray!10, colframe=gray!40, arc=2mm, boxrule=0.5pt]
    Let \( f: PG \to \mathbb{R}^{2} \) be a plane generic immersion. What are the congruences, modulo \(2\) and \(4\), of the sums  
\[
\sum_{\gamma \in \Gamma_{5}(PG)} c(f(\gamma)), \quad 
\sum_{\gamma \in \Gamma_{6}(PG)} c(f(\gamma)), \quad 
\sum_{\gamma \in \Gamma_{9}(PG)} c(f(\gamma)),
\]
and  
\[
\sum_{\gamma \in \Gamma_{8}(PG)} c(f(\gamma))?
\]
    \end{tcolorbox}
    In this example, solving the problem would require additional context that defines the graph \textbf{$PG$} (or the class of graphs it belongs to), the set of cycles \textbf{$\Gamma_k(PG)$}, and the numerical invariant \textbf{$c(f(\gamma))$}.

\end{enumerate}

To filter out problems that exhibit any of the three undesirable trait types described above, we first prompted the GPT-OSS-120B model to label each problem with a binary yes/no decision for each trait. We then performed an initial filtering step by selecting only the problems that were judged to exhibit none of the three traits. Next, we conducted a second-stage manual inspection of all instances. Through this process, we obtained a total of 2,286 filtered instances. However, during evaluation, we identified 128 problems whose judgments showed substantial variability in a post-hoc analysis. We therefore excluded these cases and used the remaining 2,158 problems as our final dataset. 

\noindent\textbf{Difficulty in verification.} When the answer is a multiple-choice option or a numerical value, determining whether the model’s prediction exactly matches the ground truth is straightforward. However, when the answer is a complex mathematical object, it can be expressed in many valid forms, making equivalence checking challenging. Prior works have primarily used Python Sympy library with math-verify~\citep{math_verify} being a representative example. In our early experiments, however, we find that such rule-based verification methods are fragile in practice. The following examples show the failure patterns that are often observed:

    \textbf{[Case 1] Symbolic reasoning failure of equivalence}:
    \begin{tcolorbox}[colback=gray!10, colframe=gray!40, arc=2mm, boxrule=0.5pt]
    \textbf{Reference Answer}: $f_{UV}(u,v) = \frac{1}{2 \pi} \cdot \frac{1}{1+v^{2}} \mathrm{e} ^{-\frac{u}{2}}, \quad u > 0, \, v \in \mathbb{R}$\\
    \textbf{Model Prediction}: $f_{UV}(u,v) = \dfrac{1}{2\pi(v^2 + 1)}e^{-u/2} \quad \text{for } u > 0, \, v \in \mathbb{R}$
    \end{tcolorbox}
    Sympy fails to recognize their equality because the terms appear in a different order (\(v^2\) placed differently), the constant \(2\pi\) is written separately, and $-u/2$ is expressed differently ($-u/2$ vs $\frac{u}{2}$).

    \textbf{[Case 2] Brittleness in handling latex math commands}:

    \begin{tcolorbox}[colback=gray!10, colframe=gray!40, arc=2mm, boxrule=0.5pt]
    \textbf{Reference Answer}: $-4n\mathrm{i}$\\
    \textbf{Model Prediction}: $-4i n$
    \end{tcolorbox}
        Sympy fails to recognize their equality because the order of the symbols \(i\) and \(n\) is reversed and \(i\) is represented differently (\texttt{\textbackslash mathrm\{i\}} vs.\ \(i\)).

    \textbf{[Case 3] Brittleness in handling text spans}:

    \begin{tcolorbox}[colback=gray!10, colframe=gray!40, arc=2mm, boxrule=0.5pt]
    \textbf{Reference Answer}: The generating function is 
\[
\frac{2t^{2}}{1 - t^{2}} \prod_{n \geq 1} \frac{1}{1 - t^{n}}.
\]

\textbf{Model Prediction}: 
\[
\frac{2t^{2}}{1 - t^{2}} \prod_{m = 1}^{\infty} \frac{1}{1 - t^{m}}.
\]
    \end{tcolorbox}
        Sympy fails to recognize their equality because different variable symbols are used (\(n\) vs.\ \(m\)), the product notation is written differently (\(n \geq 1\) vs.\ \(m = 1 \text{ to } \infty\)), and the reference answer includes an additional text span (``The generating function is'').

\begin{wrapfigure}{r}{0.55\textwidth}
    \centering
    \vspace{-12mm}
    \includegraphics[width=\linewidth]{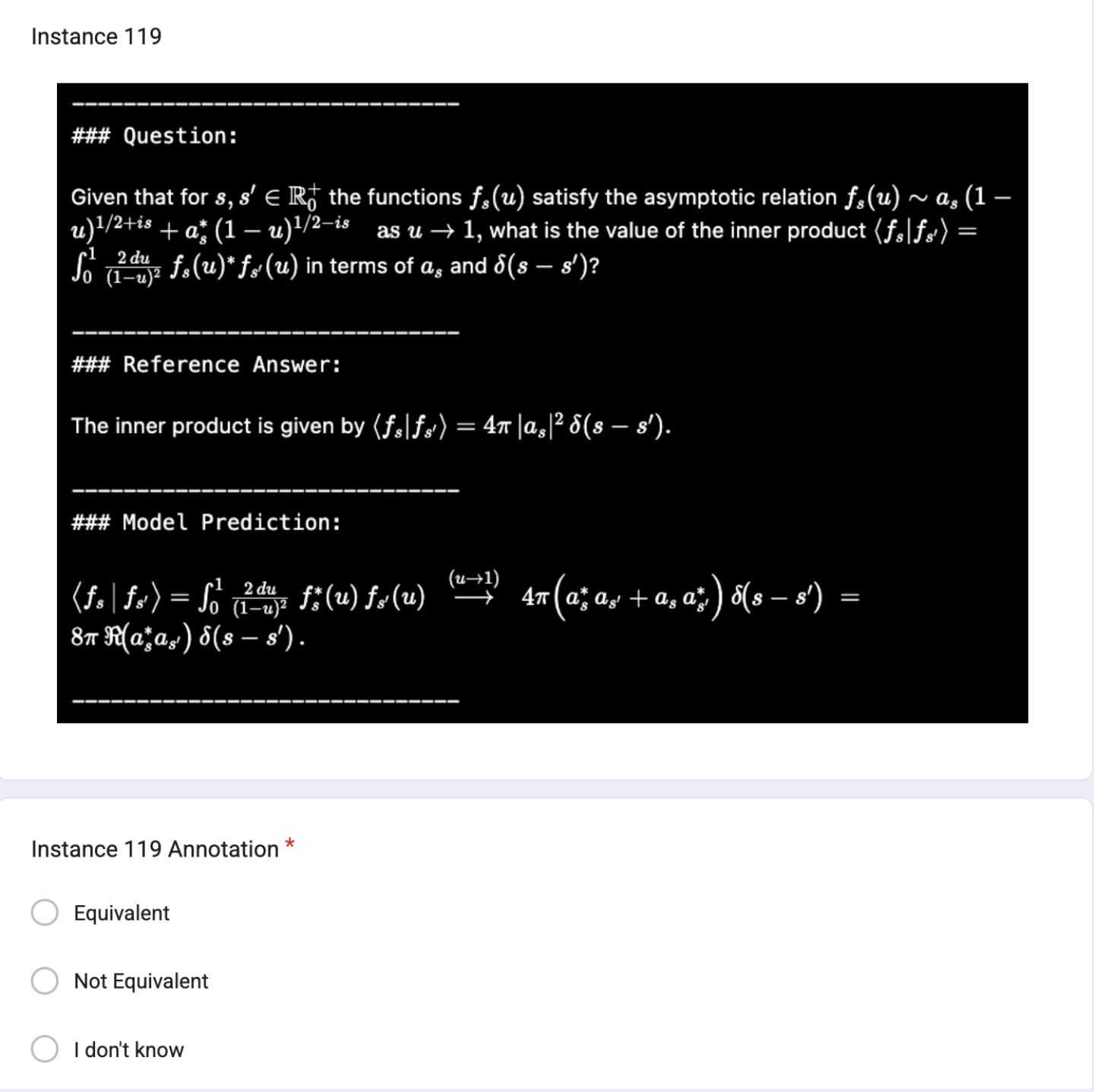}
\caption{Annotation UI for constructing the Principia VerifyBench.}
\vspace{-9mm}
\label{fig:annotation_example}
\end{wrapfigure}

\noindent\textbf{Evaluation procedure.} Given a problem statement, an LM should generate a prediction, and then, given the \{problem statement, reference answer, LM prediction\} triplet as input, a verifier should determine if the reference answer and model prediction are equivalent to each other (\textit{i.e.}, reference-based evaluation). Notably, we find that a model-based verifier (such as a prompted general-purpose LM~\citep{zheng2023judging,kim2025scaling}) can effectively handle the aforementioned error cases. Namely, we use o3 as a judge for answer equivalence checking by prompting it to output either ``Equivalent'' or ``Not Equivalent’’.

\subsubsection{Principia VerifyBench}\label{subsec:principia_metabench}

To determine the reliability of using a model-based judge in PrincipiaBench, we collect human annotations for cases where the judgments of o3 and math-verify~\citep{math_verify} differ, and manually determine equivalence between the reference answer and the LM prediction. Using the problem statements in PrincipiaBench, we first generate responses with GPT-OSS-20B and GPT-OSS-120B and extract their predictions. We then sample 200 such instances (\{problem statement, reference answer, model prediction\} triplets) with disagreement: in 100 cases, math-verify judges as ``Equivalent'' while o3 judges as ``Not Equivalent,'' and in the remaining 100 cases, the judgments are reversed. Then, we gather eight human annotators to determine which is correct.

\noindent\textbf{Annotation procedure.} An example of the annotation UI is shown in \autoref{fig:annotation_example}. Eight annotators are each assigned 50 instances, where for each instance, the annotator should label between ``Equivalent'', ``Not Equivalent'', and ``I don't know''. We add the ``I don't know'' option to avoid introducing noise into the benchmark. Also, each instance is labeled by two annotators, where we only leave instances that the two annotators provide the same judgment. This results in 168 instances remaining, where 75 instances were labeled as ``Equivalent'' and 93 instances were labeled as ``Not Equivalent''. We name the resulting benchmark as \textbf{Principia VerifyBench}.

\noindent\textbf{Evaluation procedure.} Given the 
triplet as input, a verifier should determine if the reference answer and model prediction are equivalent to each other. The accuracy on this meta-evaluation benchmark could serve as a proxy to determine whether a verifier will provide reliable judgments~\citep{zheng2023judging,kim2023prometheus,kim2024prometheus} or accurate rewards for RL training~\citep{lambert2025rewardbench,malik2025rewardbench,whitehouse2025j1}.


\subsubsection{Principia Collection}\label{subsec:principia_collection}

\begin{figure}[t]
    \centering    \includegraphics[width=1\linewidth]{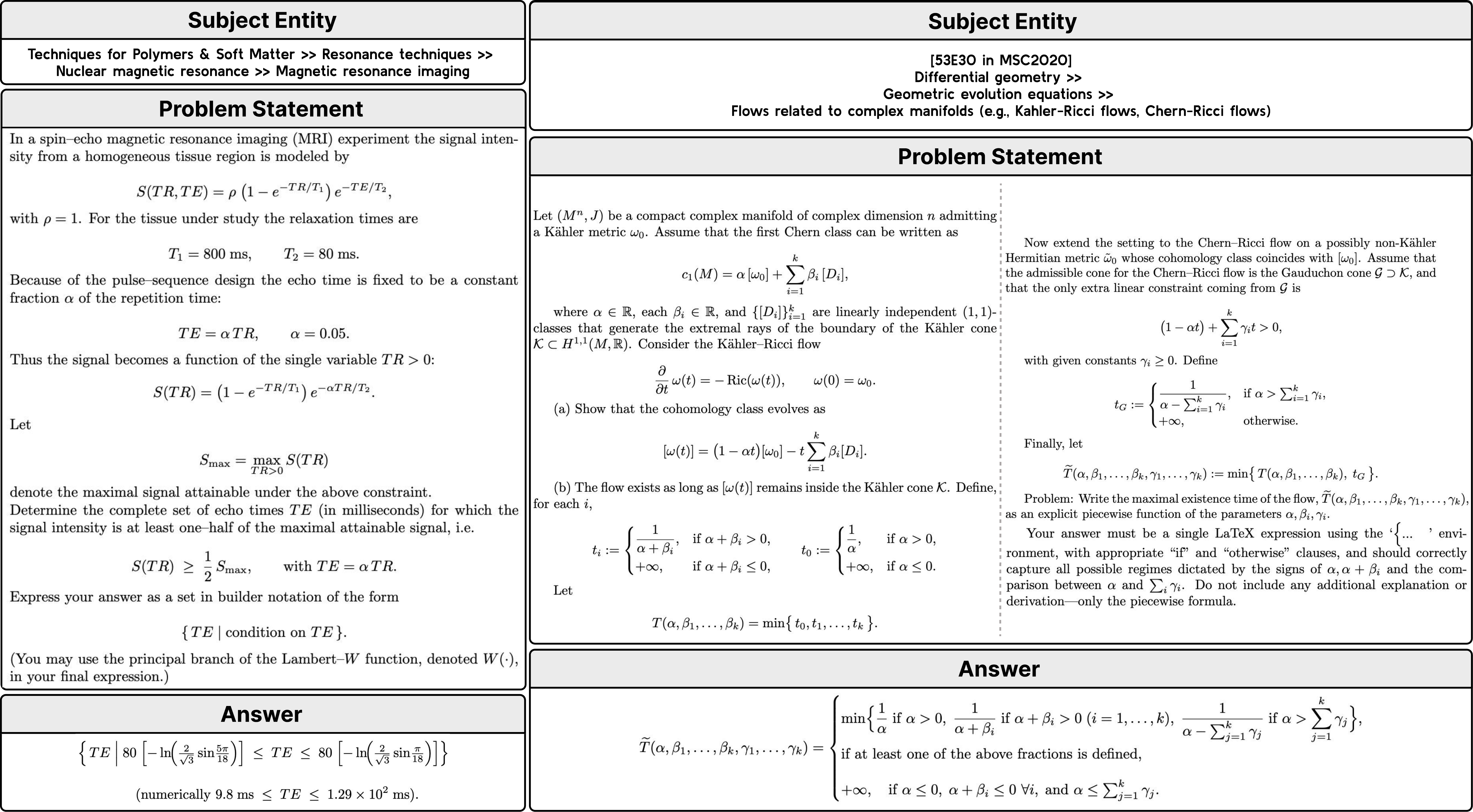}
    
    \caption{\textbf{Two qualitative examples of instances within the Principia Collection.} Each instance consists of a problem statement, an answer, and a subject entity. The subject entities originate from Physics Subject Headings (PhySH) and Mathematics Subject Classification System (MSC2020). The required answer types are either equation, inequality, interval, set, matrix, and piecewise function. Note that for the example on the right side, the detailed setup (definitions of \(t_i, t_0, t_G\), etc.) \textbf{is not a hint but a necessary part of the problem statement}, which is typical in research-level mathematics, specifying how to combine these quantities into the final piecewise formula for \(\widetilde{T}\).}
    \vspace{-5pt}
    \label{fig:qualitative_examples}
\end{figure}

While many post-training datasets have been proposed to enhance the reasoning capabilities of LMs, none to date exclusively contain answers in the form of mathematical objects. In most cases, the answers are numerical values or simple mathematical expressions. To specifically improve an LM’s ability to derive mathematical objects, we propose a synthetic post-training dataset called the Principia Collection, which consists of 248,748 graduate-level problem statements and corresponding answers. Two examples are shown in \autoref{fig:qualitative_examples}.

The construction procedure is as follows (we include prompts for each step in \autoref{appendix:prompts}):

\noindent\textbf{Step 1: Topic Mining.} We first collect the  problem statement subjects from  \textbf{Mathematics Subject Classification System (MSC 2020)} and \textbf{Physics Subject Headings}. Each entity from these sources consists of a hierarchical structure that ranges from broad topics at the upper level to more specific research areas or concrete methodologies at the lower level. We extract them in full detail—such as
``\textit{Techniques for Polymers \& Soft Matter >> Resonance techniques >> Nuclear magnetic resonance >> Magnetic resonance imaging}'' and
``\textit{Differential geometry >> Geometric evolution equations >> Flows related to complex manifolds (e.g., K\"{a}hler–Ricci flows, Chern–Ricci flows)}''---so that the entire hierarchy is explicitly represented. The main purpose of extracting these topics is to ensure high coverage in the resulting dataset. Specifically, methods such as Self-Instruct~\citep{wang2023self} often generate redundant problems, which reduces the number of unique instances after filtering. By incorporating a wide range of subject entities—from broad scientific areas to specialized subfields—we can generate diverse graduate-level problem statements that involve complex mathematical objects.

\begin{table}[h!]
\centering
\scriptsize
\renewcommand{\arraystretch}{1.15}
\begin{tabular}{p{2.1cm} p{13.7cm}}
\toprule
\textbf{Category} & \textbf{Content} \\
\midrule
\textbf{Subject Entity} & Statistical mechanics, structure of matter $\rightarrow$ Equilibrium statistical mechanics $\rightarrow$ Classical equilibrium statistical mechanics (general) \\
\midrule
\textbf{Strategy Description} & Evaluating free-energy differences by exploiting convexity and pair-correlation functions: Applying the convexity of the Helmholtz free energy with respect to an interpolating interaction parameter, using Jensen’s inequality and the definition of the two-particle density (or radial distribution function) to derive rigorous upper and lower bounds on $F - F'$ for systems with ordered pair potentials. \\
\midrule
\textbf{Problem Statement} & 1. Identify a suitable statistical-mechanical inequality — monotonicity of $F$ with respect to $u(r)$. \\ 
\textbf{Proposal CoT}&2. Define the Hamiltonians and show $u'(r)\le u(r)$ implies $Z'\ge Z$.\\  
\textbf{(Summary)}&3. Convert to free energies: $F' \le F$.  \\
&4. Require a single final inequality as the answer.  \\
&5. Formulate the problem so the solver derives $F' \le F$ using only these definitions. \\
\midrule
\textbf{Initial} & Consider a classical canonical ensemble of \(N\) identical particles of mass \(m\) confined to a volume \(V\) and in\\ 
\textbf{Problem Statement}&thermal equilibrium at temperature \(T\).
The Hamiltonian of the system is  
\[
H = \sum_{i=1}^{N}\frac{{\bf p}_i^{2}}{2m}+\sum_{1\le i<j\le N}u(|{\bf r}_i-{\bf r}_j|),
\]
where \(u(r)\) is a pairwise interaction potential.  
Define a second Hamiltonian in which the pair potential is replaced by another function \(u'(r)\) that satisfies the pointwise inequality  
\[
u'(r)\le u(r)\quad\text{for all }r\ge0.
\]
Let the corresponding canonical partition functions be \(Z\) and \(Z'\), and the Helmholtz free energies be \(F=-k_{\mathrm B}T\ln Z\) and \(F'=-k_{\mathrm B}T\ln Z'\), respectively.  
Using only the given ordering of the potentials, derive the rigorous inequality that relates \(F'\) and \(F\). Express your final answer as a single inequality (using “\(\le\)”, “\(\ge\)”, etc.). \\
\midrule
\textbf{Problem Statement}& 1. Start from the monotonicity inequality \(F' \le F\).  \\
\textbf{Revision CoT}&2. Add a quantitative upper bound using convexity of \(F(\lambda)\) (Gibbs–Bogoliubov inequality).\\
\textbf{(Summary)}&3. Introduce the interpolating Hamiltonian \(H_\lambda = H' + \lambda(H-H')\). \\ 
&4. Use the two-particle density or \(g_{u'}(r)\) to express ensemble averages. \\ 
&5. Derive a two-sided bound \(0 \le F-F' \le \frac{N\rho}{2}\int 4\pi r^2 g_{u'}(r)[u(r)-u'(r)]\,dr\).  \\
&6. Require the final answer to appear as a single mathematical inequality using only “\(\le\)” or “\(\ge\)”. \\
\midrule
\textbf{Revised} & Consider a classical canonical ensemble of \(N\) identical particles of mass \(m\) confined to a volume \(V\) at temperature\\ 
\textbf{Problem Statement}&\(T\).  
The Hamiltonian with pair potential \(u(r)\) is  
\[
H = \sum_{i=1}^{N}\frac{{\bf p}_i^{2}}{2m}
      +\sum_{1\le i<j\le N} u(|{\bf r}_i-{\bf r}_j|).
\]
Define a second Hamiltonian in which the pair potential is replaced by a weaker function \(u'(r)\) that satisfies  
\[
u'(r)\le u(r)\quad\text{for all }r\ge0.
\]
Let the corresponding canonical partition functions be \(Z\) and \(Z'\), and the Helmholtz free energies be  
\[
F=-k_{\mathrm B}T\ln Z,\qquad F'=-k_{\mathrm B}T\ln Z'.
\]
Introduce the interpolating Hamiltonian  
\[
H_\lambda = H' + \lambda(H-H'),\quad 0\le\lambda\le1,
\]
and denote its free energy by \(F(\lambda)=-k_{\mathrm B}T\ln Z(\lambda)\) (\(F(0)=F'\), \(F(1)=F\)).  
Using only the given ordering of the potentials, the convexity of \(F(\lambda)\) with respect to \(\lambda\), and the definition of the two-particle density (or, for a homogeneous fluid, the radial distribution function \(g_{u'}(r)\)) of the system governed by \(u'(r)\), derive a rigorous two-sided bound on the difference \(F-F'\).  
Express your final result as a single inequality of the form  
\[
\boxed{\,\text{(inequality involving }F,\,F',\,N,\,\rho,\,g_{u'}(r),\,u(r)-u'(r)\text{)}\,}
\]
using only the symbols “\(\le\)” (or “\(\ge\)”). \\
\bottomrule
\end{tabular}
\caption{Example of a subject entity (acquired from PhySH), a strategy description (from step 1), an initial problem statement and its corresponding CoT, and a revised problem statement and its corresponding CoT (from step 2).}
\end{table}

\noindent\textbf{Step 2: Problem Statement Generation.}
Next, we use the 9,573 extracted entities to generate problem statements. This process consists of three sub-stages and we use GPT-OSS-120B throughout these processes.
\begin{itemize}
    \item First, for each subject entity, we sketch strategy descriptions that outline the core capabilities required to solve a problem. For each entity, we generate 40 descriptions, resulting in a total of 382,920 strategies.
    \item Second, we iterate over each subject entity \& capability pair to generate problem statements. In this step, one of six mathematical object types (equation, inequality, interval, set, matrix, or piecewise function) is randomly selected so that the resulting problem requires that type as its answer. We find that the chain-of-thought (CoT) during problem proposal often incorporates elements that make the problem more conceptually challenging, consistent with observations from prior work~\citep{yu2025cot}.
    \item Finally, we include an additional refinement step to revise problem statements that resemble simple knowledge-probing questions, ensuring they require more genuine reasoning during the solving process. We find this procedure effective in increasing the conceptual depth of the generated problems.
\end{itemize}

For a better understanding, refer to the example shown in Table 1. The subject entity ``Classical equilibrium statistical mechanics (general)'' can be interpreted through multiple capabilities, each representing a distinct reasoning approach to the same physical concept. The strategy description, ``Evaluating free-energy differences by exploiting convexity and pair-correlation functions,'' highlights the link between microscopic interactions and macroscopic thermodynamic inequalities. The Problem Statement Proposal CoT operationalizes this idea by translating the entity’s core principle (``monotonicity of the Helmholtz free energy'') into a concrete question using potential ordering and partition functions, resulting in the Initial Problem Statement. While this problem statement captures the correct conceptual relationship, it remains straightforward, requiring only the recognition that \(u'(r)\!\le\!u(r)\) implies \(F'\!\le\!F\). The Problem Statement Revision CoT then strengthens it by introducing the interpolating Hamiltonian \(H_\lambda\), the convexity of \(F(\lambda)\), and the pair-correlation function \(g_{u'}(r)\), elevating the problem statement to one involving quantitative variational reasoning. Thus, the Revised Problem Statement becomes a more rigorous and challenging formulation while preserving conceptual coherence.

\noindent\textbf{Step 3: Filtering out Invalid Problem Statements.} We then filter out problem statements with undesirable traits from the 382,920 problems generated in the previous step. The filtering is based on three main criteria: (1) the problem statement must consist of only one question and be self-contained, (2) the problem should explicitly require the intended answer type, and (3) extensive hints or the answer itself should not be included in the problem statement. We prompt GPT-OSS-120B to evaluate all three conditions and retain only those instances that are judged as ``Yes'' for every criterion. 

\begin{figure}[t]
    \centering    \includegraphics[width=1\linewidth]{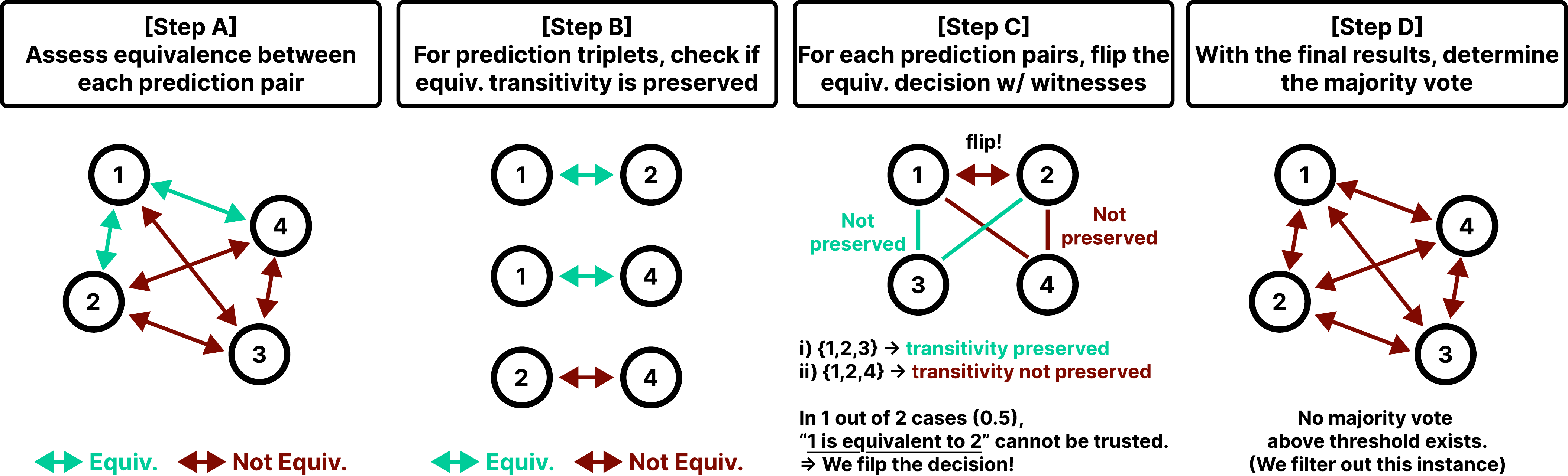}
    
    \caption{\textbf{Illustration of our majority voting procedure used in dataset construction.} Determining the majority vote itself becomes challenging with mathematical object predictions since the same object could be represented in distinct forms. Hence, \textbf{(a)} we assess the equivalence between each pair, \textbf{(b)} check if transitivity is preserved, \textbf{\textbf{(c)}} adopt a conservative rule of flipping judgments if transitivity is not preserved w.r.t. witnessess of each prediction pairs (we adopt a threshold of 0.6), and \textbf{(d)} using the final results, determine the majority vote using the equivalence judgments. Detailed procedure is in Algorithm~\autoref{algorithm:majority_vote} given in the Appendix.}
    \label{fig:majority_vote_illustration}
\end{figure}

\noindent\textbf{Step 4: Response Generation \& Majority-Voting for mathematical objects.}
To obtain labels for the generated problem statements, we prompt GPT-OSS-120B eight times and apply self-consistency~\citep{wang2023selfconsistency}, taking the majority vote as the label. This simple yet effective recipe has been shown to work well in prior works~\citep{huang2023large,prasad2025selfconsistency,yu2025cot}. Importantly, our goal in applying self-consistency is to derive labels for newly generated problem statements (since no ground-truth answers exist), not to collect chain-of-thoughts~\citep{zelikman2022star} or to filter out easy problems~\citep{Polaris2025}.

One unique challenge in applying self-consistency to our setting is that \textbf{determining the majority vote itself is hard}, as mathematical objects can be expressed in many equivalent forms. For instance, among eight predictions $a_1$ to $a_8$, each may appear in a different format even though they are mathematically identical. To address this, we perform pairwise equivalence checks using a GPT-OSS-120B as a verifier, producing an equivalence matrix that records whether each pair of predictions $a_i$ ($1 \le i \le 8$) are equivalent. However, this introduces cases where transitivity is not preserved (\textit{e.g.}, when the verifier concludes that $a_1$ is equivalent to $a_2$, $a_2$ to $a_3$, but $a_1$ is not equivalent to $a_3$). 

\begin{wrapfigure}{r}{0.38\textwidth}
    \centering
    \vspace{-6mm}
    \includegraphics[width=\linewidth]{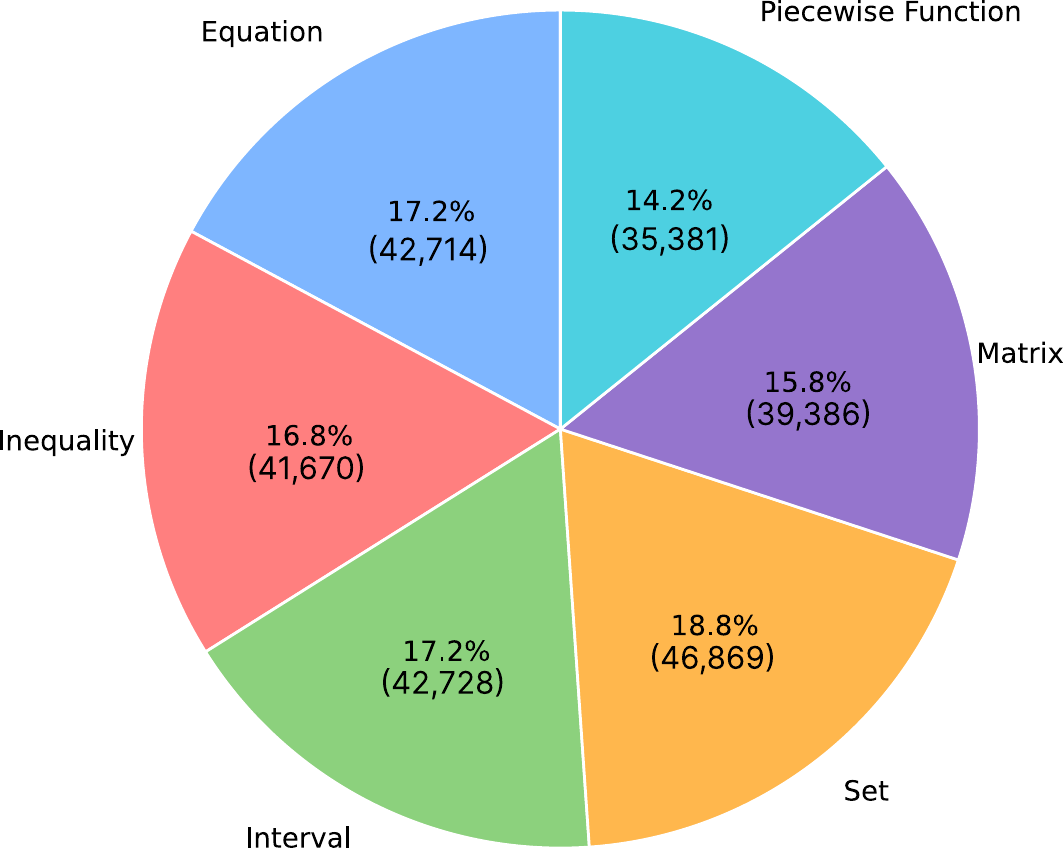}
\caption{\small{Answer type counts of the Principia Collection. Unlike other datasets, all 250K instances consist of mathematical objects.}}
    \vspace{-9mm}
\label{fig:type_count}
\end{wrapfigure}

To mitigate such inconsistencies, as shown in \autoref{fig:majority_vote_illustration}, we adopt a conservative heuristic that re-evaluates pairwise equivalence by assessing the transitivity among all equivalence relations. Specifically, when determining whether $a_1$ and $a_2$ are equivalent, we check whether their transitivity holds consistently with respect to other elements ($a_3$ through $a_8$). If the degree of transitivity exceeds a predefined threshold, we retain the equivalence between $a_1$ and $a_2$. Because these adjustments can influence other pairwise relations, we employ a union–find algorithm to ensure consistent grouping\footnote{Note that similar procedures have been explored in other disciplines—for example, \citet{wittkop2011comprehensive} use transitivity-based clustering in biological data, and  \citet{nguyen2007consensus} develop clustering methods that aggregate multiple partitionings via pairwise similarity and voting mechanisms.}. Finally, if a prediction is found to be equivalent to at least a threshold number of other predictions (i.e., at least 5 out of 8), we designate this group as the majority vote and use it as the reference answer. 

\autoref{fig:type_count} shows the answer type counts of the Principia Collection after all the aforementioned procedures. Overall, we filter out 134,172 instances and use the remaining 248,748 instances in the following sections.

\subsection{Experimental Results}

\subsubsection{Experimental Setup}

\noindent\textbf{Baselines.} In our main results (\autoref{subsec:main_results}), we demonstrate the effectiveness of our Principia Collection dataset by training multiple base LMs on the Principia Collection and comparing the trained LMs to (1) existing post-trained LMs that share the same base LMs; and (2) existing post-training datasets by training on the same set of base LMs. The specific baselines we employ from these two groups are listed below:
\begin{itemize}
    \item \textbf{Post-Trained LM Baselines}: We use post-trained LMs that share the same base LMs (Qwen2.5-7B-Base, Qwen3-4B-Base, Qwen3-4B, and OctoThinker-8B-Long-Base) and provide publicly available checkpoints on huggingface as our baselines. These include Qwen2.5-7B-Instruct~\citep{yang2024qwen2}, Qwen3-4B~\citep{yang2025qwen3}, General-Reasoner-4/7B~\citep{ma2025general}, OpenReasoner-Zero~\citep{hu2025open}, SimpleRL-Qwen-2.5-7B-Zoo~\citep{zeng2025simplerl}, and Polaris-4B~\citep{Polaris2025}. Note that we use the available checkpoints without replicating them and employ the hyperparameters specified in each model card.
    \item \textbf{Post-training Dataset Baselines}: We choose two baseline datasets, (1) WebInstruct-Verified~\citep{ma2025general} and (2) DeepScaleR~\citep{deepscaler2025}. WebInstruct-Verified is constructed by extracting problem–answer pairs from pre-training corpora and adeptly transforming them using Gemini-2.5-Pro~\citep{comanici2025gemini}. It serves as our most direct baseline because it also encompasses mathematical objects as part of their dataset. DeepScaleR, on the other hand, collects problems from AMC, AIME, IMO, IMC, and the AoPS forum, thus containing a large proportion of challenging math-competition problems. Notably, both datasets are based on human-written materials, making them well-suited for \textbf{comparing how effective our synthetic data (Principia Collection)} is. We train Qwen3-4B-Base and OctoThinker-8B-Long-Base using these datasets, employing the same  hyperparameters as our setting.
\end{itemize}

\noindent\textbf{Training configuration.} We use Group Relative Policy Optimization (GRPO)~\citep{shao2024deepseekmath}, a widely adopted RL algorithm, to train LMs to generate CoTs using the Principia Collection. As our base models, we employ four LMs: Qwen2.5-7B-Base~\citep{yang2024qwen2}, Qwen3-4B-Base~\citep{yang2025qwen3}, Qwen3-4B (thinking)~\citep{yang2025qwen3}, and OctoThinker-8B-Long-Base~\citep{wang2025octothinker}. These models are selected to cover a range of parameter scales (4B–8B), different model families (Qwen-based and Llama-based), and training configurations that include both zero RL and continual training of reasoning models—all within a computational budget that allows training on the full 250K-instance dataset in a reasonable time frame. 

To match the training compute when training on different datasets, we train up to at most 1K steps (which is equivalent to training on 1 epoch on the Principia Collection under our hyper-parameters) and choose the best performing checkpoint (per 50-step) using the overall validation scores. We use train batch size of 256, max response token length of 8192, PPO mini batch size of 256, clip ratio (low) of 0.2, clip ratio (high) of 0.3, temperature 0.7, top\_p 1.0, 32 response rollouts per prompt, log prob micro batch size of 256, and reference log prob micro batch size of 256. We set the KL coefficient as 0.0 for Qwen3-4B-Base and 0.001 for other base LMs. An exception for this default setting is LMs that is trained upon Qwen3-4B (thinking), where we use train batch size of 128, PPO mini batch size of 128, clip ratio (low) of 0.2, clip ratio (high) of 0.28, top\_p of 1.0. Following \citet{Polaris2025}, we gradually increase the max response token length from 52K to 90K and also increase the temperature from 1.2 to 1.4. All experiments are held on 8 H200 nodes with 64 GPUs.

\noindent\textbf{Evaluation configuration.} For all of the LMs we trained, we use temperature 0.7, top\_p 1.0, and max token length of 16,384 during evaluation. An exception is LMs that are trained from Qwen3-4B (thinking), where we use temperature 1.4, top\_p 1.0, and max token length of 90,000, following the settings of \citet{Polaris2025}.

\noindent\textbf{Validation set.} Unlike prior works that use AIME for validation purposes~\citep{zeng2025simplerl,hu2025open,Polaris2025}, we prepare a held-out validation set to prevent overfitting. The validation set are as follows.

\begin{itemize}
    \item \textbf{Held-out set of the Principia Collection}: We randomly sample 40 problem statements for each answer type (equation, inequality, iterval, set, matrix, piecewise function), which leads to a total of 240 instances.
    \item \textbf{GPQA (non-Diamond)}: We use GPQA questions that are not included in the diamond set. Note that we evaluate on both MCQA (with options provided) and free-form (without options provided) formats.
    \item \textbf{Physics (val set)}: We use the validation set of the Physics benchmark~\citep{feng-etal-2025-physics}. We filter out invalid problem statements following the procedure discussed in Subsection~\ref{subsec:principia_bench}.
    \item \textbf{WebInstruct-Verified (val set)}: We use the validation set of the WebInstruct-Verified training data~\citep{ma2025general}. This data consists of diverse formats such as MCQA, numerical, and mathematical objects.
\end{itemize}

\noindent\textbf{Evaluation benchmarks (Test set).} Our test set consists of three main groups based on the required answer type.
\begin{itemize}
    \item \textbf{Mathematical objects}: We use our proposed PrincipiaBench. Note that the PrincipiaBench and Principia Collection use problem statements from different sources. We report \texttt{mean@8} scores.
    \item \textbf{MCQA}: We use GPQA-Diamond~\citep{rein2024gpqa} and the mathematics and engineering subset of SuperGPQA~\citep{du2025supergpqa}. We report \texttt{mean@16} and \texttt{mean@8} scores for each benchmark, respectively.
    \item \textbf{Numerical}: We use AIME-2024 and AIME-2025. For both benchmarks, we report \texttt{mean@32} scores.
\end{itemize}

\begin{table*}[t!]
\fontsize{12}{14}\selectfont
\centering
\resizebox{\textwidth}{!}{
\begin{tabular}{lccccccccccc} 
    \toprule
    \multicolumn{1}{c}{\multirow{2}{*}{\textbf{Model}}} &
    \multicolumn{1}{c}{\multirow{2}{*}{\textbf{Training Data}}} &
    \multicolumn{5}{c}{\textbf{PrincipiaBench (Math Obj.)}} &
    \multicolumn{2}{c}{\textbf{MCQA}} &
    \multicolumn{2}{c}{\textbf{Numerical}} &
    \multicolumn{1}{c}{\multirow{2}{*}{\textbf{Total Avg.}}} \\
    \cmidrule(lr){3-7}\cmidrule(lr){8-9}\cmidrule(lr){10-11}
    & & \textbf{RealMath} & \textbf{Physics} & \textbf{ARB} & \textbf{SuperGPQA$^{*}$}
      & \textbf{Avg.} & \textbf{SuperGPQA$^{*}$} & \textbf{GPQA-Diamond}
      & \textbf{AIME-24} & \textbf{AIME-25}
      &  \\ 
    \midrule
    \multicolumn{12}{@{}c}{\cellcolor{red!10}\textbf{Qwen2.5-7B-Base}} \\
    \midrule
    Qwen2.5-7B-Base & - & 11.19 & 9.32 & 16.76 & 13.75 & 12.75 & 20.41 & 27.40 & 8.02 & 3.65 & 13.81\\
    Qwen2.5-7B-Instruct & - & 12.95 & 10.45 & 19.15 & 20.05 & 15.65 & 32.61 & 35.29 & 12.08 & 7.81 & 18.80\\
    SimpleRL-7B-Zoo & SimpleZoo-Data. & 14.00 & 10.68 & 26.86 & 21.17 & 18.18 & 28.49 & 28.85 & 15.31 & 8.75 & 19.26\\
    General-Reasoner-7B & WebInstruct-Ver. & 15.96 & 12.39 & 26.86 & 23.15 & 19.59 & 32.95 & 36.55 & 10.63 & 7.71 & 20.78\\
    OpenReasoner-Zero & ORZ-Math-Coll. & 15.09 & 13.75 & 30.85 & 25.12 & 21.20 & 37.10 & 37.25 & \textbf{17.19} & \textbf{13.65} & 23.75\\
    Principia-7B-Zero (\textbf{Ours}) & Principia Collec. & \textbf{15.59} & \textbf{15.11} & \textbf{32.45} & \textbf{28.34} & \textbf{22.87} & \textbf{38.80} & \textbf{39.71} & 15.52 & 10.73 & \textbf{24.53}\\
    \midrule
    \multicolumn{12}{@{}c}{\cellcolor{yellow!10}\textbf{OctoThinker-8B-Long-Base}} \\
    \midrule
    OctoThinker-8B-Long-Base & - & 3.16 & 2.73 & 5.32 & 3.79 & 3.75 & 6.53 & 11.74 & 0.83 & 0.63 & 4.34\\
    (\textit{Reimplemented}) & DeepScaleR & 10.66 & 11.02 & 19.95 & 16.82 & 14.61 & \textbf{34.36} & 29.42 & 13.54 & 19.27 & 19.38\\
    (\textit{Reimplemented}) & WebInstruct-Ver. & 10.66 & 11.02 & 19.95 & \textbf{20.56} & 17.67 & 34.31 & 29.42 & 13.54 & 19.27 & 19.84\\
    Princpia-8B-Zero (\textbf{Ours}) & Principia Collec. & \textbf{13.57} & \textbf{14.20} & \textbf{28.46} & 19.96 & \textbf{19.05} & 33.49 & \textbf{33.62} & \textbf{18.33} & \textbf{20.73} & \textbf{22.80}\\
    \midrule
    \multicolumn{12}{@{}c}{\cellcolor{green!10}\textbf{Qwen3-4B-Base}} \\
    \midrule
    Qwen3-4B-Base & - & 9.43 & 5.80 & 17.81 & 12.18 & 11.31 & 20.46 & 18.09 & 12.5 & 10.21 & 13.31\\
    Qwen3-4B (no-thinking) & - & 17.86 & \textbf{22.39} & 39.89 & 28.93 & 27.27 & 39.78 & 41.89 & 22.81 & 21.04 & 29.32\\
    \textit{(Reimplemented)} & DeepScaleR & 9.24 & 20.91 & 38.30 & 31.04 & 27.42 & 43.38 & \textbf{44.63} & \textbf{23.02} & 21.77 & 29.04\\
    General-Reasoner-4B & WebInstruct-Ver. & 16.06 & 18.07 & 39.36 & 27.88 & 25.34 & 39.63 & 41.04 & 19.06 & 14.27 & 26.92\\
    Principia-4B-Zero (\textbf{Ours}) & Principia Collec. & \textbf{19.28} & 21.81 & \textbf{43.62} & \textbf{33.92} & \textbf{29.66} & \textbf{45.53} & 43.56 & 22.71 & \textbf{23.96} & \textbf{31.80}\\
    \midrule
    \multicolumn{12}{@{}c}{\cellcolor{blue!10}\textbf{Qwen3-4B (thinking)}} \\
    \midrule
    Qwen3-4B (thinking) & - & 23.81 & 40.57 & 58.78 & 41.77 & 41.23 & 58.87 & 53.70 & 71.88 & 64.90 & 51.79\\
    Polaris-4B & Polaris-Data. & 26.17 & 51.02 & 64.36 & 45.82 & 46.84 & 62.18 & 57.30 & 79.50 & 77.19 & 57.94\\
    Principia-4B (\textbf{Ours}) & Principia Collec. & \textbf{28.96} & \textbf{51.24} & \textbf{66.53} & \textbf{47.05} & \textbf{48.45} & \textbf{63.83} & \textbf{57.48} & \textbf{81.06} & \textbf{78.12} & \textbf{59.28}\\
    \midrule
    \multicolumn{12}{@{}c}{\cellcolor{gray!10}\textbf{Reference Models}} \\
    \midrule
    Llama-3.2-3B-Instruct & - & 3.70 & 3.30 & 1.33 & 4.25 & 3.14 & 12.24 & 15.88 & 3.33 & 0.52 & 5.57\\
    Llama-3.1-8B-Instruct & - & 6.01 & 5.80 & 9.31 & 7.40 & 7.13 & 16.18 & 21.46 & 3.13 & 1.15 & 8.81 \\
    Llama-3.3-70B-Instruct & - & 18.41 & 21.36 & 37.50 & 25.81 & 25.77 & 38.79 & 48.61 & 19.27 & 4.38 & 26.77\\
    Qwen3-14B (no-thinking) & - & 21.34 & 28.64 & 50.27 & 36.50 & 34.19 & 47.89 & 53.47 & 29.69 & 26.04 & 36.73\\
    Qwen3-14B (thinking) & - & 28.36 & 51.36 & 67.02 & 49.35 & 49.02 & 66.82 & 65.56 & 78.23 & 70.73 & 59.68 \\
    Qwen3-235B (no-thinking) & - & 28.54 & 44.32 & 63.03 & 45.14 & 45.26 & 55.82 & 62.94 & 37.40 & 27.08 & 45.38\\
    Qwen3-235B (thinking) & - & 36.02 & 58.64 & 73.94 & 53.74 & 55.58 & \textbf{69.33} & 74.34 & 83.33 & 80.94 & 66.29\\
    GPT-OSS-20B & - & 42.56 & 51.59 & 72.34 & 50.71 & 54.30 & 63.71 & 65.07 & 74.72 & 72.71 & 61.68\\
    GPT-OSS-120B & - & 44.05 & 59.89 & 74.47 & 53.92 & 58.08 & 66.35 & 69.31 & 79.31 & 76.15 & 65.43\\
    Claude-4.0-Sonnet & - & 32.04 & 41.82 & 59.57 & 48.19 & 45.40 & 63.37 &70.71 & 57.19 & 44.58&52.18\\
    GPT-4.1 & - & 36.30 & 51.25 & 61.44 & 46.43 & 48.85 & 60.78 & 67.20 & 46.98 & 35.71 & 50.76\\
    o3 & - & \textbf{48.74} & \textbf{63.75} & \textbf{81.91} & \textbf{57.19} & \textbf{62.90} & 69.10 & \textbf{82.29} & \textbf{85.63} & \textbf{80.99} & \textbf{71.20}\\
    \bottomrule
\end{tabular}}
\caption{\textbf{Evaluation performance on PrincipiaBench involving hard reasoning problems with mathematical objects}. Principia models are trained with new synthetic reasoning problems involving mathematical objects (Principia Collection) – which gives improvements both on PrincipiaBench and existing MCQA and numerical benchmarks.  Note that for SuperGPQA$^{*}$, we use problems that require the generation of mathematical objects from the mathematics and engineering subset. LMs are grouped based on the base model it was trained on, and the best scores from each groups are \textbf{bolded}. Note that \texttt{Reimplemented} denotes the baselines we trained ourselves with either DeepScaleR or WebInstruct-Verified datasets, since there are no publicly available checkpoints for some of the base LMs.}
\label{table:main_results}
\end{table*}  

\subsubsection{Main Results}\label{subsec:main_results}

The main results are shown in \autoref{table:main_results}.

\noindent\textbf{Training on Principia Collection improves the ability to derive mathematical objects.} Across all four base models-each colored in  red \tightcolorbox{red!10}{(Qwen2.5-7B-Base)}, yellow \tightcolorbox{yellow!10}{(OctoThinker-8B-Long-Base)}, green \tightcolorbox{green!10}{(Qwen3-4B-Base)}, and blue \tightcolorbox{blue!10}{(Qwen3-4B)}---we observe a consistent trend where training on the Principia Collection leads to performance improvements on the PrincipiaBench. Specifically, Qwen2.5-7B-Base improves +10.12\% (12.75 $\rightarrow$ 22.87), OctoThinker-8B-Long-Base improves +15.30\% (3.75 $\rightarrow$ 19.05), Qwen3-4B-Base improves +18.35\% (11.31 $\rightarrow$ 29.66), and Qwen3-4B improves +7.22\% (41.23 $\rightarrow$ 48.45). The resulting models (\tightcolorbox{red!10}{Principia-7B-Zero}, \tightcolorbox{yellow!10}{Principia-8B-Zero}, \tightcolorbox{green!10}{Principia-4B-Zero}, and \tightcolorbox{blue!10}{Principia-4B}) achieve the highest  scores on PrincipiaBench across all settings, implying that Principia Collection strengthens an LM’s ability to derive mathematical objects.

\noindent\textbf{Training on mathematical objects exhibits positive transfer to MCQA and numerical benchmarks.}
Unexpectedly, we also observe substantial improvements on MCQA and numerical benchmarks when training on the Principia Collection (which only consists of mathematical object answers). For instance, \tightcolorbox{blue!10}{Principia-4B} outperforms Polaris-4B and markedly surpasses the Qwen3-4B (thinking) model on SuperGPQA, GPQA-Diamond, and AIME. Likewise, \tightcolorbox{red!10}{Principia-7B-Zero}, \tightcolorbox{yellow!10}{Principia-8B-Zero}, and \tightcolorbox{green!10}{Principia-4B-Zero} achieve the best average performances, including PrincipiaBench, MCQA and numerical benchmarks. These results demonstrate that learning to construct complex mathematical forms strengthens reasoning abilities that generalize beyond mathematical object derivation (PrincipiaBench) to diverse answer formats (AIME, GPQA, SuperGPQA).

\begin{figure}[t]
    \centering    \includegraphics[width=1\linewidth]{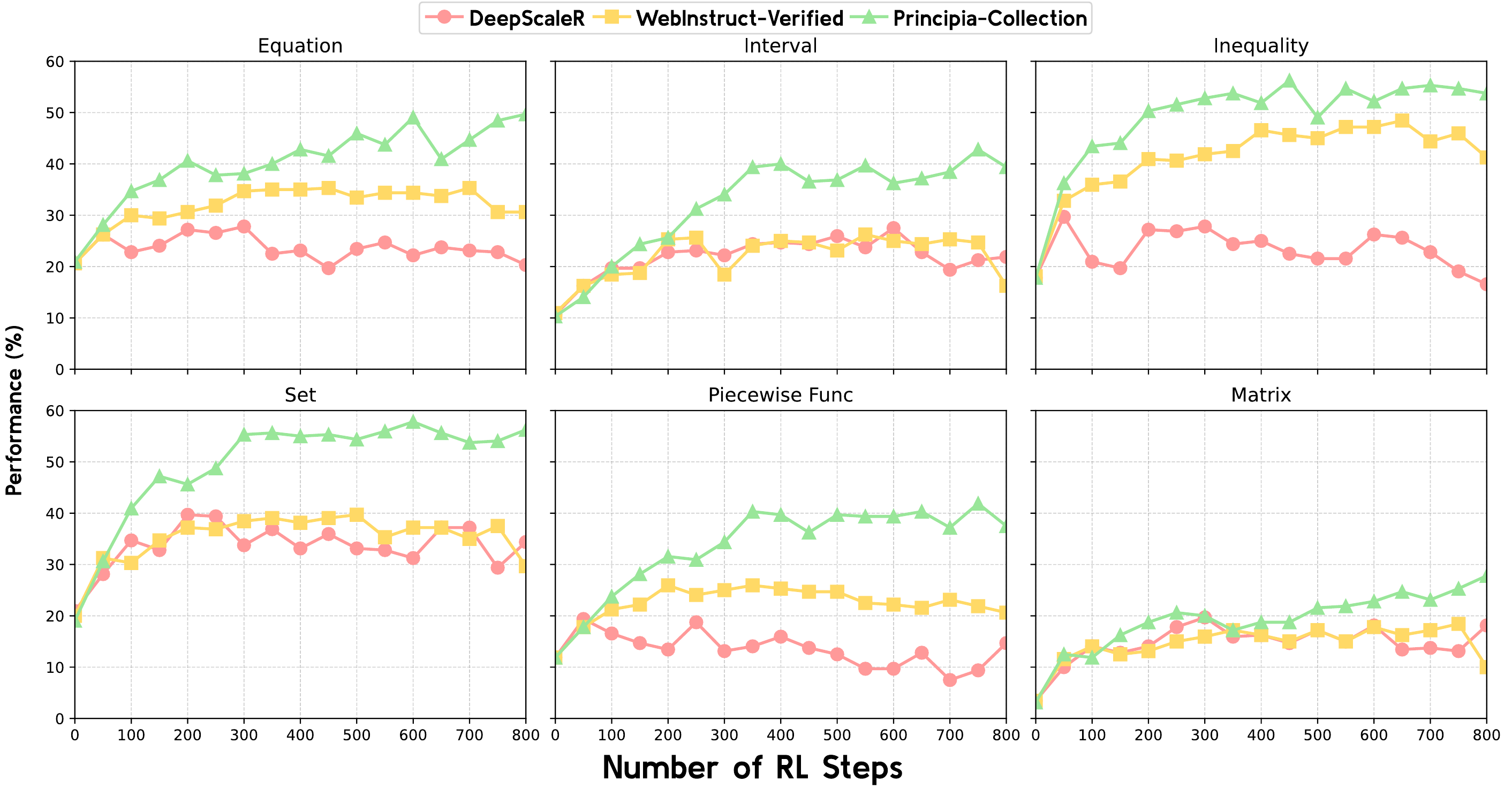}
    
    \caption{\textbf{Training directly on complex mathematical objects yields substantially better transfer than training on datasets requiring only numerical values or simple mathematical objects.} Validation performance across different answer types when training with three RL post-training datasets. DeepScaleR and WebInstruct-Verified contain problems from reliable online sources but mostly require simple mathematical objects or numerical answers. In contrast, Principia Collection is a fully synthetic dataset built around diverse and complex mathematical objects. Across all categories, Principia-trained models exhibit consistently higher and more stable improvements over training steps, demonstrating that exposure to complex answer structures during learning is crucial for generalizing to tasks involving rich mathematical forms.}
    \label{fig:validation_scores}
\end{figure}

\noindent\textbf{Frontier models struggle on PrincipiaBench.}
Compared to widely used reasoning benchmarks such as SuperGPQA, GPQA-Diamond, and AIME, the vast majority of LMs—including frontier models like o3 (62.90) and Qwen3-235B (55.58)—exhibit substantially lower performance on the PrincipiaBench. This highlights that even the strongest models struggle to derive mathematical objects. Consequently, PrincipiaBench serves as a challenging and informative testbed for evaluating the depth of reasoning capabilities in frontier LMs.

\begin{wrapfigure}{r}{0.43\textwidth}
    \centering
    \vspace{-6mm}
    \includegraphics[width=\linewidth]{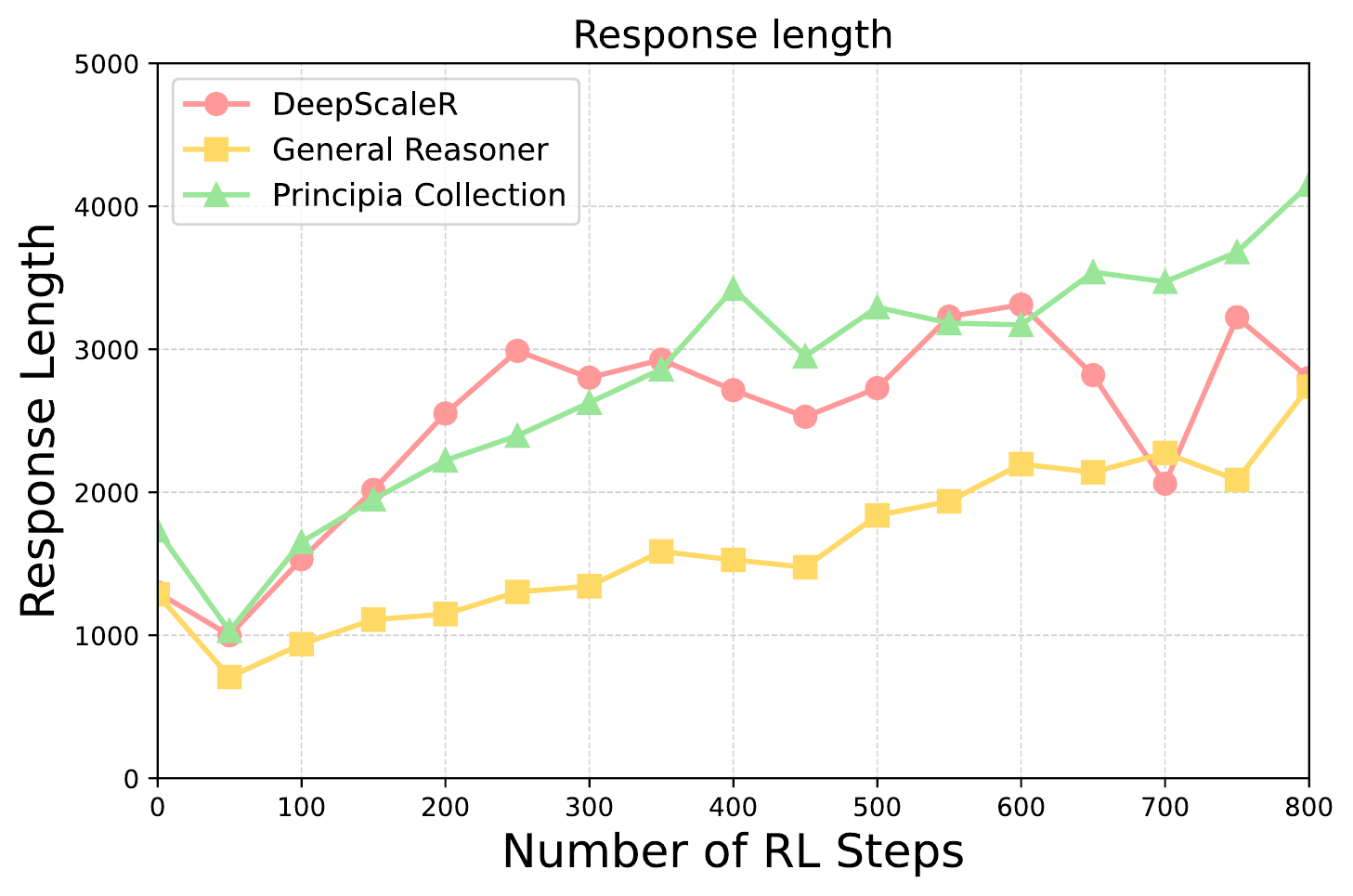}
\caption{\small{Response length gradually increases when training on the Principia Collection. Note that the token length is measured on our validation set.}}
    \vspace{-3mm}
\label{fig:validation_length_comparison}
\end{wrapfigure}

\noindent\textbf{Training directly on complex mathematical objects provides far stronger transfer than relying on datasets limited to simple objects or numerical answers.} As shown in \autoref{fig:validation_scores}, we compare the scores on our validation set when training Qwen3-4B-Base on different RL post-training. This experimental result addresses the question: ``\textit{Do we really need to train on data whose answers are complex mathematical objects, or will a model naturally generalize if it is trained only on problems with simple answers such as numerical values?}'' The results show that either training on DeepScaleR (consisting of problems from AMC, AIME, IMO, IMC, and the AoPS forum with numerical value answers) and WebInstruct-Verified (consisting of simple mathematical objects and numerical value answers) are insufficient to achieve this goal. In contrast, the Principia Collection is better aligned with the desired target (\textit{i.e.}, precisely deriving complex mathematical objects) and we observe that it leads to more consistent improvements over the course of training. Hence, this implies that practitioners who want their LMs to precisely derive mathematical objects can include the Principia Collection in their training suite. Furthermore, as a side effect, in \autoref{fig:validation_length_comparison}, training on the Principia Collection leads to a more stable increase in the LM’s response length than baselines over the course of training.

\begin{table}[t]
\centering
\fontsize{7}{9}\selectfont
\begin{tabular}{l l c c c c c}
\toprule
\multicolumn{1}{c}{\multirow{2}{*}{\textbf{Verifier Type}}} & \multicolumn{1}{c}{\multirow{2}{*}{\textbf{Verifier}}} & \multicolumn{1}{c}{\multirow{2}{*}{\textbf{Active Parameters}}} & \multicolumn{4}{c}{\textbf{Principia VerifyBench}}\\ \cmidrule(lr){4-7} & & & \textbf{Agreement with Humans (\%)} & \textbf{Precision} & \textbf{Recall} & \textbf{F1}\\
\midrule
Rule-based Verifier & math-verify &  - & 5.95\% & 5.38 & 6.67 & 5.96\\
\midrule
Fine-tuned Verifier & general-verifier & 1.5B & 82.74\% & 83.13 & 93.24 & 87.90 \\
Fine-tuned Verifier & CompassVerifier & 3B & 81.55\% & 90.74 & 65.33 & 75.97 \\
Fine-tuned Verifier & CompassVerifier & 7B & 88.69\% & 93.75 & 80.00 & 86.33 \\
Fine-tuned Verifier & CompassVerifier & 32B & 91.66\% & 94.20 & 86.67 & 90.28 \\
\midrule
General-purpose LM & Qwen3-4B & 4B & 92.26\% & 89.74 & 93.33 & 91.50 \\
General-purpose LM & Qwen3-14B & 14B & 93.45\% & 92.21 & 94.67 & 93.42 \\
General-purpose LM & GPT-OSS-20B & 3.6B & 94.64\% & 95.83 & 92.00 & 93.88 \\
General-purpose LM & GPT-OSS-120B &  5.1B & 95.24\% & 97.18 & 92.00 & 94.52\\
General-purpose LM & o3 & - & 94.05\% & 93.33 & 93.33 & 93.33 \\
\bottomrule
\end{tabular}
\caption{\footnotesize \textbf{Principia VerifyBench results}. We find that model-based verifiers are more effective at determining answer equivalence than rule-based verifiers. Also, o3 and GPT-OSS-120B achieve high agreement rate with ground truth labels, justifying our selection in employing a strong judge in PrincipiaBench and a reward model for RL training with the Principia Collection.}
\label{table:human_judgment}
\end{table}

\subsection{Analysis}
\label{sec:ablations}

\subsubsection{Benefits of model-based verifiers over rule-based verifiers} 

At the core of our recipe, we use a model-based verifier, both for training on Principia Collection and evaluating on PrincipiaBench. To examine its reliability over rule-based verifiers, we check the performance of rule-based verifiers and model-based verifiers on the Principia VerifyBench. In addition, we run ablation experiments by training LMs with rewards from rule/model-based verifiers, and comparing their downstream performance.

\noindent\textbf{Principia VerifyBench Experimental Setting.}
Using the 168 instances in Principia VerifyBench, we compare \texttt{math-verify}~\citep{math_verify} and 13 different model-based verifiers. The model-based verifiers could be categorized into: (1) \textbf{fine-tuned verifiers} which have been trained for checking equivalence between the model prediction and the reference answer and (2) \textbf{general-purpose LMs} which aren't specially trained but could serve as a verifier when prompted. For fine-tuned verifiers, we use general-verifier (1.5B)~\citep{ma2025general} and compassverifier~\citep{liu2025compassverifier} as baselines. For general-purpose LMs, we employ Qwen3-4/14B~\citep{yang2025qwen3}, GPT-OSS-20/120B~\citep{agarwal2025gpt}, and o3~\citep{o3_openai}. We measure the agreement with ground-truth labels, the precision, the recall, and F1 scores for each baseline.

\noindent\textbf{Principia VerifyBench Experimental Result.} The results are shown in \autoref{table:human_judgment}. First, we find that o3 scores 94.05\% agreement with ground truth labels, which justifies our choice behind using o3 as a judge in PrincipiaBench. Furthermore, GPT-OSS-120B scores 95.24\% agreement, which indicates that it could serve as a reliable verifier to provide accurate reward during RL training. Notably, we find that math-verify fails to adequately determine equivalence in most of the cases in Principia VerifyBench, only scoring 5.95\% agreement with the ground-truth labels. Note that Principia VerifyBench was constructed in an adversarial manner for math-verify, since we specifically sample instances where the o3 judge and math-verify disagree. Hence, the experimental results here imply that, in cases where the two judgments diverge, o3 is more likely to be correct. 

Furthermore, we find that specially fine-tuned verifiers (general-verifier, compassverifier) exhibit lower agreement rates (<90\%), primarily due to low recall, whereas even small general-purpose LMs such as Qwen3-4B achieve agreement rates as high as 92.26\%. These findings suggest that, for RL training on the Principia Collection, it is better to use a general-purpose LM as the verifier rather than a fine-tuned verifiers.

\begin{table*}[t!]
\fontsize{12}{14}\selectfont
\centering
\resizebox{\textwidth}{!}{
\begin{tabular}{lllcccccccccc} 
    \toprule
    \multicolumn{1}{c}{\multirow{2}{*}{\textbf{Data}}} &
    \multicolumn{1}{c}{\multirow{2}{*}{\textbf{Verifier Type}}} &
    \multicolumn{1}{c}{\multirow{2}{*}{\textbf{Verifier}}} &
    \multicolumn{5}{c}{\textbf{Math Obj. (PrincipiaBench)}} &
    \multicolumn{2}{c}{\textbf{MCQA}} &
    \multicolumn{2}{c}{\textbf{Numerical}} &
    \multicolumn{1}{c}{\multirow{2}{*}{\textbf{Total Avg.}}} \\
    \cmidrule(lr){4-8}\cmidrule(lr){9-10}\cmidrule(lr){11-12}
     & & & \textbf{RealMath} & \textbf{Physics} & \textbf{ARB} & \textbf{SuperGPQA$^{*}$}
      & \textbf{Avg.} & \textbf{SuperGPQA$^{*}$} & \textbf{GPQA-Diamond}
      & \textbf{AIME-24} & \textbf{AIME-25}
      &  \\ 
    \midrule
    - & - & - & 9.43 & 5.80 & 17.81 & 12.18 & 11.31 & 20.46 & 18.09 & 12.5 & 10.21 & 13.31\\
    \midrule
    DeepScaleR & Rule & math-verify&9.24&20.91&38.3&28.10&24.14&43.38&44.63&23.02&21.77&28.67\\
    DeepScaleR & Model & general-verifier&16.20&18.86&38.03&30.13&25.81&42.35&45.74&\textbf{25.00}&23.13&29.93\\
    DeepScaleR & Model & GPT-OSS-120B&15.55&19.93&39.16&31.04&27.42&45.32&\textbf{46.14}&24.17&22.29&30.45\\
    \midrule
    Principia-Collec. & Rule & math-verify&11.53&10.80&21.54&15.67&14.89&29.06&32.29&5.52&3.96&16.30\\
    Principia-Collec. & Model & general-verifier&10.48&8.18&7.98&6.72&8.34&10.27&31.09&0.42&1.15&9.54\\
    Principia-Collec. & Model & GPT-OSS-120B&\textbf{19.28} & \textbf{21.81} & \textbf{43.62} & \textbf{33.92} & \textbf{29.66} & \textbf{45.53} & 43.56 & 22.71 & \textbf{23.96} & \textbf{31.80}\\
    \bottomrule
\end{tabular}}
\caption{\footnotesize \textbf{Training with rule-based verifier vs. model-based verifier}: Using a strong model-based verifier becomes highly important when the data is more challenging to verify. While there is little difference in terms of performance when training on DeepScaleR, which consists of numerical value and simple mathematical object answers, the performance varies a lot when training on the Principia Collection with consists of complex mathematical object answers. The best performances across all baselines are \textbf{bolded}.}
\label{table:ablation_verifier}
\end{table*}

\noindent\textbf{Reward Model Ablation Experimental Setting.} Furthermore, we compare the effectiveness of a rule-based verifier and a model-based verifier by measuring the downstream performance of LMs RL trained with each verifier. In this experiment, we train Qwen3-4B-Base on either DeepScaleR or the Principia Collection using three different verifiers: math-verify, general-verifier, and GPT-OSS-120B. This results in six total training settings. 

\noindent\textbf{Reward Model Ablation Experimental Result.}
The results are shown in \autoref{table:ablation_verifier}. Notably, the performance gap across different verifier types is relatively small when training on DeepScaleR, where the total average ranges only from 28.67 to 30.45. This is likely because DeepScaleR contains many numerical-value answers sourced from math competitions (IMO, IMC, AMC, AIME), and thus using either rule-based or model-based verifiers does not make a substantial difference.\footnote{As also shown in \autoref{fig:token_count}, most answers are under 10 tokens, making the verification task comparatively simple.} In contrast, the results vary substantially when training on the Principia Collection. Performance increases only modestly with \texttt{math-verify} (13.31 $\rightarrow$ 16.30) and even decreases with \texttt{general-verifier} (13.31 $\rightarrow$ 9.54), whereas using a strong model-based verifier yields the largest gain (13.31 $\rightarrow$ 31.80). We speculate that this difference arises from the effectiveness of GPT-OSS-120B in determining the equivalence of mathematical objects. Notably, although \texttt{general-verifier} achieves an 82.74\% agreement rate on Principia VerifyBench, the LM RL-trained with \texttt{general-verifier} attains very low performance (9.54). This suggests that even a slightly imperfect verifier can significantly deteriorate RL training outcomes. Taken together, these findings indicate that employing a strong LM-based verifier is crucial when post-training on datasets whose answers consist of complex mathematical objects (Principia Collection).

\begin{figure}[t!]
    \centering    \includegraphics[width=1\linewidth]{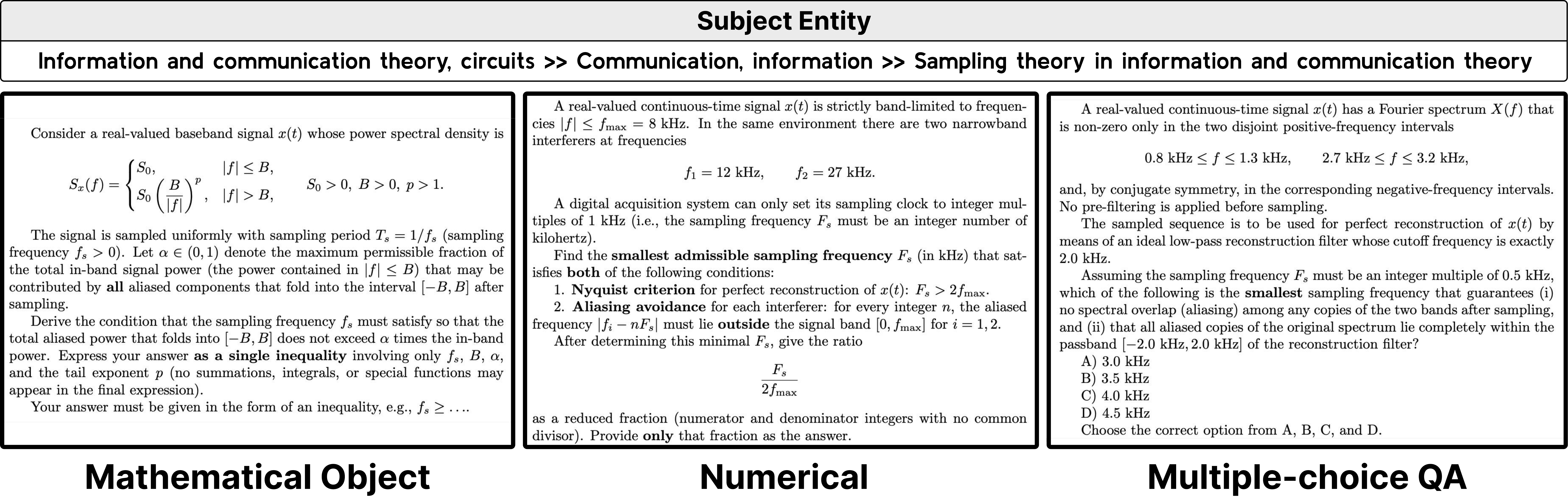}
    
    \caption{\textbf{Example of training instances used in data ablation experiments.} We prepare training data that share the same subject entity (``\textit{Information and communication theory, circuits >> Communication, information >> Sampling theory in information and communication theory}'') but requires either a mathematical object answer (our Principia Collection data), numerical answers, and multiple choice options. By training on these distinctly formatted datasets (\autoref{section:data_ablation}), we find that training to generate mathematical objects is effective at improving reasoning capabilities.}
    \label{fig:distinct_formats}
\end{figure}

\subsubsection{Unifying different answer formats}\label{section:data_ablation}

\noindent\textbf{Experimental Setting.} Frontier LMs are often post-trained on heterogeneous datasets, which raises a natural question: ``When post-training data spans multiple answer formats (mathematical objects, numerical, MCQA), how does this mixture affect performance on benchmarks associated with each format, and does mixing these formats provide any benefit?'' To answer this question, we systematically analyze how including mathematical-object data affects downstream performance when combined with other data types. 

\begin{table*}[t!]
\fontsize{12}{14}\selectfont
\centering
\resizebox{\textwidth}{!}{
\begin{tabular}{ccccccccccccccc} 
    \toprule
    \multicolumn{3}{c}{\textbf{Training Data Answer Format}} &
    \multicolumn{1}{c}{\multirow{2}{*}{\textbf{Method}}} &
    \multicolumn{5}{c}{\textbf{Math Obj. (PrincipiaBench)}} &
    \multicolumn{2}{c}{\textbf{MCQA}} &
    \multicolumn{2}{c}{\textbf{Numerical}} &
    \multicolumn{1}{c}{\multirow{2}{*}{\textbf{Total Avg.}}} \\
    \cmidrule(lr){1-3}\cmidrule(lr){5-9}\cmidrule(lr){10-11}\cmidrule(lr){12-13}
    \textbf{Math Obj.}& \textbf{Numer.} & \textbf{MCQA.} & & \textbf{RealMath} & \textbf{Physics} & \textbf{ARB} & \textbf{SuperGPQA$^{*}$}
      & \textbf{Avg.} & \textbf{SuperGPQA$^{*}$} & \textbf{GPQA-Diamond}
      & \textbf{AIME-24} & \textbf{AIME-25}
      &  \\ 
    \midrule
    \xmark & \xmark & \xmark & - & 9.43 & 5.80 & 17.81 & 12.18 & 11.31 & 20.46 & 18.09 & 12.5 & 10.21 & 13.31\\
    \midrule
    \cmark & \xmark & \xmark & Single Training & 19.28 & 21.81 & 43.62 & 33.92 & 29.66 & 45.53 & 43.56 & 22.71 & 23.96 & 31.80\\
    \xmark & \cmark & \xmark & Single Training & 17.96 & 20.14 &40.96&	30.28&27.34&46.10&43.50&\textbf{26.98}&23.75&31.19\\
    \xmark & \xmark & \cmark & Single Training&15.84&18.75&37.23&27.87&24.96&34.74&29.29&17.71&19.79&25.15\\
    \midrule
    \cmark & \cmark & \xmark & Joint Training & 17.46&19.89&40.96&32.18&27.62&43.75&42.27&22.50&20.31&29.92\\
    \cmark & \cmark & \xmark & Weight Merging & \textbf{20.35}&\textbf{22.93}&\textbf{45.21}&\textbf{34.38}&\textbf{29.97}&\textbf{47.30}&\textbf{44.79}&25.94&\textbf{24.92}&\textbf{33.23}\\
    \midrule
    \cmark & \cmark & \cmark & Joint Training&16.83&19.66&39.63&30.98&26.77&38.80&40.28&20.42&20.52&28.39\\
    \cmark & \cmark & \cmark & Weight Merging&17.68&22.84&44.68&33.59&29.70&46.02&41.95&26.04&22.81&31.95\\
    \bottomrule
\end{tabular}}
\caption{\footnotesize  \textbf{Unifying mathematical objects and numerical type data using weight merging leads to the best performances.} We systematically generate three different types of synthetic data that shares the same subject entities but require different format of answers. Then, we either train on one of them, jointly train them, or merge the weights of single trained checkpoints. The best scores are \textbf{bolded}.}
\label{table:ablation_data}
\end{table*}  

As shown in \autoref{fig:distinct_formats}, we generate additional synthetic datasets that use the same underlying entity subjects as the Principia Collection but require either a numerical value answer or a multiple-choice option answer. The three problem statements share the same entity subject but require different answer formats. This enables to minimize the effect of problem diversity (which is crucial in practice) and solely compare the effect of the answer format~\citep{kim2025evaluating}. We follow the same data-generation procedure (described in Subsection~\ref{subsec:principia_collection}), and the resulting numerical-value dataset contains 300K questions, and the MCQA dataset contains 350K questions, where fewer samples are filtered out compared to mathematical objects. For numerical-value answers, we categorize problems into six types—integer values with or without units, fractional values with or without units, and decimal values with or without units. For multiple-choice questions, we include three formats: questions with 4 options, 6 options, and 10 options. Then, we train Qwen3-4B-Base on these three different datasets and compare their performances on our evaluation suite. 

When mixing training datasets, we compare two approaches: (1) \textbf{joint-training}, where we mix the datasets and RL train them together, and (2) \textbf{weight-merging}, where we train LMs on each datasets separately, and them merge the resulting checkpoints. For weight merging, in practice, the performance might vary a lot depending on how we mix the checkpoints. For simplicity, we use linear merging (1:1 or 1:1:1 ratio).

\noindent\textbf{Experimental Results.} The results are shown in \autoref{table:ablation_data}. Our observation is that training solely on MCQA data performs poorly across nearly all benchmarks. For example, the MCQA-only model reaches only 25.04 total average performance, with weak results on both mathematical-object tasks (e.g., 24.96 on PrincipiaBench Avg.) and numerical benchmarks (17.71 on AIME-24 and 19.79 on AIME-25). This suggests that MCQA supervision alone provides little transferable signal for complex reasoning. As illustrated in \autoref{fig:cot_comparison}, a likely cause of this failure is that LMs often rely on the provided options as an anchor during their reasoning process. In an MCQA-only training setup, the model is repeatedly rewarded for behaviors that exploit this shortcut—such as backward reasoning from the answer choices or forming conclusions conditioned on option structure—because these strategies can still lead to correct predictions. We suspect that over the course of training, such option-anchored heuristics become reinforced and amplified, preventing the model from learning the deeper derivational reasoning patterns needed to generalize to mathematical-object or numerical tasks.

In contrast, training on mathematical-object data yields strong transfer to both MCQA and numerical benchmarks. The math-only model achieves 45.53 on SuperGPQA, 43.56 on GPQA-Diamond, and robust numerical performance such as 22.71 on AIME-24 and 23.96 on AIME-25, demonstrating that mathematical-object supervision teaches reasoning skills that generalize well across answer formats.

Most surprisingly, when unifying data types, \textit{weight merging} consistently outperforms joint training. For example, when combining math-object and numerical data, weight merging achieves a total average of 33.23—substantially higher than the joint-training result of 29.92. A similar pattern holds when unifying all three answer formats, where weight merging again surpasses joint training (31.95 vs.\ 28.39). The strongest PrincipiaBench scores in these settings also come from weight merging, including 22.93 on Physics, 45.21 on ARB, and 34.38 on SuperGPQA*). Notably, for both joint training and weight merging, incorporating MCQA data consistently reduces performance compared to using only mathematical-object and numerical-value data. These results suggest that merging independently trained LMs may be more effective than co-training a single model on heterogeneous answer formats, and that MCQA supervision can dilute the benefits of mixing data.

\subsection{Related Work}

\noindent\textbf{Evaluating reasoning capability of LMs.} Reasoning has long been considered a cornerstone of machine intelligence and remains one of the most important capabilities to assess~\citep{bottou2014machine,weston2015towards,chollet2019measure}. In evaluating the reasoning capabilities of LMs, it has become common in recent years to use grade-school mathematics exams or mathematics Olympiad competition problems, which typically require a numerical answer~\citep{cobbe2021training,hendrycks2021measuring,luong2025towards}. Recent efforts also assess reasoning across a broader range of disciplines through multiple-choice questions in subjects such as physics, chemistry, and biology—tasks that require both advanced university- or graduate-level knowledge and the ability to synthesize the information provided in the problem statement to infer the answer~\citep{wang2024mmlu,rein2024gpqa,du2025supergpqa}.

On the other hand, there is considerable excitement that LMs could be used in reasoning-intensive real-world applications such as autonomous scientific discovery and assisting scientists’ workflows~\citep{van2023ai,mondorf2024beyond,lu2024ai,wang2025ai,gpt5_accelerating_science}. Yet a substantial gap remains between these ambitions and the standardized tests we currently rely on. One of the most important capabilities overlooked in the literature between exam-style standardized tests and scientific discovery is the ability to manipulate and infer complex mathematical objects. Problems that require producing or reasoning over mathematical objects present a more challenging test bed than multiple-choice or numerical-answer standardized exams, enable the evaluation of a broader range of reasoning skills, and simultaneously assess a critical competency needed for conducting experiments or writing scientific papers~\citep{hestenes2009modeling,meerschaert2013mathematical,goos2023mathematics}. It is also worth noting that \citet{ma2025general}, whose motivation is closely aligned with ours, trains General-Reasoner on diverse domains and formats, yet evaluates their model only on multiple-choice questions and olympiad problems that require numerical answers. In contrast, our work introduces PrincipiaBench to assess an LM’s ability to derive mathematical objects.

\noindent\textbf{Synthetic data for improving reasoning capability of LMs.} Synthetic data generation has been widely used to improve the instruction-following capabilities of LMs, both by creating new prompts and by generating corresponding responses~\citep{wang2023self,alpaca,liu2025spice}. In contrast, when acquiring data aimed at enhancing reasoning—such as mathematical problem-solving—the highest priority is placed on ensuring answer correctness. As a result, instead of generating entirely new problems, prior work has typically augmented chain-of-thoughts (CoTs) to existing problem–answer pairs~\citep{zelikman2022star,kim2023cot,mukherjee2023orca}\footnote{When training LMs with SFT or DPO, CoTs that correctly predict the ground-truth answer are filtered beforehand; in RL settings, however, such CoT are not pre-collected but are instead rewarded when the policy model generates a correct answer.}. However, while this approach has the advantage of guaranteeing answer correctness, it also suffers from a key limitation: it restricts the diversity of problem sources to only those that already exist. Notably, commonly used post-training datasets—such as DeepScaleR~\citep{deepscaler2025}, OmniMath~\citep{gao2025omnimath}, NuminaMath~\citep{numina_math_datasets}, and Skywork-OR1~\citep{he2025skywork}—collect problems from reliable sources including AMC, AIME, IMO, IMC, and Chinese K-12 exams. Yet none of these datasets contain a lot of the types of problems that require deriving complex mathematical objects. 

Another widely used data-synthesis approach extracts problems from online forums or documents in the pre-training corpus and then slightly modifies the problem statement while keeping the original answer unchanged. However, the diversity of the resulting data is still determined by the types of problems contained in those documents. Representative datasets produced through this approach include Natural Reasoning~\citep{yuan2025naturalreasoning}, WebInstruct-Verified~\citep{ma2025general}, and TextBookReasoning~\citep{fan2025megascience}. Yet, when we analyze the token distribution of their outputs, we find that—despite the potential of document-based data synthesis—these datasets contain far too few instances that require the complex mathematical objects as well. In our work, we generate new problem statements using subjects drawn from graduate-level disciplines (\textit{i.e.}, entities from MSC2020 and PhySH) that require mathematical objects as answers. Furthermore, we demonstrate that training on the Principia Collection yields strong performance on both the PrincipiaBench and other reasoning benchmarks.

\subsection{Conclusion}
In this work, we propose \textbf{PrincipiaBench}, a benchmark that explicitly requires mathematical-object derivation; the \textbf{Principia Collection}, an RL post-training dataset designed to teach this ability; and \textbf{Principia VerifyBench}, a meta-evaluation suite for assessing the performance of verifiers used during RL training and benchmarking. While the ability to derive mathematical objects is fundamental to scientific reasoning, existing benchmarks that rely on multiple-choice formats fail to capture this capability. Our main finding is that LMs trained on the Principia Collection not only improve substantially on PrincipiaBench but they also exhibit enhanced reasoning on other answer formats, including MCQA and numerical tasks. Furthermore, through a range of experiments, we show that a model-based verifier is necessary for providing accurate rewards when training on Principia Collection.

Looking ahead, we hope the artifacts introduced in this paper will be broadly useful to the community. PrincipiaBench can serve as a rigorous benchmark for researchers developing new RL algorithms or post-training datasets to track improvements in reasoning. The Principia Collection and its data generation pipeline can be utilized for future work on synthetic data synthesis. Finally, Principia VerifyBench provides a principled way to train and evaluate model-based verifiers for fast RL training. We hope that the ability to derive mathematical objects is no longer overlooked and is instead recognized as a core capability of LMs.

\clearpage
\newpage

\section{RLLM: Unified Post-Training via On-Policy-Trained Language Model as a Reward Model} \label{sec:rllm}

\begin{figure*}[h]
    \centering
    \includegraphics[width=\textwidth]{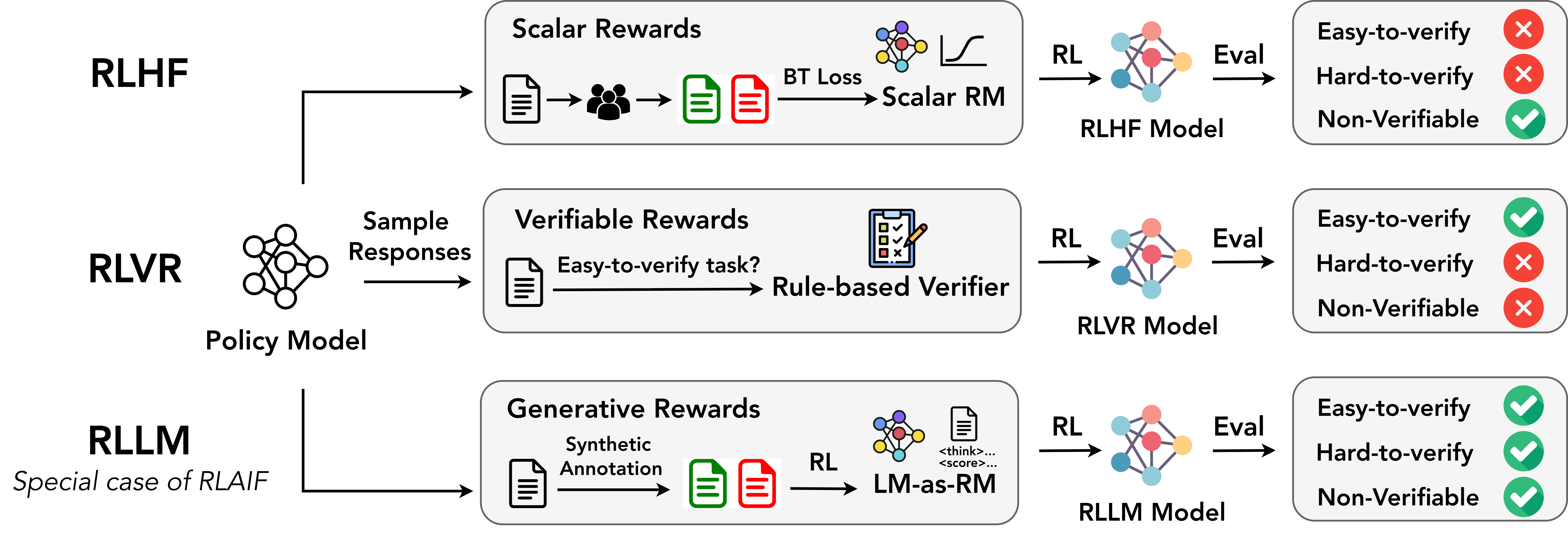}
    \caption{ \label{fig:intro} {\bf Reinforcement Learning with an LM as Reward
Model (RLLM).}  Comparison of RLHF, RLVR, and RLLM for post-training LLMs. RLHF optimizes a policy against a scalar reward model trained on human-annotated preference data using the Bradley-Terry objective. RLVR trains a policy using a rule-based verifier and hence is restricted to easy-to-verify tasks with ground-truth labels. RLLM (ours) is a special case of RL-from-AI-Feedback (RLAIF) that first trains an LM-as-RM on synthetic judgments using RL and then uses its generative rewards to optimize the policy. An LM-as-RM exploits an LLM's (1) reasoning capabilities to produce higher-quality reward signals and (2) instruction-following capabilities to allow flexible reward design. Thus, RLLM unifies the post-training paradigm, enabling the policy model to excel across easy-to-verify, hard-to-verify, and non-verifiable tasks. {\em We further show that training the RLLM reward model on-policy (via responses sampled from the policy model) improves results.}}
\end{figure*}

\subsection{Overview}

Large Language Models (LLMs) have demonstrated remarkable performance across a wide variety of tasks, from general instruction following to complex reasoning \citep{guo2025deepseek, yang2025qwen3, hurst2024gpt, dubey2024llama}. A key factor behind this progress is the post-training stage that relies on Reinforcement Learning (RL) to align models to human preferences. Traditionally, this alignment has been achieved through Reinforcement Learning from Human Feedback (RLHF), where human preference data is used to train a scalar reward model that assigns single numerical scores to guide optimization~\citep{ouyang2022training}. However, scalar reward models are limiting: they produce a final score without generating intermediate reasoning steps, making them mostly ineffective for improving reasoning capabilities and susceptible to issues such as reward hacking and poor generalization to out-of-distribution (OOD) scenarios~\citep{gao2023scaling}.

More recently, with the advancement of o1~\citep{jaech2024openai} and R1-style thinking LLMs~\citep{guo2025deepseek}, a new paradigm has emerged: Reinforcement Learning with Verifiable Rewards (RLVR)~\citep{lambert2024tulu}. As illustrated in~\autoref{fig:intro}, RLVR replaces the scalar reward model with rule-based verifiers (e.g., code compilers or math checkers) to provide grounded feedback. While RLVR offers verifiable correctness, it strictly requires access to ``gold'' answers that are often hard-to-obtain or hard-to-verify. This limits its utility primarily to domains with objective correctness criteria—such as simple math and coding—leaving a gap for complex reasoning tasks where answers are not easily checked by rules, or subjective tasks where no gold answer exists~\citep{tao2025hybrid}.

These limitations motivate a third paradigm: RLLM (Reinforcement Learning with an LM itself as a Reward Model). Unlike scalar models that lack deeper reasoning, or rule-based verifiers that lack flexibility, RLLM leverages the inherent capabilities of the LLM itself to serve as a ``thinking'' Reward Model. The effectiveness of this approach stems from an LLM's two core strengths: the model's \textit{reasoning capabilities} enable it to generate explicit thinking traces for higher-quality judgments, while its \textit{instruction-following capabilities} allow for flexible reward design through natural language prompts. Consequently, RLLM serves as a single, unified post-training framework effective across easy-to-verify, hard-to-verify, and non-verifiable domains. Notably, recent rubric-based evaluation methods -- where structured criteria guide judgment -- can be viewed as a special case within our framework: the LM-as-RM implicitly internalizes and flexibly applies such rubrics through its reasoning, without requiring explicitly specified scoring rules.

Our proposed method operates in two distinct stages. First, in the \emph{Thinking LM-as-RM Training} stage, we train an LLM to act as a thinking Reward Model. Crucially, we employ an on-policy training recipe: the RM is trained to evaluate responses sampled specifically from the policy model, using synthetic labels derived from a stronger teacher or verifier. Unlike scalar RMs, this LM-as-RM generates an explicit reasoning trace to justify its judgment before assigning a score. Second, in the \emph{Policy Post-Training} stage, we use the generative rewards produced by our trained LM-as-RM to optimize the policy model via reinforcement learning.

\begin{figure*}[t]
    \centering
    \begin{subfigure}[b]{0.58\textwidth}
        \centering
        \includegraphics[width=\linewidth]{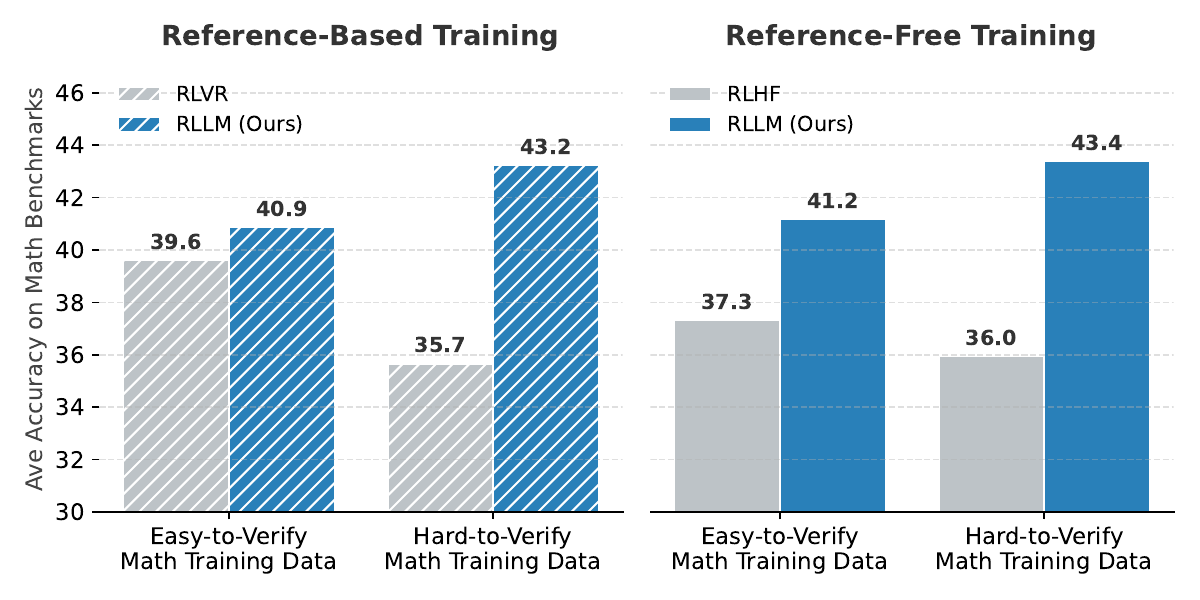}
        \caption{Verifiable Tasks (Math)}
        \label{fig:compare}
    \end{subfigure}
    \hfill
    \begin{subfigure}[b]{0.41\textwidth}
        \centering
        \includegraphics[width=\linewidth]{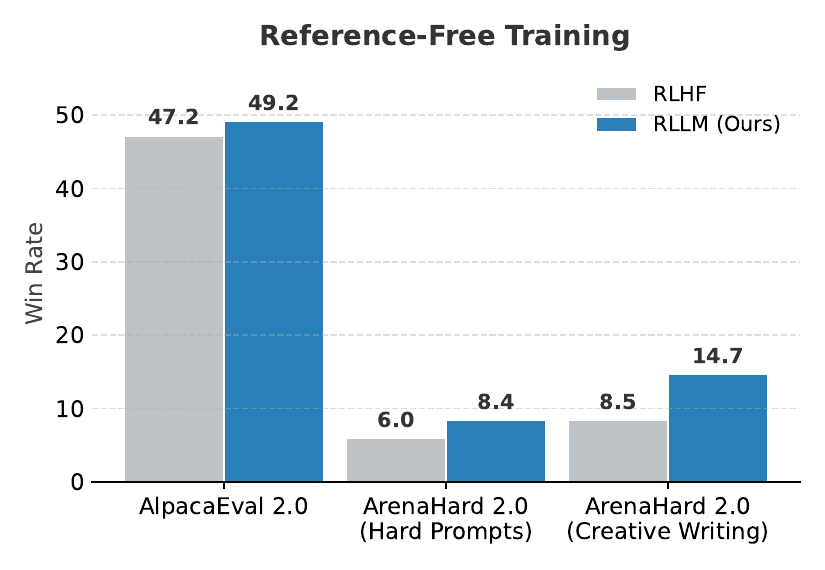}
        \caption{Non-Verifiable Tasks}
        \label{fig:non-ver}
    \end{subfigure}
    
    \caption{Performance comparison of post-trained Qwen3-1.7B models on (a) verifiable tasks (average of five math benchmarks) and (b) non-verifiable instruction-following tasks. Models are trained via RLHF (with \texttt{Skywork-Reward-V2-Llama-3.1-8B} as scalar-RM), RLVR (with \texttt{Math-Verify} as rule-based verifier) and, our RLLM (with \texttt{J1-Qwen3-32B} as LM-as-RM). Post-training data for verifiable tasks is either (1) easy-to-verify, (2) hard-to-verify, (3) reference-free, or (4) reference-based.  Across all these settings, RLLM achieves consistently higher accuracy and win rates than RLVR and RLHF, with particularly large gains when trained on hard-to-verify problems.}
    \label{fig:sampling}

\end{figure*}

\begin{figure}[t]
    \centering
    \includegraphics[width=\linewidth]{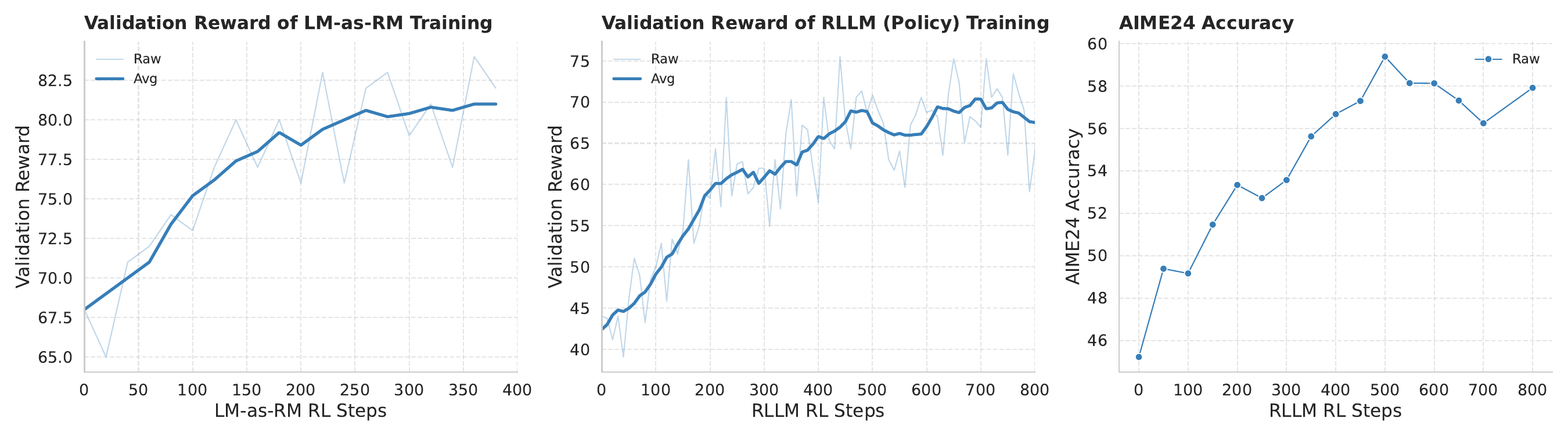}
    \caption{RL validation rewards for LM-as-RM training (left) and RLLM policy training (center), alongside the downstream AIME24 accuracy of the policy model (right). Centered moving average is shown to highlight the trend.}
    \label{fig:reward}
\end{figure}

We validate RLLM through extensive experiments on diverse benchmarks, including competition math (MathArena), physics (Principia), and open-ended instruction following (AlpacaEval, ArenaHard). We demonstrate that RLLM significantly outperforms both RLHF (with SOTA scalar RMs) and RLVR (with rule-based verifiers). Notably, RLLM achieves large gains on hard-to-verify tasks, such as math or physics problems where standard verifiers fail, while simultaneously maintaining strong performance on non-verifiable creative tasks. 
Furthermore, we identify two critical ingredients for success: 
(i) the necessity of a sufficient \textit{generator-verifier gap} (using a stronger RM to train a weaker policy), and (ii) the importance of \textit{on-policy} RM training, as we show that off-policy RMs struggle to provide accurate signals for downstream improvements.

Our contributions are summarized as follows: 
\begin{itemize} 
\item We introduce RLLM, a unified post-training framework that employs a ``thinking'' Language Model as a Reward Model, improving upon the limited rule-based verifiers (required by RLVR) or uninterpretable, shallow scalar scoring (used in RLHF).
\item We demonstrate that RLLM outperforms leading RLHF and RLVR baselines for Qwen, Llama, and Octothinker models of varying sizes, achieving significant gains on both verifiable tasks (Math, Physics) and non-verifiable tasks (AlpacaEval, ArenaHard).
\item We provide a comprehensive analysis of the recipe required for effective LM-as-RM training, empirically proving that on-policy training and a strong generator-verifier capability gap are essential for successful policy improvement. 
\end{itemize}

\subsection{Preliminaries}

\textbf{RLHF: Reinforcement Learning with Human Feedback.}
The standard RLHF~\citep{ouyang2022training, bai2022training} pipeline trains a \textit{scalar} reward model on pairwise human preference data. Each data point $(x,y_c,y_r)\in \mathcal{D}$ includes an instruction $x$, a chosen response $y_c$, and a rejected response $y_r$. The reward model $r_\phi(x,y)$ is optimized using a Bradley-Terry objective:
\begin{equation}
\label{eq:rm}
\mathcal{L}_R = -\mathbb{E}_{(x,y_c,y_r)\in \mathcal{D}}[\log \sigma(r_\phi(x,y_c) - r_\phi(x,y_r))],
\end{equation}
where $\sigma(\cdot)$ is the logistic sigmoid. This trains the model to assign higher rewards to the chosen responses. The trained scalar reward model is then used to optimize a policy model via Reinforcement Learning (e.g., PPO \citep{schulman2017proximal}).
In this paper, we use the term \emph{RLHF} to specifically refer to the classical setup of InstructGPT~\citep{ouyang2022training}, where a \textit{scalar} reward model is trained from human preference data and thereafter, used to optimize a policy model.


\textbf{RLVR: Reinforcement Learning with Verifiable Rewards.} Scalar reward models remain vulnerable to reward hacking, also known as reward over-optimization~\citep{amodei2016concrete, gao2023scaling, eisensteinhelping}, especially under distributional shift, and have shown limited ability to drive improvements on challenging reasoning problems. Thus, in an attempt to prevent such hacking, RLVR replaces the reward model in RLHF with a verification function~\citep{lambert2024tulu} such that the policy only receives a reward when its generated responses are verifiably correct.
Given an instruction $x$, a candidate response $y$, and a reference answer ${y}_{\mathrm{ref}}$, the verification function $\psi(\cdot)$ is defined as:
\begin{equation}
\psi\bigl({x}, {y}, {y}_{\mathrm{ref}}\bigr) =
\begin{cases}
\gamma, & \text{if correct, i.e., }{y} \text{ is equivalent to } {y}_{\mathrm{ref}}, \\
0, & \text{otherwise}.
\end{cases}
\end{equation}

These verifiers are typically rule-based (e.g., \texttt{math-verify} for math or unit tests for coding) but can also be model-based (e.g., a model that checks equivalence between two mathematical expressions). RLVR has shown great success on tasks with verifiable outcomes such as mathematical problem-solving \citep{shao2024deepseekmath} and verifiable instruction-following tasks \citep{lambert2024tulu}. However, its reliance on high-quality reference answers that are also easy-to-verify, makes its application limited, e.g. to easy-to-verify math and coding tasks, but not to non-verifiable or hard-to-verify tasks.

\subsection{RLLM: Reinforcement Learning with Language Models as Reward Models}

To reduce dependence on costly human preference collection, a third paradigm -- Reinforcement Learning from AI Feedback (RLAIF) -- was introduced~\citep{bai2022constitutional, lee2024rlaif} where an AI model is itself used to provide rewards. Our \textbf{RLLM} paradigm is a special case of RLAIF (see \autoref{fig:intro}), which employs a Language Model as a \emph{thinking} Reward Model (LM-as-RM) to unify post-training across diverse task types.

In contrast to scalar reward models $r_\phi(\cdot)$ used in RLHF or deterministic verifiers $\psi(x, y, y_{\text{ref}})$ used in RLVR, RLLM utilizes \textit{generative rewards} $r_{\text{LM}}(\cdot)$ obtained directly from a thinking LLM. 
This unifies the post-training recipe and allows the flexibility of computing rewards through pointwise, pairwise, or listwise judgments, in both reference-free $r_{\text{LM}}(x, y)$ and reference-based $r_{\text{LM}}(x, y_{\text{ref}}, y)$ settings, depending on the task requirements described below.

We consider the standard RL objective for maximizing expected reward:
\begin{equation}
    \max_{\pi_{\theta_{\text{policy}}}} \mathbb{E}_{x \sim \mathcal{D}, y \sim \pi(\cdot|x)} \big[ r_{\text{LM}}(x, y) \big] - \beta \mathbb{D}_{\text{KL}}(\pi_{\theta_{\text{policy}}} || \pi_{\text{ref}})
\end{equation}
where $\pi_{\theta_{\text{policy}}}$ is the policy, $\pi_{\text{ref}}$ is the reference model, and $\beta$ controls the KL-divergence penalty. Importantly, RLLM uses RL {\em both} to train an LLM as an RM  and to optimize the policy using rewards produced by this LM-as-RM.



\subsubsection{Task Settings and LM-as-RM Variants}

We consider a general post-training setting where LLMs are expected to improve across \emph{Verifiable} (e.g., Math, Code) and 
\emph{Non-Verifiable} tasks (e.g., Open-ended Chat). Verifiable tasks are those with objective correctness criteria while non-verifiable tasks are inherently subjective and lack definitive, externally verifiable references~\citep{lu2025writing}. Within verifiable tasks, we further differentiate between \textit{easy-to-verify} problems (gradeable via simple rules like \texttt{math-verify}) and \textit{hard-to-verify} problems (requiring intermediate reasoning to check equivalence). Given this set of tasks, we define the following LM-as-RM reward formulations:

\textbf{Reference-Free Pointwise LM-as-RM}: $r_{LM}(x, y) \rightarrow (t, s)$. 
In a reference-free pointwise setting, the LM-as-RM assigns rewards to individual responses independently. Given an instruction $x$ and a candidate response $y$, the LLM generates a formatted judgment consisting of (i) a reasoning trace $t$ enclosed within \texttt{<think>} tags, where the model analyzes the quality of $y$, and (ii) a final scalar score $s$ within \texttt{<score>} tags. 

\textbf{Reference-Based Pointwise LM-as-RM}: $r_{LM}(x, y_{\mathrm{ref}}, y) \rightarrow (t, s)$. 
In the reference-based setting, the LM-as-RM is additionally provided with a ground-truth reference $y_{\mathrm{ref}}$ and instructed to ground its judgment in this solution. This enables robust evaluation for both easy-to-verify and hard-to-verify problems that require substantial reasoning, effectively allowing the model to act as a \textit{generative verifier}.

\textbf{Pairwise and Listwise LM-as-RM}: $r_{LM}(x, \mathbf{y}) \rightarrow(t, \mathbf{s})$ or  $r_{LM}(x, y_{\mathrm{ref}}, \mathbf{y})\rightarrow(t, \mathbf{s})$.
Pointwise evaluation can be extended to \emph{pairwise} or more generally, to \emph{listwise} evaluation, where the LM-as-RM jointly evaluates multiple candidate responses. In this setting, the input consists of the instruction $x$ and a set of $k$ candidate responses $\mathbf{y} = \{y_1, \dots, y_k\}$.
The model jointly reasons about the candidates in $t$ and outputs a list of  scores $\mathbf{s} = \{s_1, \dots, s_k\}$. Such comparative evaluation has been shown to substantially improve zero-shot LLM-judge performance, particularly for non-verifiable tasks \citep{whitehouse2025menlo}.

In \autoref{sec:prompt}, we provide the prompt template for reference-free pointwise LMs-as-RMs in \autoref{fig:pointwise}, and pairwise in \autoref{fig:pairwise}.  In~\autoref{fig:ref} we show the prompt template for reference-based pointwise LMs-as-RMs.

\subsubsection{LM-as-RM Training via RLVR}

Here we first describe how we use RL to train an LM-as-RM. While off-the-shelf LLMs can also serve as zero-shot RMs, their judgments often become unreliable when policy outputs drift out-of-distribution or when the \textit{generator--verifier gap} is insufficient to provide trustworthy signals~\citep{shao2025deepseekmath}.
To obtain more reliable and task-aligned reward signals, we thus train an LM-as-RM via Reinforcement Learning with Verifiable Rewards using the \methodrm{} framework \citep{whitehouse2025j1}. \methodrm{} constructs synthetic judgment tasks with labels, converting diverse tasks into a unified verifiable format compatible with RLVR-style training.

\textbf{Synthetic Training Data Generation.}
Let \(\pi_{\theta_\text{policy}}\) denote the initial policy LLM that we want to optimize using an LM-as-RM. To train the LM-as-RM, we first sample \emph{on-policy} responses from \(\pi_{\theta_\text{policy}}\) and synthetically annotate the responses for the reward modeling task. Specifically, given a dataset $\mathcal{D}$ with instructions $x$ and optionally available reference answers $y_{\mathrm{ref}}$, we generate reward model training data in three steps: (i) sample a set of responses $\mathbf{y}$ from the policy \(\pi_{\theta_\text{policy}}\); (ii) employ a strong teacher LLM to rate the correctness or quality of these responses, obtaining scores $\mathbf{s}$. For mathematical reasoning tasks, these ratings are typically binary (correct/incorrect); for non-verifiable tasks, the scores span a continuous scale \([s_{\mathrm{min}}, s_{\mathrm{max}}]\) reflecting response quality. Finally, we create a balanced dataset to ensure a uniform distribution over the assigned scores.
The resulting synthetic dataset for LM-as-RM training is denoted as:
\(
    \mathcal{D}_{\text{LM}} = \big\{ (x, y_{\text{ref}}, \mathbf{y}, \mathbf{s}) \mid (x, y_{\text{ref}}) \in \mathcal{D} \big\}
\),
where $y_{\text{ref}} = \emptyset$ in reference-free settings. This would enable us to train an LM-as-RM in both reference-free or reference-based scenarios.


\textbf{RLVR Training.}
Given the scores, we now train the LM-as-RM using RLVR. Following~\citet{whitehouse2025j1}, we format examples from the synthetic dataset $\mathcal{D}_{\text{LM}}$ into seed LM-as-RM prompts (depending on the training configuration), and instruct the model to generate a judgment $(t', s')$, consisting of a reasoning trace $t'$ and a predicted score $s'$. We optimize the model using GRPO \citep{shao2024deepseekmath}, assigning a reward of $1$ if the predicted score matches the teacher score, and $0$ otherwise.




\subsubsection{RLLM Policy Training with LM-as-RM}

Given an LM-as-RM $\pi_{\theta_{\text{LM}}}$, either an off-the-shelf LLM or a trained one  (as described above), we optimize the policy model $\pi_{\theta_\text{policy}}$ using the LM-as-RM's generative rewards. All LM-as-RM variants generate thinking traces and scores; however, only the scores are used for policy optimization. In particular, we sample rollouts $y_i \sim \pi_{\theta_{\text{policy}}}(\cdot \mid x)$ and depending on whether the LM-as-RM is a pointwise, pairwise, or listwise model and if reference answers are available or not, we compute rewards from the rollouts as follows:



\textbf{Pointwise RLLM.}
In this setting, the reward for each rollout $y_i$ is computed independently as $r_i = r_{\text{LM}}(x, y_i)$ (or with reference $y_{\text{ref}}$). This approach is most computationally efficient,  providing an absolute quality score without requiring comparisons with other rollouts.

\textbf{Pairwise RLLM.}
Pairwise comparison requires constructing pairs of rollouts to be evaluated jointly by the LM-as-RM. Let $\mathbf{y} = \{y_1, \dots, y_n\}$ denote the $n$ rollouts for a given prompt. We consider two pairwise strategies and both evaluate the ordered responses $(y_a, y_b)$ and $(y_b, y_a)$ to reduce positional
bias~\citep{zheng2023judging}.

\begin{itemize}

\item \textbf{Exhaustive Pairwise.}
We compare each rollout with every other rollout, resulting in all $\binom{n}{2}$ ordered comparisons. Each rollout appears in $2(n-1)$ comparisons, and its final reward is obtained by averaging the scores across these comparisons. This provides the strongest comparative signal but has $O(n^2)$ judgment complexity.

\item\textbf{Pivot Pairwise.}
To reduce the quadratic complexity of all-pairs, we randomly choose one rollout $y_{\mathrm{pivot}}$ as a reference and pair it with every other rollout in both orders, resulting in $2(n-1)$ comparisons:
\(\quad (y_a,\, y_{\mathrm{pivot}}), \, (y_{\mathrm{pivot}},\, y_a)
\).
All non-pivot rollouts receive scores relative to the same anchor, thus producing more consistent rewards compared to randomly constructing $O(n)$ pairs. The pivot rollout’s reward is obtained by averaging the scores across all the comparisons. This setting remains lightweight, with $O(n)$ LM-as-RM evaluations.

\end{itemize}

\textbf{Listwise RLLM}. Similar constructions can be extended to a list of $K > 2$ comparisons, where the LM-as-RM
jointly evaluates sets of $K$ rollouts rather than pairs, while increasing the reward modeling complexity.

The policy and the LM-as-RM may be initialized from the same or different base LLMs; when they coincide, this corresponds to a self-rewarding RLLM setting \citep{yuan2025selfrewardinglanguagemodels}. Our experiments consider both cases.

\subsection{Experimental Setup}

\subsubsection{LM-as-RM Models} 

\noindent \textbf{Training Recipe.} For the purpose of our main experiments, we use the \methodrm{} recipe to train \emph{on-policy} LM-as-RM models starting from a \emph{large} \texttt{Qwen3-32B} (thinking) model. As we will show later in our analysis (\autoref{sec:analysis}), (1) training a large RM such that it ensures a substantial generator-verifier gap and (2) training the RM on-policy are both critical for best downstream policy improvements. Recall that LM-as-RM training can be done in both reference-free and reference-based settings, in which case we will call the resultant RMs \texttt{J1-Qwen3-32B-RM} and \texttt{J1-Qwen3-32B-Verifier}, respectively. We train these \methodrm{} models by (1) sourcing hard-to-verify prompts from OpenMathReasoning and Principia Collection, (2) generating $16$ responses per prompt from \texttt{Qwen3-1.7B} model (i.e., the policy model to be trained), and (3) labeling each response as correct or incorrect using \texttt{GPT-OSS-120B} (since the final answers are hard-to-verify). See \autoref{fig:appendix_rm_correct} and \autoref{fig:appendix_rm_incorrect} for two such training examples of correct and incorrect responses. This results in a synthetically annotated and balanced dataset of $18,774$ examples for pointwise LM-as-RM training with verifiable correctness rewards. For the purposes of ablations, whenever we train a weaker LM-as-RM model with off-policy samples, we will follow a similar \methodrm{} recipe with the exception that either the base model will change or the responses will be sampled from a different model than the policy to be aligned.

\noindent \textbf{Training Hyperparameters.} Following~\citet{whitehouse2025j1}, we implement all \methodrm{} models on top of \texttt{verl}~\citep{sheng2024hybridflow}. We use a train batch size of $512$ with a maximum sequence length of $4096$ for both input and output. We also set the KL coefficient to $0.001$, the number of rollouts to $8$, and the sampling temperature to $1.0$ during RL training. All 32B \methodrm{} models are trained using $128$ H200 GPUs. The best \methodrm{} checkpoint is chosen based on a held-out validation set of $100$ samples. 

\noindent \textbf{Inference Hyperparameters.} In addition to using \methodrm{} models during policy training, we also perform offline evaluation of these models on in-distribution validation data. Inference is done using vLLM~\citep{kwon2023efficient} with a temperature of $0.6$ and top\_p value of $0.95$. 

\subsubsection{RLLM Policy Models} 
\noindent \textbf{Training Recipe.} In our primary RLLM experiments, we optimize a \texttt{Qwen3-1.7B} (Instruct) policy using the aforementioned \texttt{J1-Qwen3-32B} reward models.\footnote{Qwen instruct models are already post-trained using RL, making them harder to improve on, compared to their base counterparts. Despite that, we conduct all RLLM experiments on top of such strong instruct models and show further improvements.} This choice of model sizes for the policy and the RM creates a substantial generator–verifier gap, which we find to be essential for achieving strong downstream policy improvements. Starting with the same \texttt{Qwen3-1.7B} policy, we experiment with four different RLLM training paradigms that vary in their training data or the LM-as-RM:
\begin{itemize}
    \item \textbf{Policy Trained on Hard-to-Verify Samples without Reference Answer.} This forms the main experimental setting of RLLM where we assume access to a post-training dataset of only hard-to-verify math problems without any reference answers. As LMs continue to grow in capability, we will increasingly confront tasks for which no human-curated reference answers exist. To test such a post-training setting, we sample $3000$ \emph{hard-to-verify} problems from the OpenMathReasoning dataset~\citep{moshkov2025aimo}. Following~\citet{tao2025hybrid}, we select a sample as hard-to-verify if the final answer cannot be deterministically validated using a rule-based symbolic verifier like \texttt{math\_verify} (see example in Appendix \autoref{fig:appendix_hard_to_verify}) and easy-to-verify, otherwise (see example in Appendix \autoref{fig:appendix_easy_to_verify}). Given the reference-free setting, our RLLM recipe uses \texttt{J1-Qwen3-32B-RM} as the reward model for optimization.
    \item \textbf{Policy Trained on Hard-to-Verify Samples with Reference Answer.} This resembles an RLLM setting where we train on the same dataset as above but this time, assuming access to reference answers. Since these reference answers are hard-to-verify, our RLLM recipe uses \texttt{J1-Qwen3-32B-Verifier} as the (reference-based) verifier for policy training.
    \item \textbf{Policy Trained on Easy-to-Verify Samples with/without Reference Answer.} Our motivation to perform RLLM on \emph{easy-to-verify} samples is to show that even when reference answers exist and they are easy-to-verify, RLLM with a strong LM-as-RM can substitute rule-based verifiers like \texttt{math-verify}. Note that an LM-as-RM has the advantage of additionally evaluating the entire CoT and identifying \emph{process} errors, unlike \texttt{math-verify} that only performs equivalence checks between the final answers. For the purpose of this study, we conduct RLLM experiments by sampling $3000$ \emph{easy-to-verify} training examples from the same OpenMathReasoning dataset~\citep{moshkov2025aimo}.
\end{itemize}

In subsequent analyses and ablations of RLLM, we also consider different combinations of policy and reward models e.g., other sizes of Qwen3 models, Octothinker \citep{wang2025octothinker}, and Llama models. For non-verifiable instruction-following tasks, we train RLLM models on 1K Wildchat prompts~\citep{zhao2024wildchat}.

\noindent \textbf{Training Hyperparameters.} We implement all policy models on top of \texttt{fairseq2}~\citep{balioglu2023fairseq2}. Models are trained using $64$ H200 GPUs, allocating $48$ GPUs for the trainer and $16$ GPUs for inference. Fairseq2's trainer is implemented as \texttt{Single Program Multiple Data (SMPD)} and all models (policy model, reference model, and \texttt{J1} reward model) run as Ray actors. RLLM training uses a batch size of $32$ with $8$ rollouts per prompt, sampled with a temperature of $1.0$. Similar to \methodrm{} training, we also set the KL coefficient to $0.001$ for policy training. RLLM models are trained for a maximum of $1000$ steps, checkpointing every $50$ steps. To mitigate the effect of mismatch in log probabilities between the trainer and the inference engine in GRPO training, we apply truncated importance sampling correction with the maximum clip ratio set to $2.0$. We also set max input and generation length to 4096 tokens and train \texttt{Qwen3-1.7B} in non-thinking mode to prevent long thinking sequences from exceeding the context window, ensuring that the reward models can evaluate the complete answers.

\noindent \textbf{Inference Hyperparameters.} We evaluate RLLM-trained policy models using vLLM with a temperature of $0.6$ and a top\_p value of $0.95$. The maximum decoding length is set to 40K tokens. Given the hybrid nature of Qwen3 models, we evaluate our post-trained models in both thinking and non-thinking modes.

\subsubsection{Evaluation Benchmarks and Metrics}

To evaluate the effectiveness of RLLM as a general post-training recipe, we conduct experiments on both verifiable reasoning tasks as well as non-verifiable tasks.

\noindent \textbf{Verifiable Reasoning Benchmarks.} Within verifiable reasoning, we experiment with both easy-to-verify benchmarks (e.g., questions with numerical answers) as well as hard-to-verify benchmarks (e.g., questions with mathematical objects as answers like equations).

\begin{itemize}
    \item \textbf{Easy-to-verify Math Benchmarks.} We evaluate on five popular competition math benchmarks from MathArena~\citep{balunovic2025matharena} -- AIME24, AIME25, BRUMO25, HMMT24, and HMMT25. For each benchmark, we report \texttt{mean@16} results.
    \item \textbf{Hard-to-verify Physics Benchmark.} We also evaluate on the Physics subset of PrincipiaBench. This serves as a test bed for both hard-to-verify reasoning problems as well as our models' generalization to a different domain (given that RLLM's training data only consists of math prompts). We use o3~\citep{o3} to evaluate the correctness of the final answers and report \texttt{mean@8} scores.
\end{itemize}

\noindent \textbf{Non-verifiable Tasks.} Finally, we also evaluate RLLM on two non-verifiable instruction following benchmarks -- AlpacaEval 2.0~\citep{alpaca_eval} and ArenaHard 2.0~\citep{licrowdsourced}. Following past work~\citep{lanchantin2025bridging}, we report win rates (with and without length-control), using GPT-4o and GPT-4.1 as the evaluators for AlpacaEval and ArenaHard respectively. 

\subsubsection{Baselines}

We compare RLLM to two groups of baselines, RLHF and RLVR.

\begin{itemize}
    \item \textbf{RLHF.} This represents RL post-training with \emph{scalar} reward models. In particular, we consider two strong general-purpose RMs -- \texttt{Skywork-Reward-V2-Llama-3.1-8B}~\citep{liu2025skywork}\footnote{This is the best-performing publicly available reward model, according to the RewardBench2 leaderboard: \url{https://huggingface.co/spaces/allenai/reward-bench}.} and \texttt{Athene-RM-8B}~\citep{Athene2024}. We also compare against a math-specific reward model \texttt{AceMath-7B-RM}~\citep{acemath2024}. 
    \item \textbf{RLVR.} This refers to RL post-training with a verifier. The verifier operates in the presence of a reference answer and can either be rule-based (e.g., \texttt{math-verify}) or model-based (e.g., \texttt{general-verifier}~\citep{ma2025generalreasoner}. Even though \texttt{general-verifier} is a model-based verifier, it only checks the equivalence between the final answers (without any CoT reasoning). We compare RLLM to both these RLVR methods. 
\end{itemize}

Within the RLLM framework, we compare our \methodrm{}-trained LMs-as-RMs to various prompted LLMs-as-RMs. These belong to different families and are of different sizes and capabilities, thus enabling us to analyze how scaling up the RM/Verifier affects policy training and downstream task improvements. In particular, we report RLLM results with \texttt{Llama-3.1-8B-Instruct}, \texttt{Qwen-1.7B}, \texttt{Qwen-32B}, and \texttt{GPT-OSS-120B} as prompted  LMs-as-RMs.

\input{rllm_tables/main1}
\input{rllm_tables/main2}
\subsection{Results}

\subsubsection{Main Results}

\noindent \textbf{RLLM (with prompted and trained LMs-as-RMs) outperforms RLHF (with scalar RMs) on both easy-to-verify and hard-to-verify reasoning tasks.} First, in~\autoref{tab:main_rllm_vs_rlhf}, we compare different post-trained Qwen3-1.7B models, optimized via either scalar RMs (RLHF) or LM-as-RM (RLLM). Within RLLM, we compare our trained J1-Qwen3-32B-RM to different prompted LMs-as-RMs of varying sizes. All models are trained on the same hard-to-verify math prompts in a reference-free setting. Our main conclusions are listed below:

\begin{itemize}
    \item All RLLM models (rows annotated in green), except for the one trained with the weaker Llama-3.1-8B-Instruct model, outperform all RLHF models, showcasing the effectiveness of LMs-as-RMs over scalar RMs ($36.12 \rightarrow 43.41$). Importantly, while scalar RMs like Skywork-Reward-V2 may be the best-performing RM on offline benchmarks (e.g., RewardBench2), such performance does not translate to best downstream performance when doing online RL. This echoes the findings of past work that also shows that highest scoring RMs on static benchmarks does not ensure a good post-trained model~\citep{malik2025rewardbench}.
    \item In a self-rewarding RLLM setting where both the policy and the RM are Qwen3-1.7B models (5th row), we observe improvements over the base policy ($35.20 \rightarrow 38.84$). This result demonstrates that a thinking LM-as-RM can provide sufficiently high-quality rewards to drive self-improvement, at least to some extent.
    
    \item Furthermore, RLLM scales with the size and the capability of the LM-as-RM, with our on-policy-trained J1-Qwen3-32B-RM obtaining the best results and even outperforming a larger GPT-OSS-120B as the RM. In particular, our best RLLM model obtains an average of 8\% absolute improvement on competition math over the Qwen3-1.7B Instruct model. This is noteworthy given that Qwen3 models are already heavily post-trained for these benchmarks.
    \item RLLM-trained models also generalize to the out-of-domain and hard-to-verify Physics benchmark, while only training on math prompts.
\end{itemize}

In summary, our results show that RLLM with strong LLMs-as-RMs can outperform RLHF. Moreover, on-policy training of an LM-as-RM can lead to further improvements. Using an LM for both the policy and as an RM also provides a natural framework for building self-rewarding and self-improving LLMs~\citep{yuan2025selfrewardinglanguagemodels}.

\noindent \textbf{RLLM also outperforms RLVR (with rule-based and model-based verifiers).} \autoref{tab:main_rllm_vs_rlvr} compares our RLLM-trained models with RLVR-trained models. In this setting, we train on the same dataset as above but assume access to reference answers. We list our main takeaways below:

\begin{itemize}
    \item When reference answers do exist, under the same RLLM framework, we can also employ strong thinking-LLMs as \emph{reference-based verifiers}. RLLM with such thinking-verifiers can thus outperform RLVR with rule-based verifiers like \texttt{math-verify} ($35.67 \rightarrow 43.24$), which is known to fail in cases where answers are complex mathematical objects~\citep{tao2025hybrid}.
    \item Similar to the reference-free setting described above, training an on-policy J1-Qwen3-32B-Verifier model improves math results by 2\% over prompted Qwen3-32B ($41.16 \rightarrow 43.24$), thus highlighting the effectiveness of RL-trained LMs-as-RMs over prompted ones. Our RLLM model also generalizes better to the hard-to-verify physics benchmark.
\end{itemize}

\noindent \textbf{RLLM on hard-to-verify training data outperforms RLVR on easy-to-verify training data.} So far, we have shown that when the training data is hard-to-verify, RLLM can outperform RLHF and RLVR. A natural question then arises that if we already have access to easy-to-verify data (in which case, a rule-based verifier like \texttt{math-verify} might also suffice), do we even need to train on hard-to-verify data? We answer this in~\autoref{tab:main_rllm_easy} by training RLLM on all different subsets of training data -- easy-to-verify, hard-to-verify, reference-free, and reference-based. Our conclusions are as follows:

\begin{itemize}
    \item Reference-free RLLM training on hard-to-verify data outperforms reference-based RLVR training on easy-to-verify data ($39.63 \rightarrow 43.41$). This suggests that one way to scale RL for post-training (beyond easily verifiable domains) is to train strong LMs-as-RMs.
    \item RLVR training on easy-to-verify data works better than on hard-to-verify data because of more accurate rewards ($35.67 \rightarrow 39.63$). However, even when training on easy-to-verify data, RLLM can match or outperform RLHF and RLVR. 
    \item RLLM training on hard-to-verify data leads to better downstream performance than training on easy-to-verify data ($41.18 \rightarrow 43.41$). 
\end{itemize}

Overall, this helps establish RLLM as a unified and superior recipe across all post-training regimes. Our results also indicate that the following RLLM setting leads to the best downstream performance: (1) hard training prompts with the (2) strongest on-policy trained LM-as-RM, (3) capable of rewarding rollouts either in reference-free or reference-based setting.

Given the hybrid nature of Qwen3 models, we also evaluate our post-trained models in \emph{non-thinking} mode. Results in Appendix~\autoref{tab:appendix_non_thinking} demonstrate that RLLM outperforms RLHF and RLVR by large margins. We draw similar conclusions in Appendix~\autoref{tab:appendix_octothinker} when experimenting with a Llama-based OctoThinker-8B-Hybrid-Base model and in Appendix~\autoref{tab:appendix_qwen8} with a larger Qwen3-8B seed model.

\input{rllm_tables/non-verifiable}

\noindent \textbf{Beyond verifiable domains, RLLM is also performant on non-verifiable instruction-following tasks.} Next, in~\autoref{tab:non_verifiable}, we evaluate the effectiveness of RLLM for non-verifiable tasks. Unlike RLLM training for verifiable tasks which used a \emph{pointwise} LM-as-RM, here we use a \emph{pairwise} LM-as-RM. This is because non-verifiable tasks do not have a strict notion of correctness and past works have shown that such responses are also easier to evaluate in a comparative setting. In particular, we consider a self-rewarding RLLM setting with Qwen3-1.7B both as the policy and the LM-as-RM, construct pairs of rollouts, and prompt the Qwen3-1.7B LM-as-RM to assign scores between 0-10 to each of the two rollouts. These scores are then averaged across all pairs to get pointwise reward estimates. Later in the ablations, we also compare pointwise, pairwise, and k-wise reward assignment for these tasks. We observe that RLLM can match or outperform RLHF, while obtaining better win rates for harder prompts. Appendix ~\autoref{tab:appendix_non_verifiable_8b} reports similar conclusions when training a Qwen3-8B policy model. This result again reinforces the effectiveness of RLLM as a general framework for post-training on all kinds of tasks. Like verifiable tasks, we expect further improvements on non-verifiable tasks upon on-policy training of the LM-as-RM, which we leave for future work.

\subsubsection{Analyses and Ablations}
\label{sec:analysis}
\noindent \textbf{Generator-Verifier Gap.} In this section, we investigate the impact of the \emph{generator–verifier gap} on RLLM training, specifically examining how the capability gap between the policy LM and the LM-as-RM influences downstream policy improvements. Recall that for our main experiments, we trained a Qwen3-1.7B policy with a J1-Qwen3-32B-RM where the RM was trained on-policy (by sampling responses from the Qwen3-1.7B policy). Now we ask if we train a weaker 1.7B LM-as-RM on its own responses i.e., J1-Qwen-1.7B-RM, can that also lead to downstream improvements? As shown in~\autoref{tab:gen_ver_gap}, we do not observe further improvements on top of the prompted Qwen3-1.7B-as-RM with J1 training. This result is further evidenced by~\autoref{fig:j1_acc_no_ref}, where we compare the raw accuracy of different LMs-as-RMs on an in-distribution validation set. We observe that \methodrm{} training of a Qwen3-32B model leads to 10\% improvement in judgment accuracy (averaged across 8 seeds) while providing almost no improvement on top of Qwen3-1.7B. In summary, training a Qwen3-1.7B model to evaluate its own responses leads to limited success and consequently, the resultant RM also does not lead to any downstream policy improvements. This underscores the importance of the capability gap between the generator and the verifier for obtaining downstream improvements. In Appendix~\autoref{fig:appendix_j1_traces}, we show examples of correct and incorrect thinking traces generated by J1-Qwen3-1.7B-RM and J1-Qwen3-32B-RM respectively.


\noindent \textbf{Off-policy vs On-policy trained LM-as-RM.} In~\autoref{tab:on_vs_off_policy}, we compare an on-policy trained LM-as-RM with two off-policy trained RMs. All three RMs are trained on top of the same Qwen3-32B model using the same \methodrm{} recipe, differing only in their training data: the off-policy RMs are trained on responses generated either by a weaker Llama model or by a stronger Qwen3-8B model. Athough~\autoref{fig:off_on_policy} shows that \methodrm{} training improves judgment accuracy for all these models on their respective in-distribution validation sets, the off-policy trained LMs-as-RMs do not transfer to downstream policy improvements.\footnote{The reward curves also reflect the respective hardness of the RM data wherein responses from the weaker Llama model are easiest to judge, followed by Qwen3-1.7B, and Qwen3-8B.} This again shows that RM capability improvements measured on static, offline benchmarks (with different data distributions) may not always be indicative of downstream task improvements because of lack of OOD generalization.

\noindent \textbf{Reference-free vs Reference-based LM-as-RM.} LMs-as-RMs have the flexibility of being trained and used in both a reference-free and a reference-based manner. In~\autoref{fig:j1_acc_ref}, we compare LMs-as-RMs with and without reference answers, indicated by the suffixes `-Verifier' and `-RM' respectively. Unsurprisingly, presence of reference answers allows for more accurate judgments with Qwen3-32B-Verifier outperforming Qwen3-32B-RM by 16\%. The performance improves further to 92\% after \methodrm{} training, showing that our recipe for training LMs-as-RMs is performant in both reference-free and reference-based settings. 

\noindent \textbf{Pointwise, Pairwise, vs Listwise LM-as-RM Rewards.} Recall that for non-verifiable tasks, we employed a pairwise LM-as-RM, primarily because non-verifiable tasks benefit from relative judgments. In~\autoref{tab:k-wise}, we compare the effect of scaling up reward modeling compute by conducting either pointwise, pairwise, or listwise scoring from the LM-as-RM. Since the complexity of pairwise scoring is quadratic in the number of rollouts, we also explore a second pairwise setting where one of the rollouts is chosen at random as a pivot (or reference) rollout to compare against. We observe that on the hard prompts, win rates improve with more judgments while for the other categories, results mostly saturate at pairwise comparisons. Overall, this highlights the flexibility of an LM-as-RM's rewarding mechanism, allowing increased compute to be spent on evaluation.

\input{rllm_tables/ablation1}
\input{rllm_tables/ablation2}

\input{rllm_tables/k-wise}

\begin{figure*}[t]
    \centering
    \includegraphics[width=0.5\textwidth]{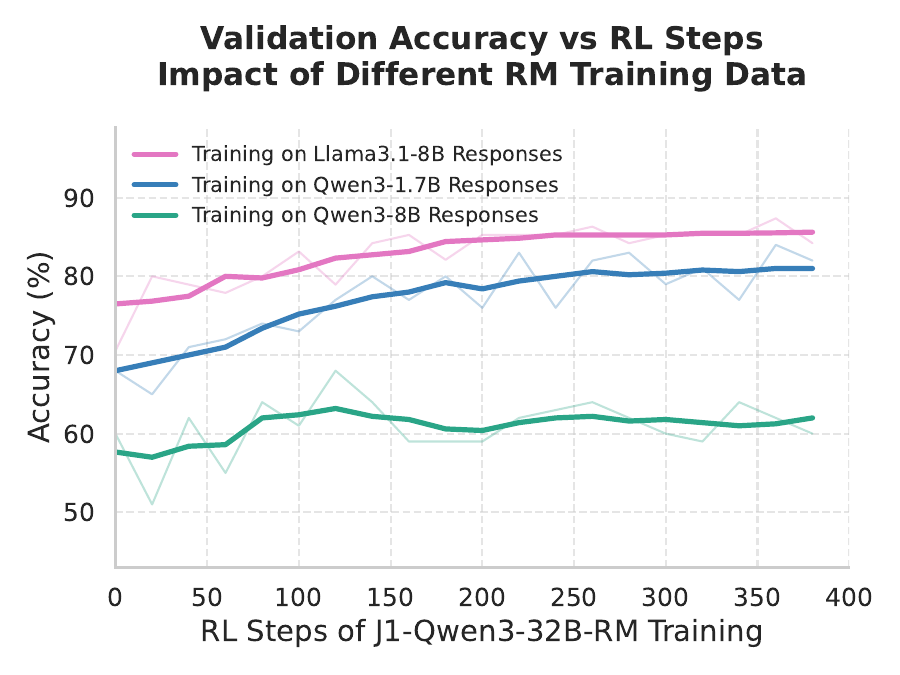}
    \caption{ \label{fig:off_on_policy}Comparison of validation reward curves for \methodrm{} training on different distributions of data. Data is either sampled from a Llama3 model (off-policy weakest), Qwen3-8B model (off-policy strongest), or Qwen3-1.7B model (on-policy). While all LMs-as-RMs improve on their respective in-distribution validation data, off-policy trained RMs do not lead to downstream task improvements potentially because of lack of OOD generalization  (\autoref{tab:on_vs_off_policy}).}
\end{figure*}

\begin{figure*}[t]
    \centering
    \begin{subfigure}[l]{0.42\textwidth}
        \centering
        \includegraphics[width=\linewidth]{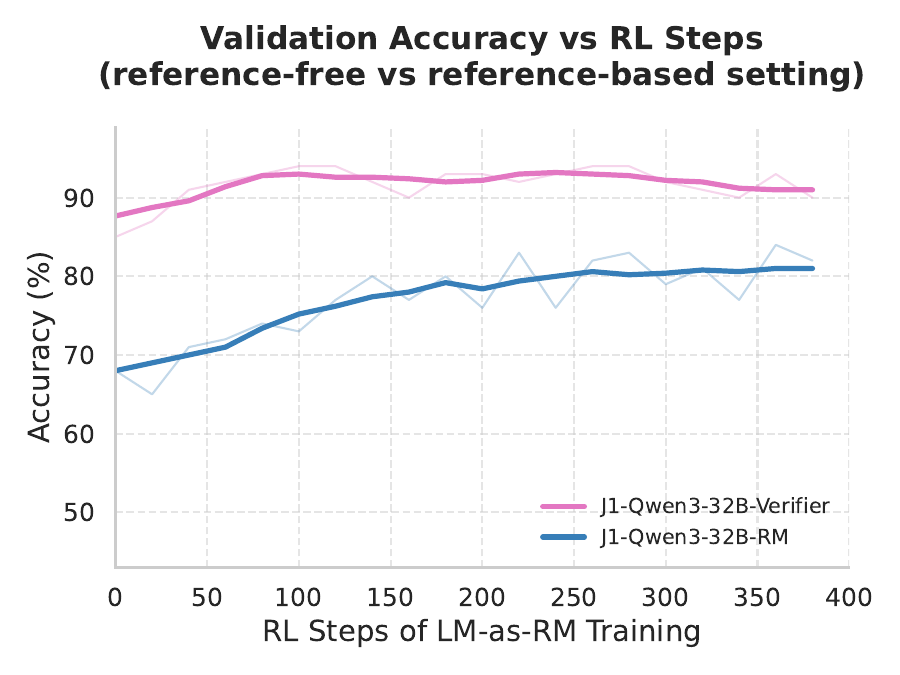}

    \end{subfigure}
    \hfill
    \begin{subfigure}[r]{0.55\textwidth}
        \centering
        \includegraphics[width=\linewidth]{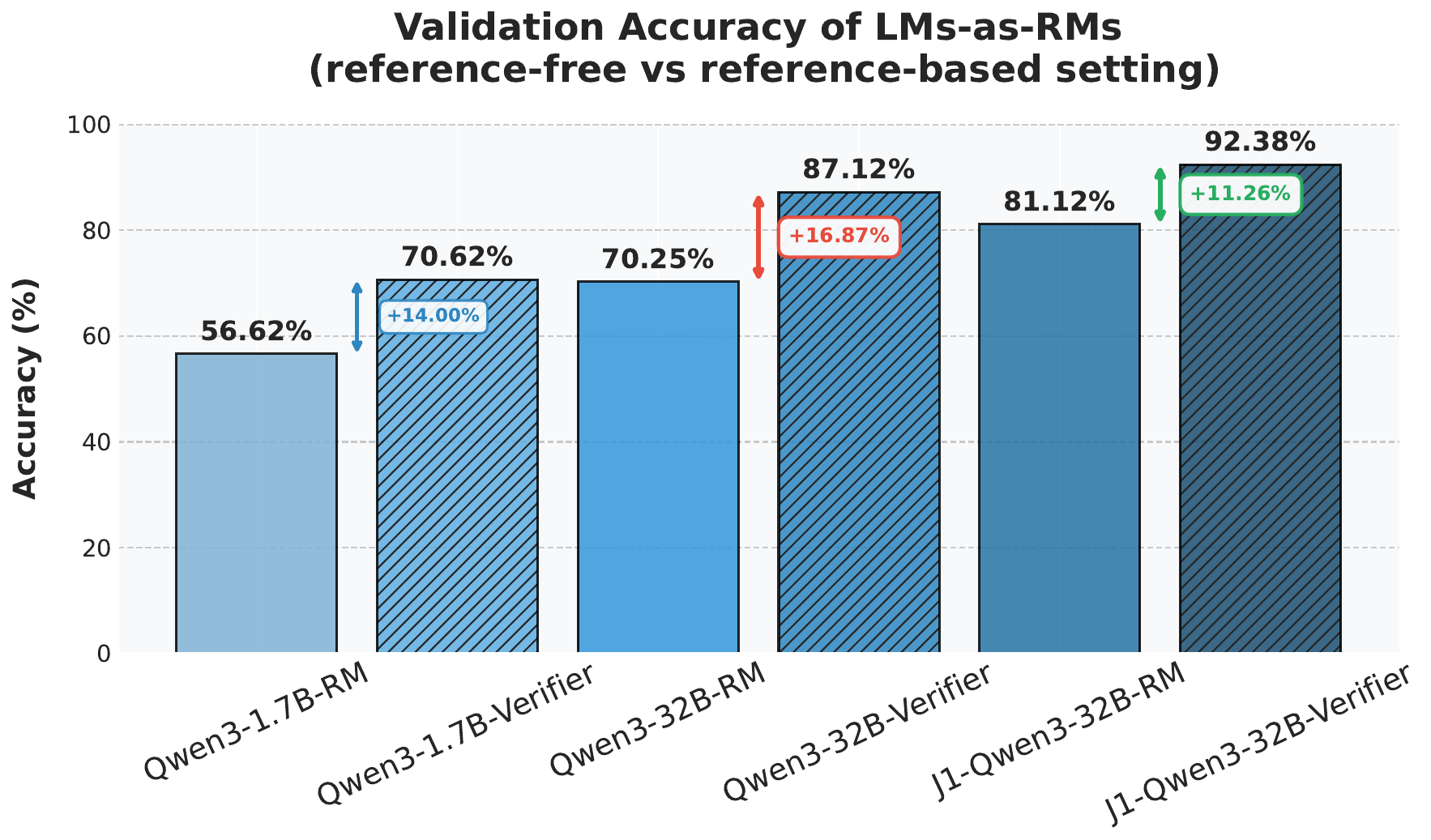}

    \end{subfigure}
    
    \caption{(a) Validation reward curves for a reference-free J1-Qwen3-32B-RM and a reference-based J1-Qwen3-32B-Verifier. (b) Comparison of different LMs-as-RMs evaluated in either a reference-free setting (suffixed with `-RM') or in a reference-based setting (suffixed with `-Verifier'). Reference-based LMs-as-RMs are more accurate than reference-free ones (e.g., J1-Qwen3-32B-Verifier is 10\% better than J1-Qwen3-32B-RM).}
    \label{fig:j1_acc_ref}

\end{figure*}

\subsection{Conclusion}

We showed that RLLM -- RL with (RL-trained) language models as reward models -- can serve as a single, unified post-training recipe across easy-to-verify, hard-to-verify, and non-verifiable tasks. Through extensive experiments, we demonstrated that RLLM outperforms both RLHF (with scalar RMs) and RLVR (with rule-based rewards), showcasing particularly large gains when training on hard-to-verify tasks. We also studied the importance of on-policy training of LM-as-RM models alongside the impact of generator-verifier gap and showed that these are important components for successful RLLM training.

\clearpage
\newpage

\definecolor{mygreen}{HTML}{228B22}
\definecolor{tempgreen}{HTML}{30C854}
\definecolor{temporange}{HTML}{E69F00}
\definecolor{divpoblue}{HTML}{565AE1}
\definecolor{myred}{HTML}{DC143C} 
\definecolor{figureblue}{HTML}{6C8EBF} 
\definecolor{matplotlibblue}{HTML}{1F77B4}
\definecolor{matplotliborange}{HTML}{FF7F0E}
\definecolor{matplotlibgreen}{HTML}{2CA02C}
\definecolor{figurepurple}{HTML}{9673A6} 
\definecolor{figureyellow}{HTML}{D6B656} 
\definecolor{forestgreen}{HTML}{228B22}

\section{Learning to Aggregate through Online Reinforcement Learning} \label{sec:para}

\begin{figure}[h!]
    \centering
\includegraphics[width=0.99\linewidth]{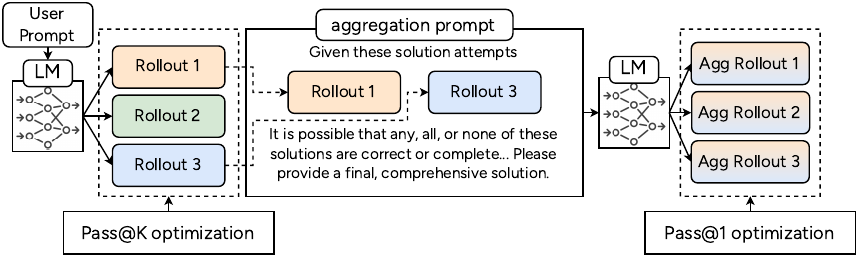}
    \caption{\textbf{\method{} Training.} Illustration of our parallel thinking scaffolding and our method. We use pass@k optimization \citep{chen2025passktrainingadaptivelybalancing} for optimizing the initial round of responses and pass@1 optimization (standard RLVR) for optimizing the aggregation rollouts, and train end-to-end.}
    \label{fig:teaser}
\end{figure}

\begin{figure}[h!]
    \centering
\includegraphics[width=0.9\linewidth]{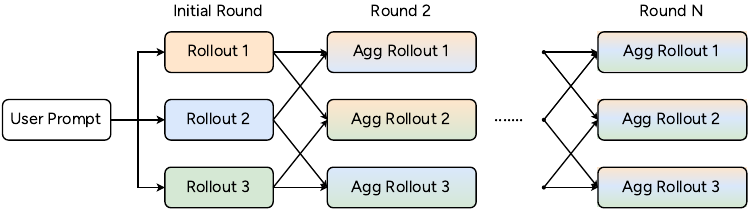}
    \caption{\textbf{\method{} Inference.} During each round, we sample rollouts from the past aggregation round, pack them into the aggregation prompt, and perform inference to obtain the next pool of rollouts.}
    \label{fig:paragator_inference}
\end{figure}

\subsection{Overview}


Scaling up test-time compute has emerged as a powerful tool for improving language models (LMs) on challenging reasoning tasks \citep{openai2024openaio1card, snell2024scaling, wu2025inference}. A widely adopted pattern is to generate multiple solution attempts in parallel and then either select or aggregate among them \citep{wang2022self, chen2023universalselfconsistencylargelanguage, brown2024largelanguagemonkeysscaling, pan2025learningadaptiveparallelreasoning, meituanlongcatteam2026longcatflashthinking2601technicalreport}. This paradigm underlies many of the strongest recent systems in complex reasoning \citep{tongyidr, li2025parallelmuseagenticparallelthinking, pacore2025}. In particular, several works train an \emph{aggregator} model to learn how to properly synthesize a final answer from a large pool of parallel-generated candidates \citep{zhao2025majorityrightrltraining, venkatraman2025recursiveselfaggregationunlocksdeep}.

However, current parallel thinking strategies suffer from two main limitations. First, most methods treat the solution generator as fixed and only optimize the aggregator \citep{zhao2025majorityrightrltraining, venkatraman2025recursiveselfaggregationunlocksdeep}. This creates a fundamental mismatch: the aggregator is trained on off-policy generations that may differ substantially from what it encounters at inference time, leading to suboptimal performance. Second, even when the generator is trained on-policy, standard outcome-based RL encourages the model to converge toward a single high-probability solution mode \citep{yue2025limit-of-rlvr, li2025jointlyreinforcingdiversityquality}. This diversity collapse leaves the aggregator with redundant candidates that provide little complementary information, thus undermining the core premise of parallel thinking, which relies on diverse solutions to enable effective aggregation.

In this work, we address both limitations through a unified online training framework. Our key insight is that \textbf{the generator and the aggregator require different optimization objectives}: the generator should produce diverse, complementary solutions, while the aggregator should synthesize them into a single best answer. To this end, we apply pass@k optimization \citep{chen2025passktrainingadaptivelybalancing} to the parallel generation stage, explicitly encouraging diversity by rewarding the model when any of its k samples succeed. For the aggregation stage, we use pass@1 optimization, training the model to reliably produce the correct final answer. Crucially, both stages are trained online and end-to-end, ensuring the generator learns to produce candidates that are not only diverse but also well-suited for downstream aggregation.

Our contributions are threefold:

\begin{itemize}[leftmargin=*]
    \item We show that the self-aggregation method is upper bounded by the pass@k performance and that the model is unable to synthesize correct solutions based on incorrect ones.
    \item We develop a framework that enables online training of self-aggregation methods, where a single model is optimized for both pass@k generation and pass@1 aggregation end-to-end.
    \item We validate the effectiveness of our method, which generalizes across competition math and scientific reasoning setups, including our newly introduced PrincipiaBench.
\end{itemize}




\subsection{Preliminaries}

Our goal is to train a language model that takes a problem as input, generates a set of candidate solutions in parallel, and then aggregates them over an arbitrary number of steps to produce a final answer. Formally, let $x$ be a given problem and $y^\star$ its ground-truth solution.
We consider two steps for a model to perform:
(i) candidate sampling: the model reads the problem and generates a set of $m$ candidate solutions $y_{\{1:m\}}$ and
(ii) self-aggregation: the model reads both the problem and the set of candidate solutions and generates a final solution $\tilde{y}$. The performance of the model can be evaluated at either an initial solution $y_i$, or the final solution $\tilde{y}$.

\subsection{Method: \method{}}

Our method jointly trains a single language model, $\mathcal{M}_\theta$ to (i) generate diverse candidate solutions and (ii) aggregate these solutions into a final answer. Both stages are optimized end-to-end using online RL.

Specifically, we train \method{} using reinforcement learning with verifiable rewards (RLVR), with rewards computed at two levels: on the initial set of candidate solutions and on the final aggregated answer. The candidate set is rewarded with pass@k, incentivizing the model to produce plausible but diverse solutions. The final answer is rewarded with pass@1, incentivizing the model to reliably produce a single correct solution. Both candidate generation and aggregation rollouts are sampled \emph{online} from the current policy during training.

\subsubsection{Candidate Solution Generation and Pool Selection}
For each problem \(x\), we first generate a pool of \(m\) candidate solutions by sampling from the model $m$ times:

\[
y_i \sim \mathcal{M}_\theta(y \mid p_C,  x), \quad i = 1,\dots,m.
\]
At this stage, we evaluate each candidate with a verifiable reward function
\[
r(y_i, y^\star) \in \{0, 1\},
\]
which indicates whether the candidate answer matches the ground-truth solution \(y^\star\).

\textbf{Pass@K Candidate Optimization} Pass@k is defined as the maximum reward given $k$ random samples $\{y_1, .. y_k\} \sim \pi_\theta(\cdot|x)$ from the policy: $\text{pass}@k := \max[r(y_1), r(y_2), ..., r(y_k)]$. We use the pass@k optimization method described in \citet{chen2025passktrainingadaptivelybalancing}, where the advantages of a correct response and an incorrect response are given by:
$$A_\text{correct} = \frac{1 - \mu(x)}{\sigma(x)}, ~~A_\text{incorrect} = \frac{1 - \mu(x) - \frac{\binom{N_{\text{incorrect}} - 1}{k- 1}}{\binom{N-1}{k-1}}}{\sigma(x)},$$ where $N$ is the group size, $N_\text{incorrect}$ is the number of incorrect rollouts in this group, and $\mu(x)$ and $\sigma(x)$ are the mean and standard deviation of the rewards for the group whose prompt is $x$. Compared to standard GRPO \citep{shao2024deepseekmath}, only the advantage of incorrect examples is modified by an offset of $\frac{\binom{N_{\text{incorrect}} - 1}{k- 1}}{\binom{N-1}{k-1}}$. In our work, we make a further modification as in Dr.GRPO \citep{liu2025understanding} by removing the division by $\sigma(x)$.

Intuitively, the model is rewarded when \textit{at least one} of its \(m\) attempts solves the problem, which encourages spreading probability mass across complementary solution modes rather than collapsing onto a single trajectory.

We do not perform any additional heuristic filtering or reranking: all generated candidates are visible to the aggregator and contribute to the candidate-level RL signal.

\subsubsection{Self-Aggregation}
Given the problem \(x\) and the candidate set \(y_{1:m}\), we prompt the same model to act as an aggregator. The input is the problem concatenated with the candidates in a fixed, structured format, for example:

\begin{tcolorbox}[colback=gray!5,colframe=gray!60,title=Aggregation Prompt ($p_A$)]
Given the following problem: \newline
\{PROBLEM\} \newline
and these solution attempts:\newline
\{CANDIDATE\_1\}\newline
\{CANDIDATE\_2\}\newline
...\newline
\{CANDIDATE\_N\}\newline
=========================\newline
it is possible that any, all, or none of these solutions are correct or complete. Carefully review the provided solutions, using them as starting points --- correcting mistakes, filling in gaps, and/or combining useful ideas --- to produce a final, comprehensive, and correct solution to the problem.
\end{tcolorbox}

Conditioned on this input, the model generates an aggregated solution
\[
\tilde{y} \sim \mathcal{M}_\theta(y \mid p_A, x, y_{1:m}).
\]
We then compute an aggregation-level verifiable reward
\[
R_{\text{agg}}(x, y_{1:m}, \tilde{y}) = r(\tilde{y}, y^\star) \in \{0,1\}.
\]

Crucially, the aggregator is always trained on-policy: during training, it sees candidate pools sampled from the current generator \(\mathcal{M}_\theta\), rather than from a frozen or separately trained model. This alignment between training and inference eliminates the off-policy mismatch common in prior self-aggregation methods and ensures that the generator learns to produce candidates that are well-suited for downstream aggregation.

\textbf{Pass@1 Aggregation Optimization}

The aggregated solutions use pass@1 performance: the aggregator receives a reward of 1 if and only if its final answer is correct. Unlike the candidate stage, only the single aggregated trajectory is rewarded, pushing the model to reliably synthesize the best answer from the available candidates.

\subsubsection{Training Details}

We optimize both stages simultaneously with RLVR. For each training example, we generate a candidate pool \(y_{1:m}\) from \(\mathcal{M}_\theta(y \mid x)\), and an aggregation trajectory \(\tilde{y}\) from \(\mathcal{M}_\theta(y \mid x, y_{1:m})\). We then update \(\theta\) using GRPO \citep{shao2024deepseekmath} with a KL-regularized RLVR loss. Because both terms are computed from online rollouts of the same policy \(\mathcal{M}_\theta\), the generator learns to produce candidate sets that are not only individually plausible and diverse, but also \textbf{useful} for the aggregator it will later become.

\subsubsection{Inference Details}
During training, we optimize a single round of aggregation over one candidate pool. At inference time, however, we naturally generalize this to multiple iterations of aggregation, enabling sequential scaling in addition to the learned parallel sampling.

Concretely, given problem $x$, we first sample an initial pool of $m$ candidates $y_{1:m}^{(0)} \sim \mathcal{M}_\theta(y \mid x)$ and sample $m$ aggregated solutions $\tilde{y}_{1:m}^{(0)} \sim \mathcal{M}_\theta(y \mid x, y_{1:m}^{(0)})$. We then form an updated candidate pool $y_{1:m}^{(1)} = \tilde{y}_{1:m}^{(0)}$ from these aggregation rollouts. The model is then prompted again in aggregation mode on $(x, y_{1:m}^{(1)})$ to produce a refined set of solutions $\tilde{y}_{1:m}^{(1)}$, and so on. This continues for $T$ iterations before returning the final aggregated answer.

This iterative procedure preserves the same generator–aggregator interface used during training, while allowing the model at test time to repeatedly refine its reasoning over an evolving pool of candidate solutions under a fixed compute budget.

\subsection{Motivating Experiments: Inference-Only Results}

In this section, we first show that basic aggregation of parallel generations yields improvements on frontier open-sourced models, such as Kimi-K2-Thinking \citep{kimiteam2025kimik2openagentic} (\S \ref{sec:inference_on_frontier}). 
This motivates employing and improving aggregation procedures will likely continue to be useful as models scale, and that the results of our training experiments should generalize beyond the smaller models we employ.
We then further show that self-aggregation requires diversity among the responses to be packed into the aggregation prompt in order to perform better (\S \ref{sec:role_of_passk}), motivating our training approach.

\subsubsection{Self-aggregation improves frontier models}
\label{sec:inference_on_frontier}

In ~\autoref{fig:paragator_inference_results}, we plot the performance of self-aggregation on Kimi-K2-Thinking \citep{kimiteam2025kimik2openagentic}, Qwen3-4B-Thinking-2507, and Qwen3-4B-Instruct-2507 \citep{yang2025qwen3}, 3 strong open-sourced models, on 4 competition math benchmarks: AIME 2025 \citep{aops2025aime}, Brumo 2025 \citep{balunovic2025matharena}, HMMT 2025 \citep{balunovic2025matharena}, and IMO-Answerbench \citep{luong-etal-2025-towards}.

\begin{figure}[h!]
    \centering
\hspace{-5pt}\includegraphics[width=0.99\linewidth]{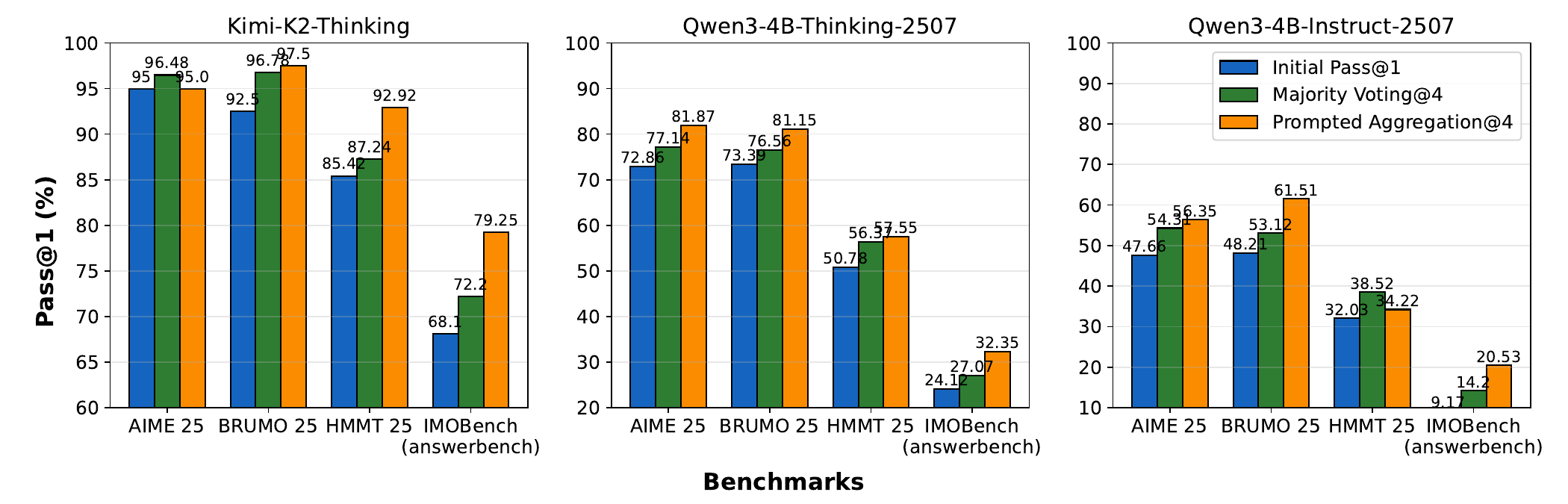}
    \caption{Parallel generation + aggregation (\textcolor{orange}{orange}) brings gains across 4 competition math benchmarks (AIME, Brumo, HMMT and IMO-Answerbench) on top of 3 strong models: Kimi-K2-Thinking, Qwen3-4B-Thinking-2507, and Qwen3-4B-Instruct-2507, compared to standard generation (\textcolor{blue}{blue}) and majority voting (\textcolor{forestgreen}{green}).}
    \label{fig:paragator_inference_results}
\end{figure}

\subsubsection{The role of candidate diversity (pass@k) in self-aggregation}
\label{sec:role_of_passk}

In \autoref{fig:repeated_aggregation}, we plot the performance of multiple rounds of aggregation using the scaffold detailed in \autoref{fig:paragator_inference}. We show pass@1, pass@4, and majority voting@4 curves. We see that the pass@1 performance (\textcolor{blue}{blue}) never exceeds the initial pass@4 (\textcolor{forestgreen}{green}), showing that the asymptotic performance is bounded by pass@k at the initial round, motivating our pass@k optimization method.

\begin{figure}[h]
    \centering
    \begin{subfigure}[h]{0.48\textwidth}
        \centering
        \includegraphics[width=\linewidth]{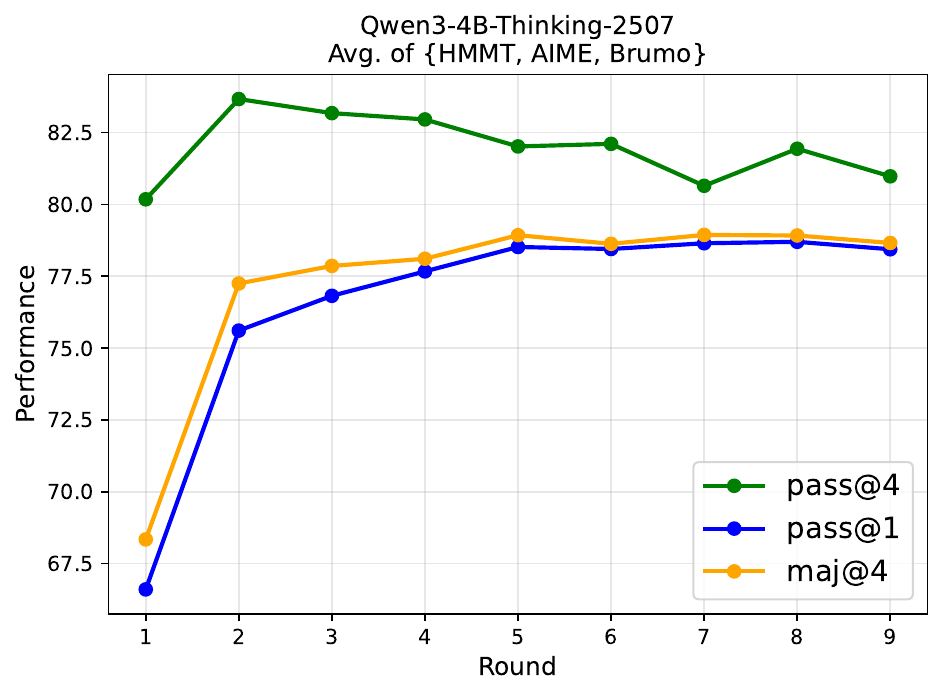}
        \caption{Self-aggregation performance on Qwen3-4B-Thinking-2507.}
        \label{fig:repeated_agg_thinking}
    \end{subfigure}
    \hfill
    \begin{subfigure}[h]{0.48\textwidth}
        \centering
        \includegraphics[width=\linewidth]{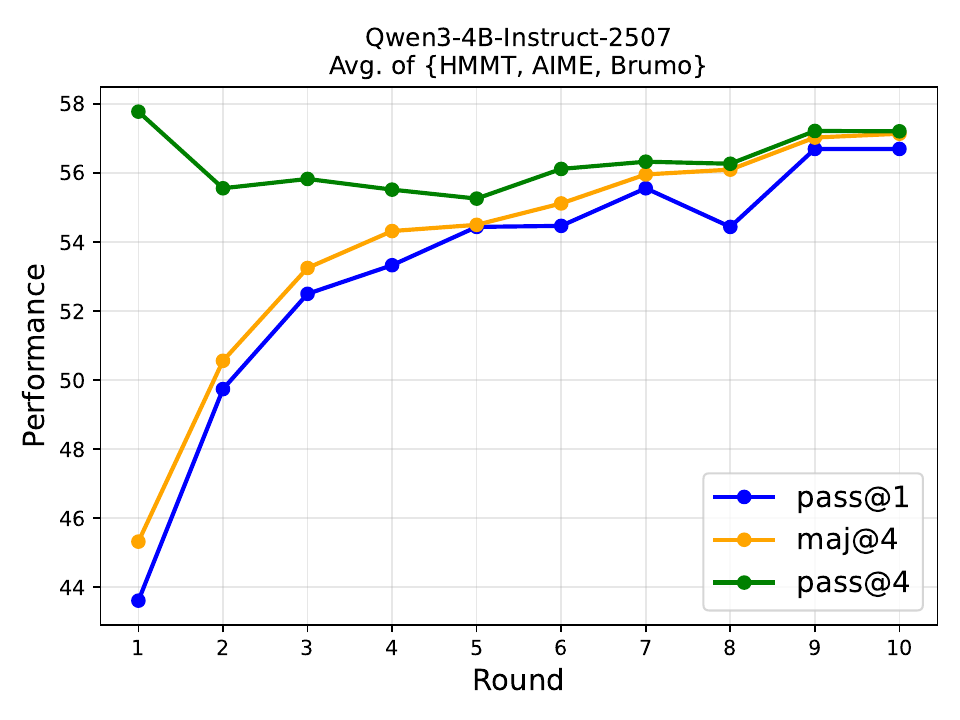}
        \caption{Self-aggregation performance on Qwen3-4B-Instruct-2507.}
        \label{fig:repeated_agg_instruct}
    \end{subfigure}
    \caption{Performance of repeated aggregation is upper bounded by the initial pass@k (\textcolor{forestgreen}{green}) for both Qwen3-4B-Thinking-2507 (left) and Qwen3-4B-Instruct-2507 (right). The asymptotic performance is upper-bounded by the pass@k at the initial round.}
    \label{fig:repeated_aggregation}
\end{figure}

To further illustrate the importance of pass@k, we vary the initial sampling temperature (0.6, 0.8, 1.0) while keeping the aggregation sampling temperature fixed (1.0). We report the scores in \autoref{tab:temp_ablation}. Pass@1 performance is similar at the initial round but a higher initial pass@k results in a higher aggregation pass@1.

\begin{table}[t]
\centering
\begin{tabular}{lccc}
\toprule
 & \multicolumn{3}{c}{Avg. of HMMT, Brumo, and AIME} \\
\cmidrule(lr){2-4}
Initial Temperature & Initial Pass@1 & Initial Pass@k & $\rightarrow$ Aggregation Pass@1 \\
\midrule
T=0.6 & 65.8 & 81.01 & 74.11 \\
T=0.8 & 65.8 & 81.3 & 74.9 \\
T=1.0 & 65.7 & \textbf{81.7} & \textbf{75.2} \\
\bottomrule
\end{tabular}
\caption{Model = Qwen3-4B-Thinking-2507. Effect of initial sampling temperature on decoding performance, averaged over HMMT, Brumo, and AIME. Increasing the initial temperature leaves pass@1 nearly unchanged while improving pass@k, resulting in higher aggregation performance.}
\label{tab:temp_ablation}
\end{table}

\subsection{Main Experiments}

We validate training our \method{} in two regimes: competition math (\S \ref{subsec:deepscaler}) and scientific reasoning (\S \ref{subsec:principia}).

\subsubsection{Competition Math}
\label{subsec:deepscaler}
\paragraph{Setup} We train on a subset of the DeepScaleR dataset \citep{deepscaler2025}, consisting of 10k prompts. Following previous work \citep{hu2025openreasonerzeroopensourceapproach}, we train on a base model that has not been post-trained: Qwen3-4B-Base \citep{yang2025qwen3technicalreport}. We report scores on the following 4 benchmarks: HMMT 2025 \citep{balunovic2025matharena}, Brumo 2025 \citep{balunovic2025matharena}, AIME 2025 \citep{aops2025aime}, and OlympiadBench \citep{he-etal-2024-olympiadbench}. We generate 32 solutions with the following sampling parameters: $\{\text{Temp} = 1.0, \text{top-p}=1, \text{top-k}=-1\}$ and report pass@1 for up to 3 aggregation rounds.

We compare \method{} against prompting the base model, Dr.GRPO \citep{liu2025understanding} (GRPO with token-level loss aggregation and the standard-deviation divisor removed), and the offline aggregation training approach proposed independently by AggLM \citep{zhao2025majorityrightrltraining} and RSA \citep{venkatraman2025recursiveselfaggregationunlocksdeep}. \autoref{table:comparison} summarizes which components of the pipeline are trained by each method.

\begin{table}[h]
\centering
\setlength{\tabcolsep}{6pt}  
\begin{tabular}{ccc}
\toprule
Training Method &
\benchcenter{Initial Round}{\includegraphics[width=0.17\textwidth]{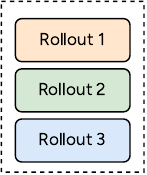}} &
\benchcenter{Aggregation Round}{\includegraphics[width=0.17\textwidth]{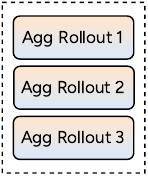}} \\
\midrule
\benchcenter{No Aggregation}{(Only Dr.~GRPO on initial candidates)}
& pass@1 optimization
& No updates \\
\midrule
\benchcenter{Offline Aggregation}{\citep{ zhao2025majorityrightrltraining,venkatraman2025recursiveselfaggregationunlocksdeep}}
& No updates
& \benchcenter{pass@1 optimization}{(data prepared offline)} \\
\midrule
Online Aggregation
& No updates
& pass@1 optimization \\
\midrule
Online Multitask Aggregation
& pass@1 optimization
& pass@1 optimization \\
\midrule
\method~(Ours)
& pass@k optimization
& pass@1 optimization \\
\bottomrule
\end{tabular}
\caption{
Comparison of training strategies across the initial and aggregation rounds.
Columns show whether model parameters are updated via pass@1 or pass@k optimization, or kept fixed.
}
\label{table:comparison}
\end{table}

\paragraph{Results} We plot the reward curves for both the initial round and the aggregation round at \autoref{fig:paratrooper_reward_cuves_deepscaler}, and report the detailed scores at \autoref{tab:deepscaler_results}. The curves show a clear trade-off in prior baselines: Dr.GRPO attains reasonable initial-round reward but lags in aggregation-round reward, while offline aggregation training exhibits the opposite pattern, improving aggregation performance at the expense of the initial round. Online multitask training partially mitigates this mismatch by optimizing both rounds jointly, but still underperforms our method. In contrast, \method{} consistently achieves the highest reward in both rounds, which translates into the strongest overall pass rates across aggregation steps. As shown in \autoref{tab:deepscaler_results}, \method{} delivers the best average performance after aggregation and achieves the strongest results on most benchmarks, demonstrating the benefit of jointly training initial-generation and aggregation behaviors within a unified online multitask framework.

\begin{figure}[ht!]
    \centering
    \begin{subfigure}[t]{0.48\textwidth}
        \centering
        \includegraphics[width=\linewidth]{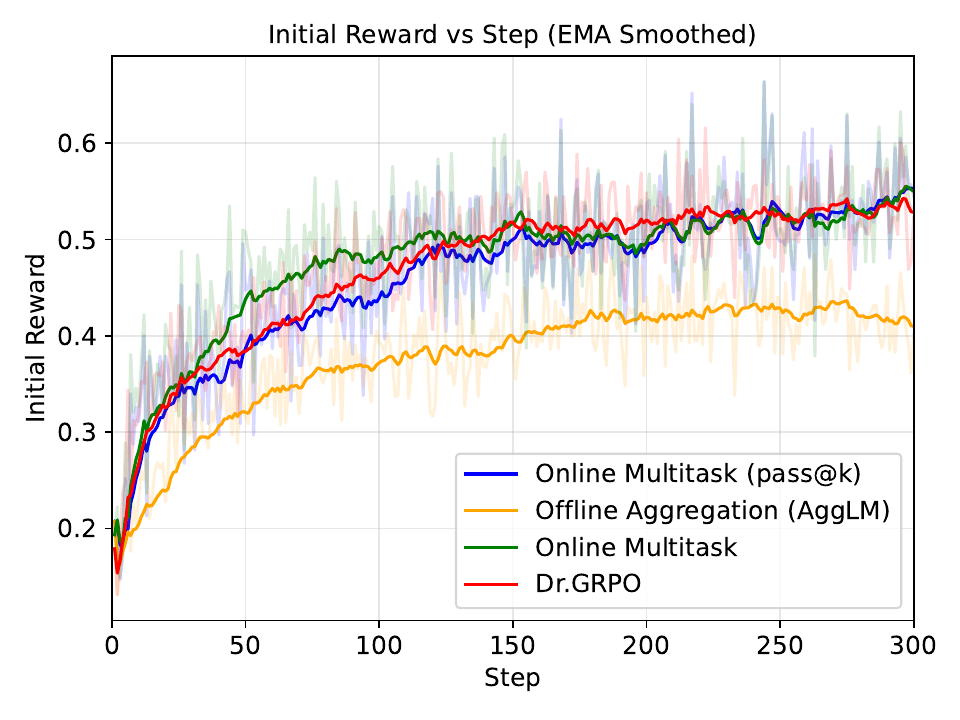}
        \caption{Reward by training step of the initial round.}
        \label{fig:deepscaler_init_reward}
    \end{subfigure}
    \hfill
    \begin{subfigure}[t]{0.48\textwidth}
        \centering
        \includegraphics[width=\linewidth]{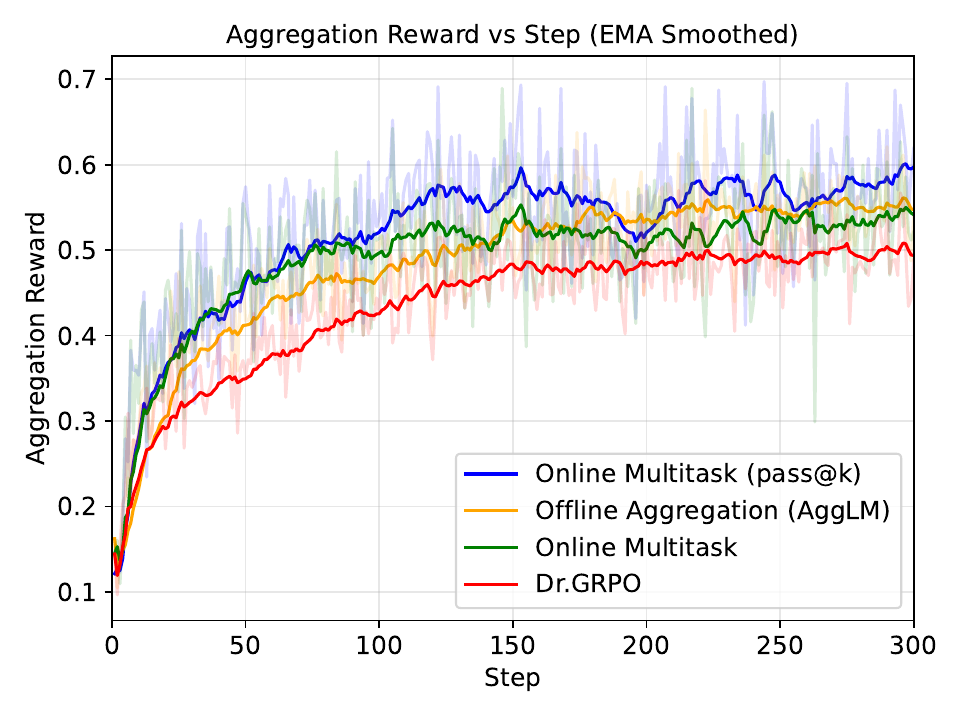}
        \caption{Reward by training step for the aggregation round.}
        \label{fig:deepscaler_agg_reward}
    \end{subfigure}
    \caption{Reward curves for training Qwen3-4B-Base on deepscaler using different methods. The baseline (\textcolor{red}{Dr.GRPO}) only optimizes the initial round, making the performance lag behind during the aggregation round. Offline aggregation only training (\textcolor{orange!80}{AggLM}) only optimizes the aggregation round, making it lag behind during the initial round. Our proposed method (\textcolor{blue}{\method}) achieves the highest reward.}
    \label{fig:paratrooper_reward_cuves_deepscaler}
    \vspace{-5pt}
\end{figure}

\begin{table}[ht!]
\centering
\setlength{\tabcolsep}{4pt}
\begin{tabular}{lcccccc}
\toprule
Method & Round & HMMT & Brumo & AIME & Olympiad & Avg. \\
\midrule
\multirow{4}{*}{Qwen3-4B-Base}
 & Initial       &  3.20/5.15 & 16.67/29.04 &  8.20/21.33 & 33.72/47.41 & 15.45/25.73 \\
 & Agg 1         &  2.50 & 14.25 & 10.30 & 30.28 & 14.33 \\
 & Agg 2         &  1.80 & 14.39 &  9.20 & 28.15 & 13.38 \\
 & Agg 3         &  1.60 & 11.38 &  8.03 & 25.16 & 11.54  \\
\midrule

\multirow{4}{*}{\bench{Dr.GRPO}{\citep{liu2025understanding}}}
 & Initial       &  9.79/15.94 & 25.05/35.27 & 19.27/28.11 & \textbf{45.65}/54.92 & 24.94/33.56 \\
 & Agg 1         & 11.30 & 26.35 & 21.15 & 47.25 & 26.51 \\
 & Agg 2         & 12.40 & 27.60 & 21.88 & 48.33 & 27.55 \\
 & Agg 3         & 13.70 & 28.10 & 21.22 & 49.50 & 28.13 \\
\midrule
\multirow{4}{*}{\bench{Offline Aggregation Training}{\citep{zhao2025majorityrightrltraining}}}
 & Initial       & 6.34/9.02 & 19.31/27.94  & 14.49/23.65  & 38.62/49.10 & 19.69/27.43 \\
 & Agg 1         & 12.08 & 24.54 & 21.05  & 44.46 & 25.53 \\
 & Agg 2         & 15.10 & 29.32  & \textbf{26.44} & 47.67 & 29.63 \\
 & Agg 3         & 17.01 & 31.33 & \textbf{27.37} & 50.11 & 31.46 \\
\midrule
\multirow{4}{*}{Online Multitask}
 & Initial       & 10.16/14.49 & 26.15/35.11 & 18.12/27.95 & 45.33/55.20 & 24.94/- \\
 & Agg 1         & 10.05 & 29.11 & 20.62 & 46.90 & 26.67 \\
 & Agg 2         & 10.68 & 27.24 & 19.38 & 47.22 & 26.13 \\
 & Agg 3         & 12.90 & 29.30 & 20.50 & 47.15 & 27.46 \\
\midrule

\multirow{4}{*}{\bench{\method-Zero-4B}{(Online Multitask w/ pass@k)}}
 & Initial       & \textbf{11.15}/22.11 & \textbf{28.07}/39.96 & \textbf{20.05}/32.40 & 45.42/54.49 & 26.17/33.19 \\
 & Agg 1         & \textbf{13.44} & \textbf{32.24} & \textbf{23.59} & \textbf{51.93} & 30.30 \\
 & Agg 2         & \textbf{15.29} & \textbf{34.67} & 25.45 & \textbf{53.09} & 32.12 \\
 & Agg 3         & \textbf{17.60} & \textbf{36.25} & 27.25 & \textbf{54.89} & 34.00 \\
\bottomrule
\end{tabular}
\caption{Competition Math evaluation results by method and round, with fine-tuning starting from  Qwen3-4B-Base. Best values in each column bolded. Numbers = Pass@1/Pass@4. \method{} gives the overall best results.}
\label{tab:deepscaler_results}
\end{table}

\subsubsection{Scientific Reasoning}
\label{subsec:principia}
\paragraph{Setup} We train on a subset of the Principia dataset, consisting of a total of 30,000 questions. 15,000 are sampled from the \texttt{mathematical\_objects} subset, and 15,000 are sampled from the \texttt{numerical} subset. We report pass@1 scores both on the original PrincipiaBench, consisting of 2558 questions\footnote{Note that 128 filtered problems described in \Cref{subsec:principia_bench} were used in this evaluation as well due to concurrent experimentation.}, and scores in Pass@1 for HMMT 2025, BrumoMath 2025, and AIME 2025. We train on two different models: Qwen3-4B-Base and Qwen3-4B-Instruct-2507 \citep{yang2025qwen3}. The former resembles the ``RL-Zero" paradigm, where we perform RL directly on a base model that has not been post-trained. The latter resembles a more practical setup where we enhance parallel thinking behavior on a strong post-trained model.

\begin{figure}[h!]
    \centering
    \begin{subfigure}[t]{0.48\textwidth}
        \centering
        \includegraphics[width=\linewidth]{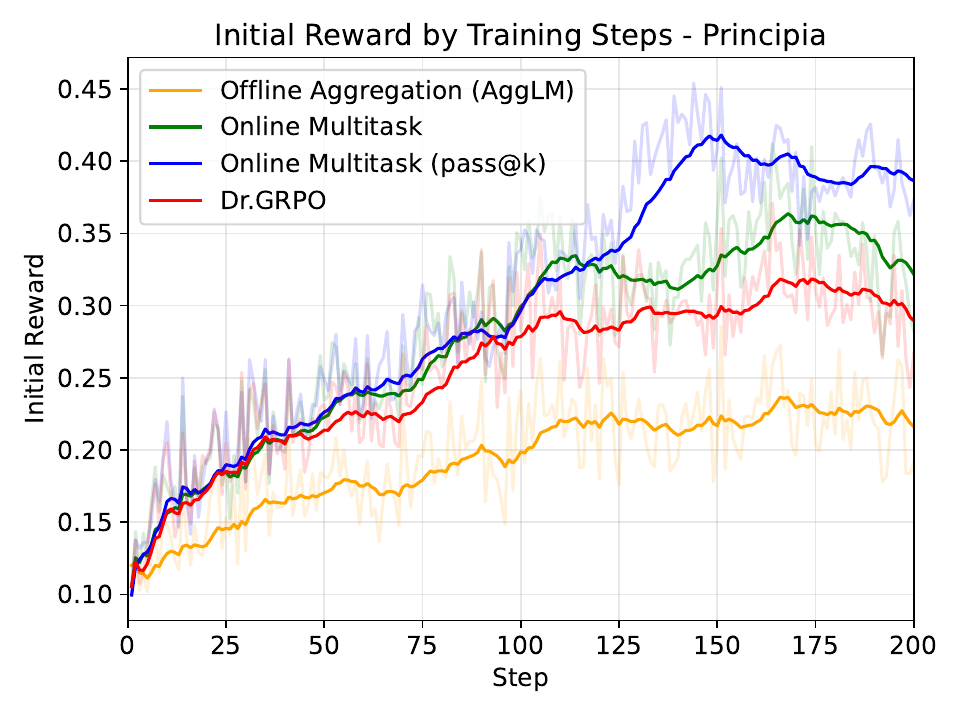}
        \caption{Reward by training step of the initial round.}
        \label{fig:principia_init_reward}
    \end{subfigure}
    \hfill
    \begin{subfigure}[t]{0.48\textwidth}
        \centering
        \includegraphics[width=\linewidth]{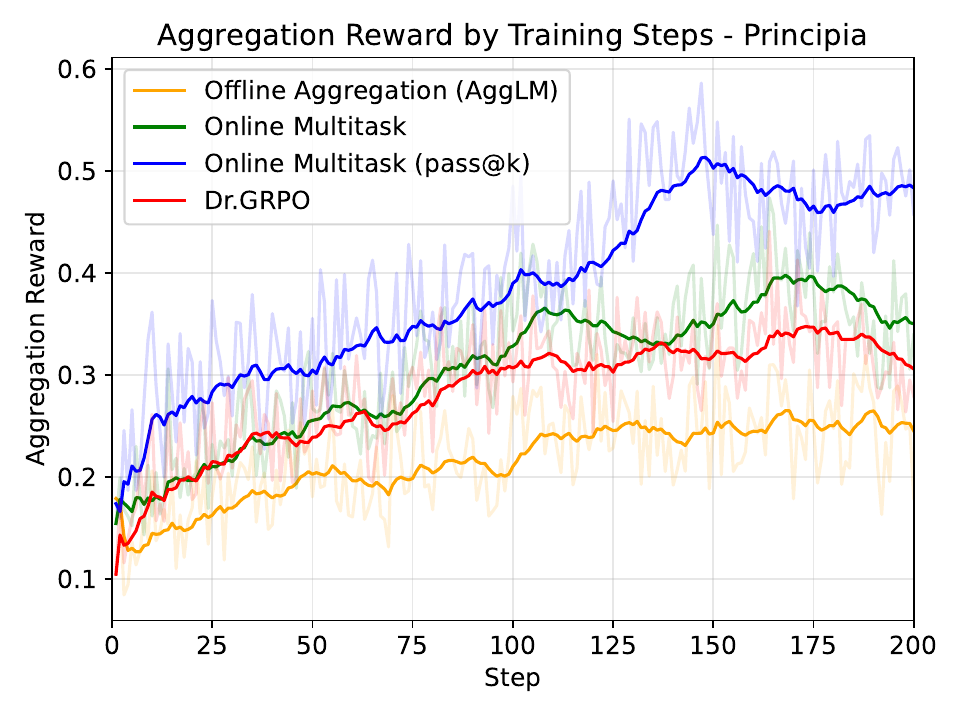}
        \caption{Reward by training step for the aggregation round.}
        \label{fig:principia_agg_reward}
    \end{subfigure}
    \caption{Reward curves for training Qwen3-4B-Base on Principia. Optimizing for pass@k during the initial round (\textcolor{blue}{\method}) achieves the highest reward on both the initial round and the aggregation round.}
    \label{fig:main}
\end{figure}

\begin{table}[h!]
\setlength{\tabcolsep}{3.5pt}
\small
\begin{tabular}{l*{9}{c}}
\toprule
& \multicolumn{5}{c}{\textbf{Scientific Reasoning (PrincipiaBench)}}
& \multicolumn{4}{c}{\textbf{Competition Math}} \\
\cmidrule(lr){2-6} \cmidrule(lr){7-10}
\textbf{Model}
& \textbf{RealMath} & \textbf{Physics} & \textbf{ARB} & \textbf{SuperGPQA} & \textbf{Avg.}
& \textbf{AIME} & \textbf{Brumo} & \textbf{HMMT} & \textbf{Avg.} \\
\midrule
\multicolumn{10}{c}{Experiments on Qwen3-4B-Base} \\
\midrule
Qwen3-4B-Base              & 9.11 & 5.69 & 17.44 & 11.01 & 10.81 & 8.20 & 16.67 & 3.20 & 9.36 \\

~~~~+aggregation             & 8.35 & 4.79 & 18.24 & 10.68 & 10.52 & 10.30 & 14.25 & 2.50 & 9.02 \vspace{5pt} \\
Dr.GRPO                    & 17.40   & 15.65 & 35.26 & 29.3 & 24.40 & 20.93 & 23.19 & 8.14 & 17.42 \\
~~~~+aggregation             & 19.32 & 21.06  & 44.17 & 32.28 & 29.21 & 25.95 & \textbf{31.07} & \textbf{13.10} & 23.37 \vspace{5pt} \\
Offline aggregation (AggLM)& 10.15 & 12.37 & 25.40 & 15.26 & 15.80 & 9.13 & 17.11 & 7.42 & 11.22 \\

~~~~+aggregation             & 12.10 & 19.64 & 31.27 & 21.62  & 21.15 & 16.91 & 25.65 & 12.40 & 18.32 \vspace{5pt} \\
Online Multitask           & 18.05 & 16.33 & 33.26 & 24.18 & 22.96 & 19.14 & 22.37 & 7.98 & 16.50 \\
~~~~+aggregation             & 23.07 & 25.31  & 43.40 & 29.65 & 30.36 & 26.19 & 29.54 & 14.11 & 23.28 \vspace{5pt} \\

\method-Zero-4B-Principia          & 17.71 & 21.09 & 38.62 & 28.75 &  26.54 & 21.05 & 22.58 & 7.04 & 16.89 \\
~~~~+aggregation             & \textbf{27.40} & \textbf{29.24} & \textbf{45.17} & \textbf{33.24} & \textbf{33.77} & \textbf{27.96} & 31.06 & 12.15 & \textbf{23.72} \\
\midrule
\multicolumn{10}{c}{Experiments on Qwen3-4B-Instruct-2507} \\
\midrule
Qwen3-4B-Instruct-2507     & 22.94 & 33.75 & 63.30 & 41.87 & 40.47 & 47.71 & 30.83 & 55.42 & 44.65 \\
~~~~+aggregation             & 29.45 & 44.12 & 72.56 & 49.09 & 48.80 & 56.46 & 35.94 & 61.88 & 51.43  \vspace{5pt}\\

Dr.GRPO (with KL)                    & 28.13 & 49.95 & 68.74 & 52.31 & 49.78 & 58.02 & 44.69 & 61.56 & 54.20 \\

~~~~+aggregation             & 35.44 & 57.14 & 76.67 & \textbf{58.94} & 57.05 & 63.65 & \textbf{47.29} & 65.52 & 58.79 \vspace{5pt} \\

Offline aggregation (AggLM)& 25.19 & 40.05 & 62.75 & 44.46 & 43.11 & 51.06 & 37.47 & 54.91 & 47.81 \\
~~~~+aggregation             & 33.19 & 51.04 & 72.13 & 52.30 & 52.17  & 62.42 & 43.16 & 65.14 & 56.91 \vspace{5pt} \\

 Online Multitask           & 29.16 & 47.07 & 67.65 & 53.44 & 49.32 & 58.36 & 41.73 & 59.45 & 53.18 \\

~~~~+aggregation             & \textbf{37.12} & \textbf{60.10} & \textbf{79.23} & 58.25 & \textbf{58.68} & 64.24 & 45.26 & \textbf{67.33} & 58.94 \vspace{5pt} \\

\method-4B-Instruct-Principia           & 27.93 & 46.18 & 66.70 & 52.39 & 48.30 & 59.44 & 42.32 & 59.67 & 53.81 \\

~~~~+aggregation             & 36.84 & 59.41 & 79.19 & 56.85 & 58.07  & \textbf{71.20} & 46.65 & 66.28 & \textbf{61.38} \vspace{5pt} \\
\bottomrule
\end{tabular}
\caption{Scientific reasoning (PrincipiaBench) and competition math evaluation results. Numbers denote Pass@1. Best values per column and model group are bolded. \method{} gives the overall best results.}
\label{tab:principia_results}
\end{table}

\paragraph{Results} Across both backbones, we observe the same specialization trade-off: Dr.GRPO improves the initial-round reward but under-optimizes the aggregation round, while offline aggregation training (AggLM) does the opposite. Online multitask alleviates this mismatch by optimizing both stages jointly, and the pass@k variant achieves the strongest reward in both rounds, consistent with its goal of improving the quality of the candidate set consumed by aggregation.

\textbf{Qwen3-4B-Base.} \method{} yields the best post-aggregation performance overall, achieving the highest PrincipiaBench average (33.77) and the best competition-math average (23.72). It is strongest on all PrincipiaBench subsets (RealMath/Physics/ARB/SuperGPQA), and delivers the best AIME score (27.96), indicating broad improvements in scientific reasoning and hard math when both rounds are trained end-to-end.

\textbf{Qwen3-4B-Instruct-2507.} In this stronger, post-trained regime, online multitask attains the best PrincipiaBench average after aggregation (58.68), while \method{} remains competitive (58.07). In contrast, \method{} achieves the largest gains on competition math, reaching the best overall average (61.38) and the top AIME score (71.20). This suggests that pass@k-aware training is particularly effective for difficult math, where aggregation benefits most from a higher-quality, more solution-bearing candidate pool.

Overall, \method{} consistently improves end-to-end aggregation performance.

\subsection{Related Work}

\textbf{Parallel Inference Scaling} Recent works have explored scaling up test-time compute by generating multiple candidate solutions in parallel, moving beyond foundational majority-voting methods like Self-Consistency \citep{wang2022self}. Rather than simply selecting a single best output, modern approaches focus on dynamically selecting, compressing, or aggregating intermediate candidates to form a cohesive final response. One prominent direction is dynamic branching: \citet{pan2025learningadaptiveparallelreasoning} train models to adaptively spawn child threads for specific sub-tasks and join their results back into the main reasoning process. Similarly, \citet{zheng2025parallelr1parallelthinkingreinforcement} trigger multiple reasoning threads at critical steps to be merged into the main context. \citet{li2025parallelmuseagenticparallelthinking} adapt this parallel scaling to agentic contexts by branching partial rollouts only at highly uncertain steps, utilizing compressed reports during answer aggregation to maintain computational efficiency. To handle the massive context overhead of parallel scaling, \citet{pacore2025} utilize message passing to compress long reasoning trajectories, enabling test-time compute to scale to millions of tokens under a fixed context length. Alternatively, other works emphasize multi-round iterative refinement; \citet{venkatraman2025recursiveselfaggregationunlocksdeep} generate parallel solutions and aggregate them over successive rounds. In a closely related approach, \citet{madaan2025rethinkingthinkingtokensllms} employ a cyclical pipeline that generates parallel solutions, distills them into a condensed summary, and generates refined solutions conditioned on that summary. To ensure these iterative generation cycles do not amplify errors, \citet{sharma2026prismpushingfrontierdeep} leverage Process Reward Models to provide step-level verification during aggregation of the candidate population. Alternatively, \citet{singh2026v_1} proposes a pairwise self-verification method to select the single best solution from the generated pool, rather than aggregating a new solution. In this work, we utilize end-to-end reinforcement learning to train a model to operate within a minimal scaffolding of parallel generation and subsequent aggregation.

\textbf{Optimizing for Solution Variety in RL for LLMs} The effectiveness of aggregation frameworks is fundamentally tied to the diversity of the underlying candidate solutions. To encourage this during training, many recent works directly incorporate diversity metrics into the reinforcement learning reward signal, utilizing techniques such as embedding models \citep{zhou2026evolvinglanguagemodelslabels}, trained classifiers \citep{li2025jointlyreinforcingdiversityquality}, and LLM-based diversity judges \citep{hu2026rewardingrareuniquenessawarerl}. A particularly effective subset of these methods focuses on pass@k optimization \citep{tang2025optimizing, walder2025passkpolicyoptimizationsolving, peng2025simkosimplepasskpolicy}. Building on this direction, our work adopts the direct pass@k optimization strategy from \citet{chen2025passktrainingadaptivelybalancing}, which specifically up-weights the advantage of incorrect solutions to prevent mode collapse and expand the exploration space.

\subsection{Conclusion} Scaling test-time compute is only as effective as the diversity and quality of the reasoning paths that are explored. Traditional parallel decoding and self-aggregation methods are bottlenecked by off-policy generations and mode collapse. To overcome these limitations, we introduced \method{}, a unified online reinforcement learning framework that explicitly aligns and optimizes candidate generations with downstream aggregation.

Our core insight is that generation and aggregation require distinct but complementary optimization strategies. In \method{}, the generator actively explores a diverse, complementary set of solutions through pass@k optimization. Simultaneously, the aggregator is trained via pass@1 optimization to reliably synthesize the on-policy candidates into a final answer.

Extensive evaluations across competition math and scientific reasoning benchmarks validate the strength of this approach. In both base models (e.g., Qwen3-4B-Base) and strong post-trained reasoners (e.g. Qwen3-4B-Instruct-2507), \method{} consistently improves standard offline self-aggregation. The gains are particularly pronounced on highly complex tasks, such as AIME and Principia, where synthesizing diverse reasoning trajectories is critical. By co-training generation and aggregation end-to-end, \method{} provides a robust, scalable recipe for improving inference-time reasoning.

\bibliographystyle{assets/plainnat}
\bibliography{paper}

\clearpage
\newpage
\include{appendix}

\include{rllm_tables/rm_examples}
\include{rllm_tables/policy_examples}
\include{rllm_tables/j1_eval_examples}

\end{document}

%% file: rllm_tables/main1.tex
\begin{table}[]
\centering
\caption{{\bf Reference-free setting}: Comparison of different post-trained Qwen3-1.7B (Instruct) models using RLLM or RLHF on easy-to-verify and hard-to-verify reasoning benchmarks. All models are trained on hard-to-verify samples in a reference-free setting. RLHF'ed models are optimized using SOTA scalar RMs. RLLM models are optimized using either prompted LM-as-RM or our trained \methodrm{} LM-as-RM. We observe improved RLLM results by scaling up the LM-as-RM, with J1-Qwen3-32B-RM improving AIME24 by 12\% on top of a Qwen3-1.7B (Instruct) model.} 
\label{tab:main_rllm_vs_rlhf}
\setlength{\tabcolsep}{5pt}
\resizebox{\textwidth}{!}{
\begin{tabular}{lllcccccccc}
\toprule
\multirow{2}{*}{\sc Method} & \multirow{2}{*}{RM} & \multirow{2}{*}{RM Type} & \multirow{2}{*}{RM Size} & \multicolumn{6}{c}{\sc Math (Easy-to-Verify)} & {\sc Hard-to-Verify}                          \\ \cmidrule(lr){5-10}
& & & & \textbf{AIME24} & \textbf{AIME25} & \textbf{BRUMO25} & \textbf{HMMT24} & \textbf{HMMT25} & \textbf{Average} & \textbf{Physics} \\
\midrule 
\texttt{Qwen3-1.7B}& -- & -- & -- &	45.22 & 36.20 &	49.78 &	23.30 &	21.44 &	35.20 & 15.57 \\ \midrule
\rowcolor{gray!15}\multicolumn{11}{l}{\textit{\textbf{Policy Trained on hard-to-verify samples w/o Reference Answer}}} \\
RLHF & \texttt{Nexusflow/Athene-RM-8B} 
& Scalar & 8B	& 40.22	&32.92	&38.95	&17.08&	19.58&	29.75 & 13.18 \\
RLHF &  \texttt{Skywork-Reward-V2-Llama-3.1-8B}
& Scalar & 8B &	48.33&	36.26&	47.91	&22.92	&24.36	&35.96 & 16.93\\					
RLHF & \texttt{nvidia/AceMath-7B-RM}
& Scalar & 7B &
48.76& 38.34& 47.69& 23.12& 22.71& 36.12 & 16.24
\\
\rowcolor{mint} \textbf{RLLM} & \texttt{Llama-3.1-8B-Instruct} & Generative & 8B & 48.54 & 36.68 & 46.47 & 22.51 & 24.18 & 35.67 & 15.68 \\
\rowcolor{mint} \textbf{RLLM} & \texttt{Qwen3-1.7B} & Generative & 1.7B & 51.25 & 37.71 & 51.67 & 26.70 & 26.88 & 38.84 & 17.84 \\ 
\rowcolor{mint} \textbf{RLLM} & \texttt{Qwen3-32B} & Generative & 32B & 54.38 & 43.53 & 53.76 & 26.70 & \textbf{31.02} & 41.88 & 16.48 \\
\rowcolor{mint} \textbf{RLLM} & \texttt{GPT-OSS-120B} & Generative & 120B & 52.09 & 39.79 & 52.91 & 33.30 & 28.34 & 41.29 & 16.36 \\
\rowcolor{mint} \textbf{RLLM} & \texttt{J1-Qwen3-32B-RM} & Generative & 32B & \textbf{57.91} & \textbf{44.17} & \textbf{54.16} & \textbf{33.30} & 27.50 & \textbf{43.41} & \textbf{18.75} \\
\rowcolor{mint} \textbf{$\Delta$ w/ Qwen3-1.7B} & -- & -- & -- & +12.59 & +7.97 & +4.38 & +10.0 & +6.06 & +8.21 & +3.18  \\
                        \bottomrule
    \end{tabular}}
\end{table}

\begin{table}[t]
\centering
\caption{{\bf Reference-based setting}: Comparison of different post-trained Qwen3-1.7B (Instruct) models using RLLM or RLVR on easy-to-verify and hard-to-verify reasoning benchmarks. All models are trained on hard-to-verify examples in a reference-based setting. RLVR models are optimized using either rule-based or model-based verifiers. RLLM models are optimized using either prompted or trained LM-as-RM (functioning as reference-based verifiers). All RLLM variants outperform all RLVR variants.} 
\label{tab:main_rllm_vs_rlvr}
\setlength{\tabcolsep}{5pt}
\resizebox{\textwidth}{!}{
\begin{tabular}{llllccccccc}
\toprule
\multirow{2}{*}{\sc Method} & \multirow{2}{*}{Verifier} & Verifier & Verifier & \multicolumn{6}{c}{\sc Math (Easy-to-Verify)} & {\sc Hard-to-Verify}                          \\ \cmidrule(lr){5-10}
& & Type & Size & \textbf{AIME24} & \textbf{AIME25} & \textbf{BRUMO25} & \textbf{HMMT24} & \textbf{HMMT25} & \textbf{Average} & \textbf{Physics} \\
\midrule 
\texttt{Qwen3-1.7B}& -- & -- & -- &	45.22 & 36.20 &	49.78 &	23.30 &	21.44 &	35.20 & 15.57 \\ \midrule
\rowcolor{gray!15} \multicolumn{11}{l}{\textit{\textbf{Policy Trained on hard-to-verify samples w/ Reference Answer}}}\\
RLVR & Math-Verify
& Rule-based & -- & 48.96	&34.79	&47.51&	21.24	&25.83&	35.67 & 15.95 \\
RLVR & \texttt{TIGER-Lab/general-verifier} 
& Generative & 1.5B &50.42&	41.66	&49.38&	23.55	&23.33&	37.67 & 17.39\\
\rowcolor{mint} \textbf{RLLM} & \texttt{Qwen3-1.7B} & Generative & 1.7B & 52.50 & 41.46 & 49.59 & \textbf{33.30} & 26.66 & 40.70 & 16.93\\
\rowcolor{mint} \textbf{RLLM} & \texttt{Qwen3-32B} & Generative & 32B & \textbf{57.29} & 43.33 & 52.93 & 23.30 & 28.97 & 41.16 & 18.32 \\
\rowcolor{mint} \textbf{RLLM} & \texttt{J1-Qwen3-32B-Verifier} & Generative & 32B & 55.83 & \textbf{46.05} & \textbf{53.32} & 30.00 & \textbf{31.03} & \textbf{43.24} & \textbf{19.36} \\
\rowcolor{mint}\textbf{$\Delta$ w/ Qwen3-1.7B} & -- & -- & -- & +10.61 & +9.85 & +3.54 & +6.70 & +9.59 & +8.04 & +3.79  \\
\bottomrule
    \end{tabular}}
\end{table}

%% file: rllm_tables/main2.tex
\begin{table}[]
\centering
\caption{Comparison of RLLM, RLHF, and RLVR across different training datasets -- easy-to-verify, hard-to-verify, reference-free, and reference-based. RLLM on hard-to-verify data with a strong LM-as-RM outperforms all models trained on easy-to-verify data.} 
\label{tab:main_rllm_easy}
\setlength{\tabcolsep}{5pt}
\resizebox{\textwidth}{!}{
\begin{tabular}{lllccccccc}
\toprule
\multirow{2}{*}{\sc Method} & \multirow{2}{*}{RM/Verifier} & \multirow{2}{*}{RM Type} & \multirow{2}{*}{RM Size} & \multicolumn{6}{c}{\sc Math} \\ \cmidrule(lr){5-10}
& & & & \textbf{AIME24} & \textbf{AIME25} & \textbf{BRUMO25} & \textbf{HMMT24} & \textbf{HMMT25} & \textbf{Average} \\
\midrule 
\texttt{Qwen3-1.7B}& -- & -- & -- &	45.22 & 36.20 &	49.78 &	23.30 &	21.44 &	35.20\\ \midrule
\rowcolor{gray!15} \multicolumn{10}{l}{\textit{\textbf{Policy Trained on easy-to-verify samples w/o Reference Answer}}}\\ 
RLHF &  \texttt{Skywork-Reward-V2-Llama-3.1-8B} & Scalar & 8B & 48.96 & 35.21 & 43.96 & 33.30 & 25.20 & 37.33 \\
\textbf{RLLM} & \texttt{J1-Qwen-32B-RM} & Generative & 32B & 53.34 & 45.43 & 52.08 & 26.70 & 27.33 & 41.18 \\
\midrule
\rowcolor{gray!15} \multicolumn{10}{l}{\textit{\textbf{Policy Trained on easy-to-verify samples w/ Reference Answer}}}\\ 
RLVR & Math-Verify
& Rule-based & -- & 53.75 & 40.84 & 50.42 & 26.70 & 26.44 & 39.63\\
\textbf{RLLM} & \texttt{J1-Qwen-32B-Verifier} & Generative & 32B & 54.99 & 44.36 & 52.29 & 25.83 & 26.88 & 40.87 \\ \midrule
\rowcolor{gray!15} \multicolumn{10}{l}{\textit{\textbf{Policy Trained on hard-to-verify samples w/o Reference Answer}}}\\
RLHF &  \texttt{Skywork-Reward-V2-Llama-3.1-8B}
& Scalar & 8B &	48.33&	36.26&	47.91	&22.92	&24.36	&35.96 \\
\textbf{RLLM} & \texttt{J1-Qwen3-32B-RM} & Generative & 32B & \textbf{57.91} & 44.17 & \textbf{54.16} & \textbf{33.30} & 27.50 & \textbf{43.41} \\ \midrule
\rowcolor{gray!15} \multicolumn{10}{l}{\textit{\textbf{Policy Trained on hard-to-verify samples w/ Reference Answer}}} \\
RLVR & Math-Verify
& Rule-based & -- & 48.96	&34.79	&47.51&	21.24	&25.83&	35.67 \\
\textbf{RLLM} & \texttt{J1-Qwen-32B-Verifier} & Generative & 32B & 55.83 & \textbf{46.05} & 53.32 & 30.00 & \textbf{31.03} & 43.24 \\
\bottomrule
    \end{tabular}}
\end{table}

%% file: rllm_tables/non-verifiable.tex
\begin{table}[]
\centering
\caption{\label{tab:non_verifiable}Comparison of Win Rate (WR) and Length Controlled Win Rate (LCWR) of RLLM and RLHF on non-verifiable instruction-following tasks when training a Qwen3-1.7B policy (either in thinking or non-thinking mode). For AlpacaEval 2.0, we use GPT-4o as the evaluator and for ArenaHard 2.0, we use GPT-4.1 as the evaluator. RLLM matches or outperforms RLHF, obtaining best win rates on hard prompts of ArenaHard 2.0.} 
\resizebox{\textwidth}{!}{
\begin{tabular}{llllccc}
\toprule
\multirow{3}{*}{\sc Method} & \multirow{3}{*}{RM} & \multirow{3}{*}{RM Type} & \multirow{3}{*}{RM Size} & \multicolumn{3}{c}{\sc Datasets}                          \\ \cmidrule(lr){5-7}
& & & & {\textbf{AlpacaEval 2.0}} & \textbf{ArenaHard 2.0}  & \textbf{ArenaHard 2.0}  \\ 
& & & & (LCWR / WR) & (Hard Prompts) & (Creative Writing) \\
\midrule
\rowcolor{gray!15} \multicolumn{7}{l}{\textbf{Policy Trained on non-verifiable WildChat samples in thinking mode}} \\
\texttt{Qwen3-1.7B} & -- & -- & --& 37.5 / 39.2 & 6.4  (-0.6 / +0.7) & 6.9  (-1.0 / +1.1) \\
RLHF & \texttt{Nexusflow/Athene-RM-8B} & Scalar & 8B & 38.0 / 42.6 & 4.3  (-0.6 / +0.7) & \textbf{15.8 (-2.0 / +1.9)} \\
RLHF & \texttt{Skywork-Reward-V2-Llama-3.1-8B} & Scalar & 8B & 43.0 / 47.2 & 6.0  (-0.6 / +0.6) & 8.4  (-0.9 / +1.0)	\\					
\rowcolor{mint} \textbf{RLLM} & \texttt{Qwen3-1.7B} & Generative & 1.7B & \textbf{43.9 / 49.2} & \textbf{8.5  (-0.8 / +0.8)} & 14.7  (-1.4 / +1.8) \\ 
\midrule 
\rowcolor{gray!15} \multicolumn{7}{l}{\textbf{Policy Trained on non-verifiable WildChat samples in non-thinking mode}} \\
\texttt{Qwen3-1.7B} & -- & -- & -- & 27.8 / 30.0 & 3.1  (-0.6 / +0.4) & 3.1  (-0.9 / +0.7)\\
RLHF & \texttt{Nexusflow/Athene-RM-8B} & Scalar & 8B & 37.7 / 38.1 & 3.9  (-0.6 / +0.7) & 8.0  (-1.4 / +1.3) \\
RLHF & \texttt{Skywork-Reward-V2-Llama-3.1-8B} & Scalar & 8B & 34.9  / 40.0 & 3.6  (-0.8 / +0.6) & 7.8  (-1.2 / +1.2) \\
\rowcolor{mint} \textbf{RLLM} & \texttt{Qwen3-1.7B} & Generative & 1.7B & \textbf{41.4 / 43.7} & \textbf{6.0  (-0.8 / +0.7)} & \textbf{9.1  (-1.2 / +1.2)} \\
                        \bottomrule
\end{tabular}}
\end{table}

%% file: rllm_tables/ablation1.tex
\begin{table}[]
\centering
\caption{Analysis of Generator-Verifier Gap. RLLM post-training of a Qwen3-1.7B policy with a J1-Qwen3-1.7B LM-as-RM does not improve performance over the prompted LM-as-RM baseline while post-training with a stronger J1-Qwen3-32B LM-as-RM improves over the corresponding prompted baseline.} 
\label{tab:gen_ver_gap}
\setlength{\tabcolsep}{5pt}
\resizebox{\textwidth}{!}{
\begin{tabular}{lllcccccccc}
\toprule
\multirow{2}{*}{\sc Method} & \multirow{2}{*}{RM/Verifier} & \multirow{2}{*}{RM Type} & \multirow{2}{*}{RM Size} & \multicolumn{6}{c}{\sc Math (Easy-to-Verify)} & {\sc Hard-to-Verify}                          \\ \cmidrule(lr){5-10}
& & & & \textbf{AIME24} & \textbf{AIME25} & \textbf{BRUMO25} & \textbf{HMMT24} & \textbf{HMMT25} & \textbf{Average} & \textbf{Physics} \\ \midrule
\texttt{Qwen3-1.7B}& -- & -- & -- &	45.22 & 36.20 &	49.78 &	23.30 &	21.44 &	35.20 & 15.57 \\ \midrule
\rowcolor{gray!15} \multicolumn{11}{l}{\textit{\textbf{Policy Trained on hard-to-verify samples w/o Reference Answer}}}\\
\textbf{RLLM} & \texttt{Qwen3-1.7B} & Generative & 1.7B & 51.25 & 37.71 & 51.67 & 26.70 & 26.88 & \textbf{38.84} & \textbf{17.84} \\ 
\textbf{RLLM} & \texttt{J1-Qwen3-1.7B} & Generative & 1.7B & 52.29 & 39.59 & 45.63 & 30.00 & 21.46 & 37.79 & 15.11 \\ \midrule
\rowcolor{gray!15} \multicolumn{11}{l}{\textit{\textbf{Policy Trained on hard-to-verify samples w/o Reference Answer}}}\\
\textbf{RLLM} & \texttt{Qwen3-32B} & Generative & 32B & 54.38 & 43.53 & 53.76 & 26.70 & 31.02 & 41.88 & 16.48 \\
\textbf{RLLM} & \texttt{J1-Qwen3-32B-RM} & Generative & 32B & 57.91 & 44.17 & 54.16 & 33.30 & 27.50 & \textbf{43.41} & \textbf{18.75} \\
\bottomrule
\end{tabular}}
\end{table}

\begin{figure*}[]
    \centering

        \begin{subfigure}[l]{0.43\textwidth}
        \centering
        \includegraphics[width=\linewidth]{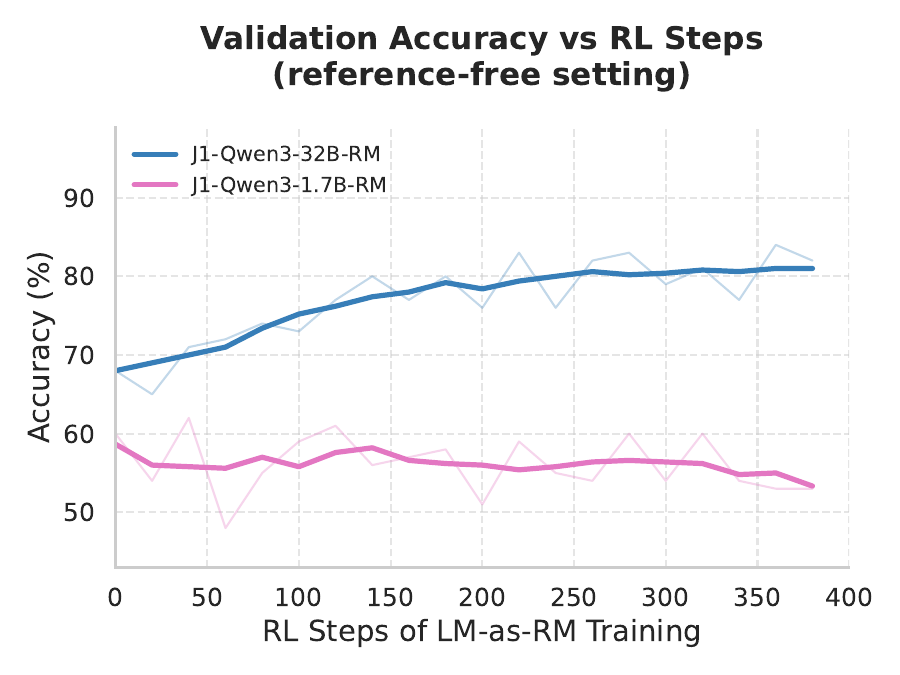}

    \end{subfigure}
    \hfill
        \begin{subfigure}[r]{0.54\textwidth}
    \centering
    \includegraphics[width=\linewidth]{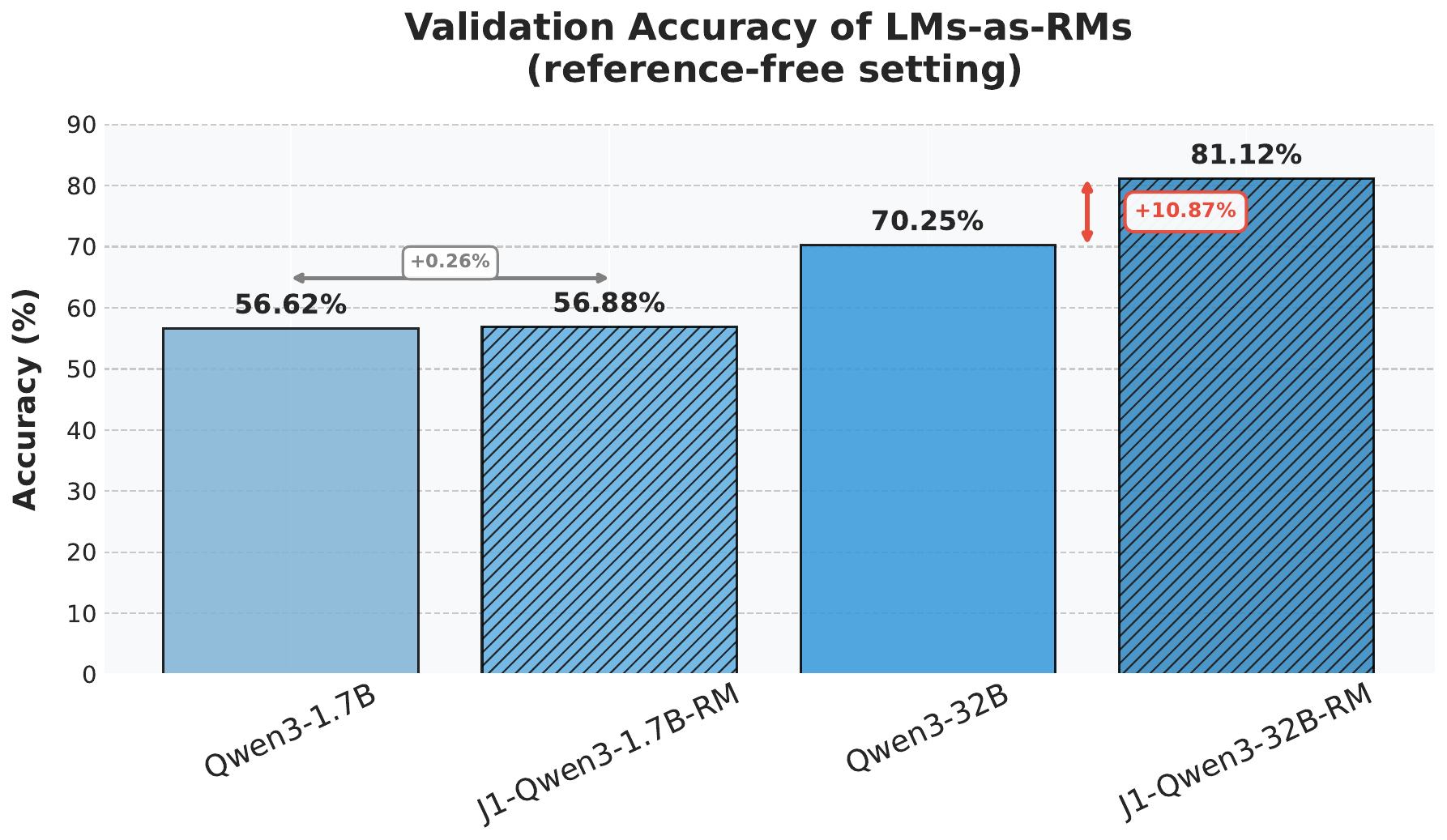}

    \end{subfigure}
    \caption{Analysis of Generator-Verifier Gap. (a) Comparison of different LMs-as-RMs in a reference-free setting on a held-out validation set (of correct/incorrect responses). \methodrm{} training on top of a weaker Qwen3-1.7B does not lead to further improvements, while the same on top of a stronger Qwen3-32B leads to 10\% absolute improvement. Results are averaged across 8 seeds. (b) Corresponding validation reward curves for \methodrm{} training across RL steps.}
    \label{fig:j1_acc_no_ref}

\end{figure*}

%% file: rllm_tables/ablation2.tex
\begin{table}[]
\centering
\caption{Comparison of RLLM post-training of Qwen3-1.7B with on-policy versus off-policy \methodrm{}-trained LMs-as-RMs. On-policy J1-Qwen3-32B-RM is trained on Qwen3-1.7B responses while off-policy models are trained on either weaker Llama responses or stronger Qwen3-8B responses. On-policy trained LM-as-RM outperforms off-policy trained ones.} 
\label{tab:on_vs_off_policy}
\setlength{\tabcolsep}{5pt}
\resizebox{\textwidth}{!}{
\begin{tabular}{lllcccccc}
\toprule
\multirow{2}{*}{\sc Method} & \multirow{2}{*}{RM/Verifier} & \multirow{2}{*}{RM Training Data} & \multicolumn{6}{c}{\sc Math}                          \\ \cmidrule(lr){4-9}
& & & \textbf{AIME24} & \textbf{AIME25} & \textbf{BRUMO25} & \textbf{HMMT24} & \textbf{HMMT25} & \textbf{Average} \\ \midrule
\texttt{Qwen3-1.7B}& -- & -- &	45.22 & 36.20 &	49.78 &	23.30 &	21.44 &	35.20 \\
\rowcolor{gray!15} \multicolumn{9}{l}{\textit{\textbf{Policy Trained on hard-to-verify samples w/o Reference Answer}}}\\
\textbf{RLLM} & \texttt{Qwen3-32B (prompted)} & -- & 54.38 & 43.53 & 53.76 & 26.70 & \textbf{31.02} & 41.88 \\
\textbf{RLLM} & \texttt{J1-Qwen3-32B-RM (off-policy-trained)} & Llama-3.1-8B-Instruct & 57.09 & 44.37 & 48.96 & 26.87 & 28.12 & 41.08 \\
\textbf{RLLM} & \texttt{J1-Qwen3-32B-RM (off-policy-trained)} & Qwen3-8B & 54.99 & \textbf{45.82} & 50.00 & 23.55 & 27.71 & 40.41 \\
\textbf{RLLM} & \texttt{J1-Qwen3-32B-RM (on-policy-trained)} & Qwen3-1.7B & \textbf{57.91} & 44.17 & \textbf{54.16} & \textbf{33.30} & 27.50 & \textbf{43.41} \\
\bottomrule
\end{tabular}}
\end{table}


%% file: rllm_tables/k-wise.tex
\begin{table}[t]
\centering
\caption{\label{tab:k-wise} Effect of scaling up reward modeling compute in RLLM via pointwise, pairwise, pairwise with a pivot rollout, and triplet-based scoring between rollouts.} 
\resizebox{\textwidth}{!}{
\begin{tabular}{lllcccc}
\toprule
\multirow{3}{*}{\sc Method} & \multirow{3}{*}{\sc RM} & \multirow{3}{*}{\sc RM Type} & \multirow{3}{*}{\sc \#RM Judgments} & \multicolumn{3}{c}{\sc Datasets}                          \\ \cmidrule(lr){5-7}
& & & & {\textbf{AlpacaEval 2.0}} & \textbf{ArenaHard 2.0}  & \textbf{ArenaHard 2.0}  \\ 
& & & & (LCWR / WR) & (Hard Prompts) & (Creative Writing) \\ \midrule
\texttt{Qwen3-1.7B} & -- & -- & -- & 37.5 / 39.2 & 6.4  (-0.6 / +0.7) & 6.9  (-1.0 / +1.1) \\
\midrule
\rowcolor{gray!15} \multicolumn{7}{l}{\textbf{Policy Trained on WildChat non-verifiable samples in thinking mode}} \\
\textbf{RLLM} & Qwen3-1.7B (prompted) & Pointwise & $n$ & 40.1 / 46.2 & 5.3  (-0.7 / +0.8) & 12.4  (-1.6 / +1.4) \\
\textbf{RLLM} & Qwen3-1.7B (prompted) & Pairwise with pivot & $2n$ & 41.0 / 42.8 & 6.3  (-0.8 / +0.8) & 13.9  (-1.3 / +1.2) \\
\textbf{RLLM} & Qwen3-1.7B (prompted) & Pairwise & $\binom{n}{2}$ & \textbf{43.9 / 49.2} & 8.5  (-0.8 / +0.8) & \textbf{14.7  (-1.4 / +1.8)} \\
\textbf{RLLM} & Qwen3-1.7B (prompted) & Triplet & $\binom{n}{3}$ & 42.0 / 48.1 &  \textbf{10.1  (-0.9 / +1.0)} & 13.1  (-1.4 / +1.8) \\
                        \bottomrule
\end{tabular}
}
\end{table}

%% file: appendix.tex
\begin{algorithm}[t]
\caption{\textcolor{black!80!black}{Conservative Equivalence Checking for Majority Voting (Principia Collection Construction)}}
\label{algorithm:majority_vote}
\small
\begin{algorithmic}[1]
\Require Predictions $A=\{a_1,\dots,a_n\}$; model-based verifier $\mathsf{V}$; agreement threshold $\tau=0.6$; majority threshold $\theta=\tfrac{5}{8}$
\State \textbf{Notation:} $\text{unique}(A)$ returns the set $U$ of distinct predictions in $A$; $|U|$ is the number of uniques; $|A|$ is the total predictions (e.g., $|A|{=}8$).

\Statex \vspace{2pt}\textbf{\color{blue!70!black}Pairwise Equivalence Checking}
\State $U \leftarrow \text{unique}(A)$; \quad $n_U \leftarrow |U|$
\For{$i \gets 1$ to $n_U$}
  \For{$j \gets i+1$ to $n_U$}
    \State evaluate $\mathsf{V}(u_i,u_j)\in\{\text{True},\text{False}\}$ and record in matrix $\mathsf{V}[u_i,u_j]$
  \EndFor
\EndFor

\Statex \vspace{2pt}\textbf{\color{cyan!70!black}Measuring Transitivity Among Elements}
\If{$n_U>2$}
  \ForAll{ordered pairs $(x,y)\in U\times U,\;x\neq y$}
    \State $\text{agree}(x,y)\gets\dfrac{1}{n_U-2}\sum_{w\in U\setminus\{x,y\}}\mathbf{1}\{\mathsf{V}(x,w)=\mathsf{V}(y,w)\}$
    \If{$\mathsf{V}(x,y)=\text{True}$ \textbf{and } $\text{agree}(x,y)<\tau$}
      \State $\mathsf{V}(x,y)\gets\text{False}$ \Comment{break weak links inconsistent with transitivity}
    \EndIf
  \EndFor
\EndIf

\Statex \vspace{2pt}\textbf{\color{pink!70!black}Grouping by Equivalence (DSU)}
\State \textbf{DSU (Disjoint-Set Union):} maintain $\mathrm{parent}[\cdot]$ with \textsc{Find} (path compression) and \textsc{Union}.
\State initialize DSU on $U$; \quad $\mathrm{cnt}[u]\gets|\{a\in A:\,a=u\}|$ \Comment{vote multiplicity per unique $u$}
\ForAll{$(x,y)$ with $\mathsf{V}(x,y)=\text{True}$}
  \State \textsc{Union}$(x,y)$ \Comment{``Merge $x,y$ in DSU''}
\EndFor
\State $\mathcal{G}\gets$ DSU groups via \textsc{Find}
\ForAll{$g\in\mathcal{G}$}
  \State $c(g)\gets\sum_{u\in g}\mathrm{cnt}[u]$ \Comment{class vote mass}
  \State $\mathrm{rep}(g)\gets
  \begin{cases}
    \arg\min_{u\in g}\text{len}(\text{string}(u)), & |g|\le 2\\[2pt]
    \text{element in $g$ whose length is closest to the median}, & |g|\ge 3
  \end{cases}$
\EndFor

\Statex \vspace{2pt}\textbf{\color{purple!70!black}Majority Decision (at least 5 of 8 votes)}
\If{$\mathcal{G}\neq\emptyset$}
  \State $g^\star\gets\arg\max_{g\in\mathcal{G}} c(g)$; \quad $a^\star\gets\mathrm{rep}(g^\star)$
  \If{$c(g^\star)\ge \lceil \theta\,|A|\rceil$}
    \State \Return $a^\star$ \textbf{(majority vote)}, $c(g^\star)$, refined map $\mathsf{V}$ \Comment{$a^\star$ is chosen if majority count is beyond threshold}
  \Else
    \State \Return \texttt{"[NO MAJORITY]"}, $c(g^\star)$, refined map $\mathsf{V}$
  \EndIf
\Else
  \State \Return \texttt{"[NO VALID PREDICTIONS]"}, $0$, refined map $\mathsf{V}$
\EndIf
\end{algorithmic}
\end{algorithm}




\section{Appendix}

\subsection{Contributions}

Chapter 1 contributors: Seungone Kim, Pranjal Aggarwal, Bo Liu, Swarnadeep Saha, Ping Yu, Anaelia Ovalle$^{\dagger}$, Jack Lanchantin, Jing Xu, Weizhe Yuan, Wenting Zhao, Adina Williams, Marjan Ghazvininejad, Graham Neubig$^{\dagger\dagger}$, Sean Welleck$^{\dagger\dagger}$, Jason Weston, Ilia Kulikov.

Chapter 2 contributors: Chenxi Whitehouse, Ilia Kulikov, Ping Yu, Jason Weston, Xian Li, Swarnadeep Saha.

Chapter 3 contributors: Tianjian Li, Jingyu Zhang, Ping Yu, Swarnadeep Saha, Sainbayar Sukhbaatar, Jason Weston, Ilia Kulikov, Jack Lanchantin.

All authors have Meta AI affiliation except: $^{\dagger}$: ULCA,  $^{\dagger\dagger}$: CMU.


\subsection{Hyperparameters for RL training}\label{appendix:hyper_parameter}

For training LMs using a model-based verifier, we load the policy model and the verifier model on separate nodes. For running inferences with the verifier model (determining answer equivalence), we use vLLM (v0.8.4) online inference and for training the policy model, we use verl (v0.4.0) with vLLM for rollouts. For inter-communications between the nodes (\textit{i.e.}, sending the verification results to verl), we use litellm (v1.72.2).

\subsection{Prompts}\label{appendix:prompts}

\subsubsection{Prompts for Training and Evaluating Principia Models}

\begin{promptbox}[Default system prompt for LMs trained on Principia Collection]
<instructions>
The user asks a question, and you should solve it as an intelligent assistant. You should first write the reasoning process step-by-step and then provide the user with the answer.
The reasoning process and answer should be enclosed within <reason> </reason> and <answer> </answer> tags, respectively (i.e., <reason> reasoning process here </reason> <answer> answer here </answer>). You should always start your response with <reason>, reason step-by-step, close your reasoning with </reason>, and then provide your final answer within <answer> and </answer> tags. Note that you should write <reason>, </reason>, <answer>, and </answer> only once in your response for clear separation of reasoning and answer. Most importantly, there should be no text before <reason> and after </answer> tags.
Between <answer> and </answer>, you should be concise and only provide the final prediction without any additional explanations (e.g., <answer> L = \\frac{1}{2} m \\dot{x}^2 \\left(1 + \\frac{4x^2}{a^2}\\right) - \\frac{mgx^2}{a} </answer>).
</instructions>
\end{promptbox}


\subsubsection{Prompts for Principia Collection Construction}

\begin{promptbox}[Prompt for Sketching Strategy Description (first step of Problem Statement Generation)]
### Task Description:
You are given a topic within mathematics or natural sciences.
Your task is to generate eight capabilities relevant to this topic as a list.
A capability is a concise, actionable description of a specific technique, method, or conceptual tool that an expert in the topic would use to solve problems within that area. Each capability should be:
    - Specific: Focus on a single, well-defined skill, method, or conceptual approach.
    - Action-oriented: Phrase each capability as something that can be done (e.g., "Applying X to compute Y", "Bounding Z using W", "Framing A in terms of B").
    - Relevant: Directly related to the topic and commonly used by researchers or advanced students in that field.
    - Non-trivial: Go beyond basic definitions or elementary facts; focus on techniques that require some expertise or insight.
    - Self-contained: Each capability should make sense on its own, without requiring reference to the others.
    - Non-overlapping: Each capability should not overlap with one another.

### Examples:
The following are some examples of a topic and corresponding capabilities:

(1) For the topic "Number theory >> Multiplicative number theory >> Distribution of integers with specified multiplicative constraints":
<start>
CAPABILITIES:
1. Framing prime conditions via Frobenius elements and splitting in number fields
2. Applying Chebotarev's density theorem to compute the density of primes with specified residue properties
3. Using the inclusion-exclusion principle to combine multiple conditions on primes (no smaller power residues)
4. Exploiting linear disjointness of field extensions to treat residue conditions independently ("quasi-multiplicative" densities)
5. Identifying and handling exceptional small primes that violate independence assumptions (special correlation cases)
6. Deriving explicit formulas for the density of primes satisfying complex power-residue conditions
7. Bounding tail contributions of infinite series using number-theoretic estimates (e.g., Euler's totient function growth)
8. Estimating and controlling error terms to ensure the final computed density is accurate to the required precision
<end>

(2) For the topic "Algebraic geometry >> Curves in algebraic geometry >> Coverings of curves, fundamental group":
<start>
CAPABILITIES:
1. Analyzing polynomial coverings via monodromy groups (permutation of 19 sheets) in algebraic geometry
2. Modeling irreducibility as transitivity of the monodromy action on fibered points
3. Utilizing branch loci and monodromy representations to encode the polynomial's structure
4. Applying Burnside's lemma and Sylow's theorem to deduce properties of a prime-order cycle in the permutation group
5. Establishing the normality of the 19-cycle subgroup to constrain the group's orbit structure (number of orbits)
6. Using the Riemann-Hurwitz formula to relate branch points to cycle structure and bound the number of components
7. Deducing the cycle decomposition (e.g., 9 disjoint transpositions) from group-orbit analysis to determine irreducible components
8. Identifying the degree-19 Chebysshev polynomial (via the Riemann existence theorem) as the unique polynomial with the required monodromy, and adjusting it to satisfy oddness, monicity, and coefficient conditions
<end>

(3) For the topic "Number theory >> Sequences and sets >> Recurrences":
<start>
CAPABILITIES:
1. Solving the linear recurrence through its characteristic polynomial and eigen values (roots of the characteristic equation)
2. Writing an explicit formula for sequence terms using the contributions of each characteristic root
3. Analyzing the p-adic valuation of each characteristic root (condition $|\\text{root}|_p < 1$) to determine viability for continuous extension
4. Using divisibility conditions in the recurrence values to pinpoint the prime that permits a continuous extension (narrowing down candidates)
5. Reducing the characteristic polynomial modulo p to examine how roots behave in $\\mathbb{F}_p$ (e.g., whether they become repeated or roots of unity)
6. Employing p-adic exponential and logarithm to define $a^x$ for non-integer $x$, enabling extension of the sequence to $\\mathbb{Z}_p$
7. Ensuring convergence of the power-series definition (requiring $|\\log(a)|_p < 1$ or $|a-1|_p < 1$ for each root $a$) so that the extension is well-defined and continuous
8. Considering multiple potential branches (difference choices of $p$-adic log for roots, especially when roots involve roots of unity) and requiring a unique branch to achieve a single continuous extension
<end>

(4) For the topic "Group theory and generalizations >> Representation theory of groups >> Representations of finite symmetric groups":
<start>
CAPABILITIES:
1. Recognizing that the given matrix tuple relations correspond to the Coxeter (braid) relations of a symmetric group
2. Reformulating the classification of such matrix tuples as the classification of representations of $S_{n}$ (symmetric group) of a given dimension
3. Noting that $GL(n)$-orbits of these tuples correspond ot isomorphism classes of $n$-dimensional representations of the group
4. Using the character table of the symmetric group to obtain all irreducible representation dimensions available
5. Accounting for the fact that each irreducible dimension (for non-self-conjugate partitions) appears in two distinct irreps (partition and its conjugate)
6. Formulating the counting of $n$-dimensional representations as a partitioning of $n$ into the multiste of irreducible dimensions (with unlimited repeats allowed)
7. Setting up a generating function whose coefficients count the number of ways to sum irreducible dimensions to $n$ (each irreducible dimension contributing a factor)
8. Extracting the coefficient of $x^n$ in the power series expansion to obtain the number of distinct $n$-dimensional representation types (hence orbits)
<end>

(5) For the topic "Number theory >> Arithmetic algebraic geometry (Diophantine geometry) >> Curves over finite and local fields":
<start>
CAPABILITIES:
1. Interpreting the homogeneous equation as a projective algebraic curve over the finite field $\\mathbb{F}{p^k}$ and translating the question into counting $\\mathbb{F}{p^k}$-rational points
2. Applying the Hasse-Weil bound to constrain the possible number of points on the curve (using genus or other invariants to get an upper bound)
3. Recognizing if the curve is maximal (attains the Hasse-Weil upper bound), which would directly determine the number of points
4. Utilizing symmetry in the curve's equation (cyclic permutation of $x,y,z$) to reduce and simplify the point-counting process
5. Incorporating finite field extension techniques, such as constructing $\\mathbb{F}_{p^k}$ via an irreducible polynomial, to systematically enumerate solutions in the extension field
6. Accounting for the "up to scaling" condition by counting projective points (excluding the trivial zero vector and identifying scalar multiples as one)
7. Breaking the counting problem into cases (e.g., whether a coordinate is zero) and using inclusion-exclusion to avoid overcounting across those cases
8. Verifying the final count against theoretical bounds or known results forconsistency, and adjusting if any deviations are detected (ensuring the count respects the Hasse-Weil limits)
<end>

### Format Requirements:
(1) Before writing the list of capabilities, you should think step by step of what would constitute a good set of lists for the given topic.
(2) After you think step by step, you should write "<start>" to denote that you will generate the list of capabilites. Your generation will be considered invalid if you do not generate any thinking before "<start>".
(3) There should exactly be 8 capabilities, where each capability starts with "1. ", "2. ", "3. ", "4. ", "5. ", "6. ", "7. ", "8. ". If not, your generation will be treated as invalid.
(4) After you are done writing the list of capabilities, write "<end>" to denote that you are done generating the list of capabilities.
(5) You should not include any explanations or other text between "<start>" and "<end>". Also,  "<start>" and "<end>" should occur only once in your generation (i.e., when you are proposing the list of capabilities after your thinking). If not, your generation will be treated as invalid.

### Guidelines for Thinking before Proposing the List of Capabilities:
You should think step by step before actually writing the list of capabilities. Here are some guidelines:
    - List the main entities: Identify all relevant objects (e.g., particles, fields, operators, manifolds, equations, symmetries, primes, groups, functions, etc.).
    - Describe core and challenging problems: What are the most important, subtle, or technically difficult questions in this area? (e.g., proving existence/uniqueness, classifying solutions, computing observables, determining distributions, understanding symmetries, etc.)
    - Highlight sources of complexity: What makes these problems hard? (e.g., nonlinearity, high dimensionality, coupling between variables, deep connections to other areas, etc.)
    - List foundational results and principles: What are the essential theorems, laws, or principles? (e.g., Noether's theorem, conservation laws, Chebotarev's theorem, variational principles, gauge invariance, etc.)
    - General strategies: What are the typical approaches to analyze or manipulate these objects? (e.g., perturbation theory, symmetry reduction, use of generating functions, field extensions, variational calculus, etc.)
    - Specialized or clever techniques: Are there unique tricks, constructions, or methods specific to this topic? (e.g., Feynman diagrams, explicit class field theory, advanced inclusion-exclusion, Monte Carlo simulations, etc.)
    - Reduction steps: How do experts break down complex problems? (e.g., decomposing into simpler subsystems, using orthogonality, modular reduction, separation of variables, etc.)
    - Combining multiple conditions: What methods are used to handle simultaneous constraints or coupled systems? (e.g., inclusion-exclusion, Lagrange multipliers, coupled equations, etc.)
    - Exceptional or pathological cases: Are there special values, singularities, or degenerate cases that require separate treatment?
    - Failure of naive approaches: Where do standard methods break down, and why? (e.g., non-perturbative effects, lack of independence, divergences, etc.)
    - Technical obstacles: What are the hardest parts of the analysis? (e.g., bounding error terms, ensuring convergence, handling infinities, controlling chaos, etc.)
Always note that it is better if both your thinking and list of capabilities are as detailed as possible, including specific equations or cases in context.

### Topic:
Now it is your turn to propose eight capabilities relevant to the topic:
[HERE_IS_THE_TOPIC]
\end{promptbox}

\begin{promptbox}[Prompt for Generating Initial Problem Statement (second step of Problem Statement Generation]
### Task Description:
You are given a topic in mathematics or the natural sciences, a description of a capability, and a required answer format.
Your task is to generate a question that is relevant to the given topic, requires the specified capability in order to solve the question, and must be answered in the specified answer format.

### Examples:
Below are some examples that include a topic, a description of a capability, an answer format, and a corresponding question. Especially, check that the questions are not concept-checking question, but instead reasoning-intensive ones:

(1) For the topic "Quantum Mechanics >> Quantum systems with discrete spectra >> Time-independent perturbation theory", the capability "Identifying that |001 decouples (zero matrix elements) because $x$ and $y$ cannot connect $z$-excited states", and answer type 'Equation: A mathematical statement that includes an equals sign (=), expressing the equality of two expressions (e.g., H = \\frac{1}{2m} \\left( \\mathbf{p} - \\frac{q}{c} \\mathbf{A} \\right)^2 + q \\phi).':
<question>
The first excited state of three-dimensional isotropic harmonic oscillator (of natural angular frequency $\\omega_0$ and mass $m$) is three-fold degenerate. Use the perturbation method to calculate the splitting (to the first order) of this three-fold degenerate state due to a small perturbation of the form $H' = bxy$, where $b$ is a constant. Give the first-order wave functions of the three split levels in terms of the wave functions of the unperturbed three-dimensional harmonic oscillator, given that, for a one-dimensional harmonic oscillator,\n\n$$\n\\langle n|x|n + 1 \\rangle = \\sqrt{\\frac{(n + 1)\\hbar}{2m\\omega_0}}.$$
</question>

(2) For the topic "Group theory and generalizations >> Special aspects of infinite or finite groups >> Braid groups; Artin groups", the capability "Employing the determinant condition to further constrain the matrix variables, typically reducing the system to a single or a few key variables", and answer type 'Piecewise Function: A function that is defined by different expressions or formulas over different intervals of its domain, with each piece applying to a specific part of the domain (e.g., f(t) = \\begin{cases} 1 - |t|, & \text{if } 0 \leq t < 2 \\ t - 3, & \text{if } 2 \leq t < 4 \\end{cases}). It should use symbols such as "\\begin{cases}", "\\end{cases}", "if", "otherwise"':
<question>
Consider the braid group ( B_3 ), the group of braids on three strands, with the standard presentation given by generators ( \sigma_1, \sigma_2 ) and the relation ( \sigma_1 \sigma_2 \sigma_1 = \sigma_2 \sigma_1 \sigma_2 ). The Burau representation of ( B_3 ) is a homomorphism ( \rho: B_3 \to \text{GL}_2(\mathbb{Z}[t, t^{-1}]) ), where ( \sigma_1 ) and ( \sigma_2 ) are mapped to the matrices:

[ \rho(\sigma_1) = \begin{pmatrix} -t & 1 \ 0 & 1 \end{pmatrix}, \quad \rho(\sigma_2) = \begin{pmatrix} 1 & 0 \ t & -t \end{pmatrix}. ]

For a real parameter ( t \geq 0 ), consider the matrix ( M(t) = \rho(\sigma_1 \sigma_2 \sigma_1) ), the image of the braid word ( \sigma_1 \sigma_2 \sigma_1 ) under the Burau representation. Define the function ( f(t) ) as the sum of the absolute values of the eigenvalues of ( M(t) ), where the eigenvalues are computed over the field ( \mathbb{C} ) (treating ( t ) as a real number).

Determine the function ( f(t) ) explicitly as a piecewise function for ( t \geq 0 ).
</question>

(3) For the topic "Numerical analysis >> Numerical linear algebra >> Numerical solutions to overdetermined systems, pseudoinverses", the capability "Formulating a regression as (\min_x\frac12|Ax-b|^2), i.e.\ building the design matrix (A), response vector (b), parameter vector (x)", and answer type 'Matrix: A rectangular array of numbers, symbols, or expressions, arranged in rows and columns, that is used to represent and manipulate data or mathematical relationships (e.g., $f(\\begin{bmatrix} x \\\\ y \\end {bmatrix}) = S + \\frac{\\|G-S \\|}{\\|F-S\\|}R (\\begin{bmatrix} x \\\\ y \\end {bmatrix} - S)$). It should be formulated in latex-compatible format, using symbols such as "\\begin{pmatrix}", "\\end{pmatrix}". This also includes vectors, which can be represented as a single-column or single-row matrix.':
<question>
Formulate the data fitting problem as a least squares problem $$\underset{x\in\mathbb{R}^{n}}{\min}\ f(x)=\frac{1}{2} ||Ax-b||_{2}^{2}$$ for the model $$\hat{y}(\theta;t)=\theta_{1}+\frac{\theta_{2}}{t+1.0}+\theta_3 t^{3}$$ and measurement equation $$y_i = \hat{y}(\theta;t_i)+e_i,\ \ \ i=1,2,...,m$$ where $e_i \sim N(0, \sigma^{2})$. Write expressions for $A,b,x$ and $n$.
</question>

### Format Requirements:
(1) Before writing the question, you should think step by step to sketch how the pquestion should be formulated as.
(2) After you think step by step, you should write "<question>" to denote that you will generate the question. Your generation will be considered invalid if you do not generate any thinking before "<question>".
(3) The question should not be a compositional problem. In other words, it shouldn't be consisted of multiple sub-problems, but instead, require only a single decisive answer based on the requested answer format.
(4) After you are done writing the list of capabilities, write "</question>" to denote that you are done generating the list of capabilities.
(5) You should not include any explanations or other text between "<question>" and "</question>". Also, "<question>" and "</question>" should occur only once in your generation (i.e., when you are proposing the question after your thinking). If not, your generation will be treated as invalid.

### Properties of a Good Question:
Here are the guidelines that defines what a good question is:
(1) The question should be reasoning-intensive instead of a concept-checking question: Assume that the solver already has profound knowledge of the provided topic. The question should be challenging enough for the solver. Specifically, if the question simply requires standard calculation, combining well-known materials, and does not implicitly require any tricks or challenges to solve the question, the question should be considered as a very bad question. The question should require to reason deeply in order to derive a decisive answer.
(2) The question should be not ill-defined: The question should provide only necessary conditions or information to start reasoning.
(3) The question should not be consisted of multiple sub-problems: Instead of asking for multiple answers, the question should naturally require going through multiple steps implicitly in order to derive the final answer.
(4) The question should not require looking at images.
(5) Don't make the problem too simple by providing hints: The question should not explicitly provide the idea of how to solve the problem and it also shouldn't provide too much details that could serve as a hint. Also, do not provide too much details on the answer format (e.g., how it should be structured).
(6) The question should be grounded to the provided materials: The question should be relevant to the provided topic, require the provided capability in order to solve it, and require to output the provided answer format.

### Guidelines for Thinking before Proposing the Question:
You should think step by step before actually writing the question. Here are some guidelines:
    - Sketch multiple possible candidate of problems: The problems you provide should be relevant to the given topic while inherently requiring the provided capability. While doing this, note that the questions should ask to output the given answer format.
    - Attempt to solve the questions you wrote: This will give you a sense of which problems are either too easy, which are too hard, and which are ill-defined (e.g., setting is too synthetic or the requirement is not clear). A good problem should require to think step by step in order to derive the final answer.
    - Ensure Uniqueness and Non-Triviality: Make sure the question is not a direct copy of standard textbook examples, and that it requires more than rote memorization or a single mechanical step.
    - Check for Ambiguity: Ensure that all terms, variables, and requirements are clearly defined so that the problem cannot be misinterpreted.
    - Balance Between Simplicity and Depth: The problem should not be so simple that it can be answered without any real thought, nor so complex that it becomes overwhelming or distracts from the core concept.
    - Consider Edge Cases: Think about whether the problem could be improved by including or excluding special cases (e.g., repeated $x$ values, missing data, etc.), but avoid unnecessary complications unless they serve a pedagogical purpose.
    - Revise or filter out to choose the final question: Among the candidates you thought of, you could attempt to add more conditions, change the values or expressions to deliberately make the problem of high-quality, or discard questions that you think are not a good question based on your thinking.
    - Final check: Again, ensure that the question you propose should be relevant to the given topic, require the specified capability, and must be answered in the specified format.
Always note that it is better if your thinking is as detailed as possible, while the question itself shouldn't be overly verbose.

### Question Proposal:
Now it is your turn to propose a question based on the following inputs:

Topic: [HERE_IS_THE_TOPIC]
Description of the capability: [HERE_IS_THE_DOMAIN]
Answer Format: [HERE_IS_THE_ANSWER_TYPE]
\end{promptbox}

\begin{promptbox}[Prompt for Revising Initial Problem Statement (last step of Problem Statement Generation]
### Task Description:
You are given a simple question, a topic in mathematics or the natural sciences, a description of a capability, and a required answer format.
It is very likely that the question is straightforward and textbook-level question for experts who has a lot of knowledge in this domain.
Your task is to propose a harder variant of the question that even experts might need to think for a long time, based on the provided materials. Specifically, the question should be relevant to the given topic and original question, require the specified capability while reasoning to solve the question, and must be answered in the specified answer format.

### Examples:
Below are some examples of challenging questions. These are not concept-checking question, but instead requires extensive reasoning:

(1) 
<question>
The first excited state of three-dimensional isotropic harmonic oscillator (of natural angular frequency $\\omega_0$ and mass $m$) is three-fold degenerate. Use the perturbation method to calculate the splitting (to the first order) of this three-fold degenerate state due to a small perturbation of the form $H' = bxy$, where $b$ is a constant. Give the first-order wave functions of the three split levels in terms of the wave functions of the unperturbed three-dimensional harmonic oscillator, given that, for a one-dimensional harmonic oscillator,\n\n$$\n\\langle n|x|n + 1 \\rangle = \\sqrt{\\frac{(n + 1)\\hbar}{2m\\omega_0}}.$$
</question>

(2) 
<question>
Consider the braid group ( B_3 ), the group of braids on three strands, with the standard presentation given by generators ( \sigma_1, \sigma_2 ) and the relation ( \sigma_1 \sigma_2 \sigma_1 = \sigma_2 \sigma_1 \sigma_2 ). The Burau representation of ( B_3 ) is a homomorphism ( \rho: B_3 \to \text{GL}_2(\mathbb{Z}[t, t^{-1}]) ), where ( \sigma_1 ) and ( \sigma_2 ) are mapped to the matrices:

[ \rho(\sigma_1) = \begin{pmatrix} -t & 1 \ 0 & 1 \end{pmatrix}, \quad \rho(\sigma_2) = \begin{pmatrix} 1 & 0 \ t & -t \end{pmatrix}. ]

For a real parameter ( t \geq 0 ), consider the matrix ( M(t) = \rho(\sigma_1 \sigma_2 \sigma_1) ), the image of the braid word ( \sigma_1 \sigma_2 \sigma_1 ) under the Burau representation. Define the function ( f(t) ) as the sum of the absolute values of the eigenvalues of ( M(t) ), where the eigenvalues are computed over the field ( \mathbb{C} ) (treating ( t ) as a real number).

Determine the function ( f(t) ) explicitly as a piecewise function for ( t \geq 0 ).
</question>

(3)
<question>
Let $$a+bi\\in\\mathbb{Z}[i]\\setminus\\{0\\}$$ and suppose that $$2^j\\parallel (a+bi)$$ (i.e. $$2^j$$ is the exact power of 2 dividing $$a+bi$$). Also, let $$n$$ be the smallest integer such that\n\\[\n\\max\\left(\\left|\\frac{a}{2^j}\\right|,\\;\\left|\\frac{b}{2^j}\\right|\\right)+2\\le w_n. \n\\]\nDetermine the value of\n\\[\n\\phi_{\\mathbb{Z}[i]}(a+bi)\n\\]\nin terms of $$n$$ and $$j$$, depending on whether\n\\[\n\\left|\\frac{a}{2^j}\\right|+\\left|\\frac{b}{2^j}\\right|+3\\le w_{n+1}\n\\]\nholds or not. Please provide your solution in the form of a piecewise function using \"\\begin{cases} ... \\end{cases}\".
</question>

(4)
<question>
Assume that the potential between two point charges \\( q_{1} \\) and \\( q_{2} \\) is not the Coulomb potential \\( A q_{1} q_{2} / r \\), but instead the Yukawa potential \\( A q_{1} q_{2} \\mathrm{e}^{-K r} / r \\), where \\( A \\) and \\( K \\) are constants, and \\( K=0^{+} \\). Under this condition, determine how the Poisson equation for electrostatic potential would be modified. Please provide the your solution in the form of an equation.
</question>

(5)
<question>
Suppose there are two kinds of E. coli (bacteria), ``red'' ones and ``green'' ones. Each reproduces faithfully (no sex) by splitting into half, red $\\rightarrow$ red + red or green $\\rightarrow$ green + green, with a reproduction time of 1 hour. Other than the markers ``red'' and ``green'', there are no differences between them. A colony of 5,000 ``red'' and 5,000 ``green'' E. coli is allowed to eat and reproduce. In order to keep the colony size down, a predator is introduced which keeps the colony size at 10,000 by eating (at random) bacteria.\nAfter a very long time, what is the probability distribution of the number of red bacteria? Please provide the your solution in the form of an equation.
</question>

(6)
<question>
"$\\text{max} f = 3x_1 + 4x_2 + 4x_3$\n$$\ns.t.\\begin{cases}\n6x_1 + 3x_2 + 4x_3 \\leq [1200, 100] \\\\\n5x_1 + 4x_2 + 5x_3 \\leq [1550, 200] \\\\\nx_1, x_2, x_3 \\geq 0\n\\end{cases}\n$$\nWhat are the fuzzy optimal solution and fuzzy optimal value?
</question>

### Format Requirements:
(1) Before writing the question, you should think step by step to sketch how the question should be formulated as.
(2) After you think step by step, you should write "<question>" to denote that you will generate the question. Your generation will be considered invalid if you do not generate any thinking before "<question>".
(3) The question should not be a compositional problem. In other words, it shouldn't be consisted of multiple sub-problems, but instead, require only a single decisive answer based on the requested answer format.
(4) After you are done writing the list of capabilities, write "</question>" to denote that you are done generating the list of capabilities.
(5) You should not include any explanations or other text between "<question>" and "</question>". Also, "<question>" and "</question>" should APPEAR ONLY ONCE in your generation (i.e., when you are proposing the question after your thinking). If not, your generation will be treated as invalid.

### Properties of a Good Question:
Here are the guidelines that defines what a good question is:
(1) The question should be reasoning-intensive instead of a concept-checking question: Assume that the solver already has profound knowledge of the provided topic. The question should be challenging enough for the solver. Specifically, if the question simply requires standard calculation, combining well-known materials, and does not implicitly require any tricks or challenges to solve the question, the question should be considered as a very bad question. The question should require to reason deeply in order to derive a decisive answer.
(2) The question should be not ill-defined: The question should provide only necessary conditions or information to start reasoning.
(3) The question should not be consisted of multiple sub-problems: Instead of asking for multiple answers, the question should naturally require going through multiple steps implicitly in order to derive the final answer.
(4) The question should not require looking at images.
(5) Don't make the problem too simple by providing hints: The question should not explicitly provide the idea of how to solve the problem and it also shouldn't provide too much details that could serve as a hint. Also, do not provide too much details on the answer format (e.g., how it should be structured).
(6) The question should be grounded to the provided materials: The question should be relevant to the provided topic, require the provided capability in order to solve it, and require to output the provided answer format. Yet, it is important that the the capability is not explicitly mentioned in the revised question.

### Guidelines for Thinking to Make the Question Harder:
You should deliberately increase the complexity and depth of the question so that it cannot be solved by a single mechanical step or by recalling a standard example. Here are some guidelines:
    - Introduce Non-Obvious Constraints or Twists: Add conditions or constraints that are not immediately apparent or that force the solver to reconsider their initial approach. For example, require the answer to satisfy an additional property, or introduce a subtle exception.
    - Incorporate Realistic or Contextual Elements: Frame the problem in a context that requires interpretation or modeling, rather than just computation. This could involve a real-world scenario, or data that must be interpreted before applying mathematical tools.
    - Consider Edge Cases and Exceptions: Deliberately include or highlight edge cases (implicitly, not explicitly!) that might trip up a solver who is not thinking carefully, but avoid making the problem artificially tricky.
    - Avoid Direct Imitation of Textbook Problems: Ensure that the question is not a direct copy or minor variation of a standard example. Instead, combine elements from different problems, or introduce a novel twist.
    - Test Your Own Problem: Attempt to solve the question yourself. If you find that it requires you to pause, plan, and reason through multiple steps, it is likely at the right level. If it feels too direct, add another layer of complexity.
    - Final Review for Clarity and Depth: Make sure the problem is still clearly stated, with all terms and requirements defined, but that it cannot be solved without genuine reasoning. The question should be challenging but fair, and the answer format should be specified.
Always note that it is better if your thinking to propose a harder question is as detailed as possible. Also, don't mention the "<question>" in your thinking before actually proposing the new question.

### Question Proposal:
Now it is your turn to propose a harder question based on the following inputs:

Topic: [HERE_IS_THE_TOPIC]
Description of the capability: [HERE_IS_THE_DOMAIN]
Answer Format: [HERE_IS_THE_ANSWER_TYPE]
Original Question: [HERE_IS_THE_ORIGINAL_QUESTION]
\end{promptbox}

\begin{promptbox}[Prompt for Filtering out Invalid Problem Statements]
### Task Description:
You are given a question, a topic in mathematics or the natural sciences, and a required answer format.
Your task is to check if the question is a valid or not. Read through the following instructions and answer in either "This is problematic" or "Looks good to me":

### Guidelines for determining the validity of a question:
1. First, determine if the question requires to output the answer type provided below. It may or may not explicitly mention to generate in such answer type, what you should determine is if it is a totally irrelevant question or not. Additionally, questions that do not ask to provide an answer and only provides statements or conditions should be considered as invalid.
2. Next, determine if the question is a compositional question or not. A compositional question refers to a question that is consisted of multiple sub-questions or a question that asks to provide at least two answers at once. This differs with a question that asks to provide a single answer which is composed of multiple parts. The rule of thumb is that if it asks to provide more than two answers in the question, it should be determined as a compositional question.
3. Lastly, determine if the answer is accidentally included within the problem or not. This also includes explicitly providing guidelines of how to solve the problem, which will make the problem overly easy.

### Format Requirements:
(1) After you think step by step, you should write "<decision>" to denote that you will provide your decision of whether the question is valid or not.
(2) Write "This is problematic" if there is at least one issue with the question and "Looks good to me" if there are none.
(3) After providing your decision, write "</decision>" to denote that you are done generating the list of capabilities.
(4) All together, you should write either "<decision> This is problematic </decision>" or "<decision> Looks good to me </decision>".

### Question:
Now it is your turn to propose a question based on the following inputs:

Question: [HERE_IS_THE_QUESTION]
Answer Format: [HERE_IS_THE_ANSWER_TYPE]
\end{promptbox}

\clearpage

\section{Prompt Templates}
\label{sec:prompt}
\autoref{fig:pointwise}, \autoref{fig:pairwise}, and \autoref{fig:ref} show the prompt templates for training pointwise reference-free, pairwise, and pointwise reference-based LMs-as-RMs, respectively.

\begin{figure*}[t]
    \centering
    \begin{prompt}{Prompt Template for Pointwise Reference-free LMs-as-RMs}

\texttt{You are given a user question and a response from an AI assistant. Your task is to act as an impartial judge and evaluate how well the response fulfills the user's instructions. You will be shown multiple responses to the same prompt, but only one at a time. Evaluate each response independently.}\\

\texttt{Think carefully about how to assess the quality of the response and assign the assistant's response a score 1 if the response is correct, and 0 if not. Enclose the score within <score> and </score> tags.}\\

\texttt{Format your output like this:\\
<think> your\_thinking\_process </think>  \\
<score> 0 or 1 </score>}\\

\texttt{Below are the user’s question and the assistant’s response:\\
\\
~[User Question]\\
~~{\color{denim}{\{instruction\}}}\\
\\
~[The Start of the Assistant's Answer]\\
~~{\color{denim}{\{response\}}}\\
~[The End of the Assistant's Answer]}
\end{prompt}
\vspace{-0.5em}
    \caption{{Prompt template for pointwise reference-free LMs-as-RMs}
    \label{fig:pointwise}
    }
\end{figure*}

\begin{figure*}[h!]
    \centering
    \begin{prompt}{Prompt Template for Pairwise LMs-as-RMs for Non-verifiable Tasks.}
\texttt{You are given a user question and two responses from two AI assistants. Your task is to act as an impartial judge and evaluate which response better follows the user's instructions and provides a higher-quality answer. Avoid any position biases and ensure that the order in which the responses were presented does not influence your decision. Do not allow the length of the responses to influence your evaluation. Do not favor certain names of the assistants. Be as objective as possible.}\\

\texttt{Think carefully about how to assess the quality of the responses and assign each response a score from 0 to 10, using either an integer or a decimal with up to 0.1 precision, with a higher score indicating a higher-quality response that better satisfies the criteria. Enclose the scores within the tags <score\_A> </score\_A>, and <score\_B> </score\_B>.}\\

\texttt{Format your output like this:\\
<think> your\_thinking\_process </think>  \\
<score\_A> your\_score\_a </score\_A> <score\_B> your\_score\_b </score\_B>
}\\

\texttt{Below are the user’s question and the two responses:\\
\\
~[User Question]\\
~~{\color{denim}{\{instruction\}}}\\
\\
~[The Start of Assistant A's Answer]\\
~~{\color{denim}{\{response A\}}}\\
~[The End of Assistant A's Answer]\\
\\
~[The Start of Assistant B's Answer]\\
~~{\color{denim}{\{response B\}}}\\
~[The End of Assistant B's Answer]}
\end{prompt}
\vspace{-0.5em}
    \caption{{Prompt template for pairwise LMs-as-RMs for non-verifiable tasks. Note that unlike verifiable tasks, here we ask the model to assign a score between 0 to 10 to capture the fine-grained quality.}
    \label{fig:pairwise}
    }
\end{figure*}

\begin{figure*}[t]
    \centering
    \begin{prompt}{Prompt Template for Pointwise Reference-based LMs-as-RMs}

\texttt{You are given a user question, \textcolor{darkgreen}{a reference answer,} and a response from an AI assistant. Your task is to act as an impartial judge and evaluate how well the response fulfills the user's instructions. You will be shown multiple responses to the same prompt, but only one at a time. Evaluate each response independently.}\\

\texttt{Think carefully about how to assess the quality of the response and assign the assistant's response a score 1 if the response is correct, and 0 if not. Enclose the score within <score> and </score> tags.}\\

\texttt{Format your output like this:\\
<think> your\_thinking\_process </think>  \\
<score> 0 or 1 </score>}\\

\texttt{Below are the user’s question and the assistant’s response:\\
\\
~[User Question]\\
~~{\color{denim}{\{instruction\}}}\\
\\
~\textcolor{darkgreen}{[Reference Answer]}\\
~~{\color{denim}{\{reference\_answer\}}}\\
\\
~[The Start of the Assistant's Answer]\\
~~{\color{denim}{\{response\}}}\\
~[The End of the Assistant's Answer]}
\end{prompt}
\vspace{-0.5em}
    \caption{{Prompt template for Pointwise Reference-based LMs-as-RMs, where an additional reference answer is provided compared to the reference-free version. Note that the reference answer is only provided as additional context and the LM-as-RM is still prompted to evaluate the entire response.}
    \label{fig:ref}
    }
\end{figure*}

\begin{table}[]
\centering
\caption{Comparison of different post-trained Qwen3-8B (Instruct) models using RLLM, RLVR, or RLHF, showing generalization of our RLLM recipe to a stronger 8B policy model. All models are evaluated in thinking mode.} 
\label{tab:appendix_qwen8}
\setlength{\tabcolsep}{5pt}
\resizebox{\textwidth}{!}{
\begin{tabular}{lllcccccccc}
\toprule
\multirow{2}{*}{\sc Method} & \multirow{2}{*}{RM/Verifier} & \multirow{2}{*}{RM Type} & \multirow{2}{*}{RM Size} & \multicolumn{6}{c}{\sc Math (Easy-to-Verify)} & {\sc Hard-to-Verify}                          \\ \cmidrule(lr){5-10}
& & & & \textbf{AIME24} & \textbf{AIME25} & \textbf{BRUMO25} & \textbf{HMMT24} & \textbf{HMMT25} & \textbf{Average} & \textbf{Physics} \\ \midrule
\rowcolor{gray!15} \multicolumn{11}{l}{\textit{\textbf{Policy Trained on hard-to-verify samples w/o Reference Answer}}} \\
\texttt{Qwen3-8B} & -- & -- & -- &	76.00	&67.30& 68.75 &	35.41&	41.25&	57.74 & 51.70 \\
RLHF & \texttt{Nexusflow/Athene-RM-8B} & Scalar & 8B	& 73.34	&60.42	&65.00 &37.91&	38.33	&	55.00 & 50.35\\
RLVR & \texttt{nvidia/AceMath-7B-RM} & Scalar & 7B & 72.91	&65.00	& 65.41 &33.75	&42.49	&55.91 & 50.76 \\
\rowcolor{mint} \textbf{RLLM} & \texttt{Qwen3-8B} & Generative & 8B & 75.84 & 70.00 & 69.58 & 50.00 & 43.33 & 61.75 & 54.32 \\ 
\rowcolor{mint} \textbf{RLLM} & \texttt{J1-Qwen3-32B} & Generative & 32B & 73.32 & 70.43 & 73.74 & 43.30 & 50.82 & 62.32 & 54.17 \\
                        \bottomrule
\end{tabular}}
\end{table}

\input{rllm_tables/non-thinking}

\begin{table}[]
\centering
\caption{Comparison of different post-trained OctoThinker-8B-Hybrid-Base models using RLLM, RLVR, or RLHF. This shows the generalization of RLLM to Llama-based models.} 
\label{tab:appendix_octothinker}
\setlength{\tabcolsep}{3pt}
\resizebox{\textwidth}{!}{
\begin{tabular}{llllccccccc}
\toprule
\sc Method & RM/Verifier & RM Type & RM Size & \textbf{AIME24} & \textbf{AIME25} & \textbf{HMMT24} & \textbf{HMMT25} & \textbf{BRUMO25} & \textbf{MATH500} & \textbf{Average} \\ \midrule
\texttt{OctoThinker-8B-Hybrid-Base} & -- & -- & -- & 2.49 & 2.70 & 4.30 & 0.00 & 0.00 & 52.36 & 10.32  \\ \midrule
\rowcolor{gray!15} \multicolumn{11}{l}{\textit{\textbf{Policy Trained on hard-to-verify samples w/o Reference Answer}}} \\
RLHF &  \texttt{Skywork-Reward-V2-Llama-3.1-8B} & Scalar & 8B & 7.30 & 3.74 & 6.88 & 3.30 & 0.00 & 60.04 & 13.54 \\
\rowcolor{mint}  \textbf{RLLM} & \texttt{Qwen3-1.7B (prompted)} & Generative & 1.7B & \textbf{7.93} & \textbf{4.37}	& \textbf{9.79} & \textbf{6.70} &	\textbf{0.83} &	\textbf{60.86} & \textbf{15.08} \\ 
\rowcolor{mint} \textbf{$\Delta$ w/ OctoThinker-8B-Hybrid-Base} & -- & -- & -- & +5.44 & +1.67 & +5.49 & +6.70 & +0.83 & +8.50 & +4.76  \\
\midrule
\rowcolor{gray!15} \multicolumn{11}{l}{\textit{\textbf{Policy Trained on hard-to-verify samples w/ Reference Answer}}} \\
RLVR & Math-Verify & Rule-based & -- & 3.31 & 1.45 & 6.67 & 3.30 & \textbf{0.41}	& 55.64 & 11.80 \\
RLVR & \texttt{TIGER-Lab/general-verifier} & Generative & 1.5B & 2.29 & 1.24 & 5.00 & 3.30 & 0.00 & 47.67 & 9.92 \\
\rowcolor{mint}  \textbf{RLLM} & \texttt{Qwen3-1.7B (prompted)} & Generative & 1.7B & \textbf{10.84} & \textbf{7.08} & \textbf{7.52} & \textbf{3.30} & 0.00 & \textbf{61.75} & \textbf{15.08}\\
\rowcolor{mint} \textbf{$\Delta$ w/ OctoThinker-8B-Hybrid-Base} & -- & -- & -- & +8.35 & +4.38 & +3.22 & +3.30 & +0.00 & +9.39 & +4.76 \\
                        \bottomrule
\end{tabular}}
\end{table}

\begin{table}[]
\centering
\caption{\label{tab:appendix_non_verifiable_8b}Comparison of Win Rate (WR) and Length Controlled Win Rate (LCWR) of RLLM and RLHF on non-verifiable instruction-following tasks when training a Qwen3-8B policy. For AlpacaEval 2.0, we use GPT-4o as the evaluator and for ArenaHard 2.0, we use GPT-4.1 as the evaluator. RLLM matches or outperforms RLHF, obtaining best win rates on hard prompts of ArenaHard 2.0.} 
\resizebox{\textwidth}{!}{
\begin{tabular}{llllccc}
\toprule
\multirow{3}{*}{\sc Method} & \multirow{3}{*}{RM} & \multirow{3}{*}{RM Type} & \multirow{3}{*}{RM Size} & \multicolumn{3}{c}{\sc Datasets}                          \\ \cmidrule(lr){5-7}
& & & & {\textbf{AlpacaEval 2.0}} & \textbf{ArenaHard 2.0}  & \textbf{ArenaHard 2.0}  \\ 
& & & & (LCWR / WR) & (Hard Prompts) & (Creative Writing) \\
\midrule
\rowcolor{gray!15} \multicolumn{7}{l}{\textbf{Policy Trained on non-verifiable WildChat samples in thinking mode}} \\
\texttt{Qwen3-8B} & -- & -- & --& 63.1 / 65.1 & 22.4 (-1.5 / +1.7) &	33.3 (-2.5 / +2.7) \\
RLHF & \texttt{Nexusflow/Athene-RM-8B} & Scalar & 8B & 70.9 / 71.2 & 25.4  (-1.6 / +1.6) & 57.9  (-2.8 / +2.0)  \\
RLHF & \texttt{Skywork-Reward-V2-Llama-3.1-8B} & Scalar & 8B & 68.6 / 72.3 &  26.4  (-1.6 / +1.7) & 50.9  (-2.9 / +2.9) \\					
\rowcolor{mint} \textbf{RLLM} & \texttt{Qwen3-8B (prompted)} & Generative & 8B & \textbf{71.4 / 77.1} & \textbf{32.7  (-1.6 / +1.5)} & \textbf{61.9  (-2.3 / +2.3)} \\ 
                        \bottomrule
\end{tabular}}
\end{table}

%% file: rllm_tables/non-thinking.tex
\begin{table}[]
\centering
\caption{Comparison of different post-trained Qwen3-1.7B (Instruct) models using RLLM, RLVR, or RLHF. All models are evaluated in \emph{non-thinking} mode. Similar to our conclusions with results in thinking mode, RLLM outperforms RLHF and RLVR.} 
\label{tab:appendix_non_thinking}
\setlength{\tabcolsep}{5pt}
\resizebox{\textwidth}{!}{
\begin{tabular}{llllccccccc}
\toprule
\multirow{2}{*}{\sc Method} & \multirow{2}{*}{RM/Verifier} & \multirow{2}{*}{RM Type} & \multirow{2}{*}{RM Size} & \multicolumn{6}{c}{\sc Math (easy-to-verify)} & {\sc Hard-to-Verify}                          \\ \cmidrule(lr){5-10}
& & & & \textbf{AIME24} & \textbf{AIME25} & \textbf{BRUMO25} & \textbf{HMMT24} & \textbf{HMMT25} & \textbf{Average} & \textbf{Physics} \\ \midrule
\texttt{Qwen3-1.7B}& -- & -- & -- & 13.33 & 10.83 & 16.04 & 6.25 & 5.21 & 10.33 & 8.75 \\ \midrule
\rowcolor{gray!15} \multicolumn{11}{l}{\textit{\textbf{\textbf{Policy Trained on hard-to-verify samples w/o Reference Answer}}}} \\
RLHF & \texttt{Nexusflow/Athene-RM-8B} 
& Scalar & 8B & \phantom{0}5.84 & \phantom{0}6.25 &12.08 & \phantom{0}2.28	& \phantom{0}1.65& \phantom{0}5.62 & 7.95\\

RLHF & \texttt{Skywork-Reward-V2-Llama-3.1-8B} & Scalar & 8B  & 
21.87	&20.21&	21.87	&12.92&	10.00&	17.37 & 10.00

\\
RLHF & \texttt{nvidia/AceMath-7B-RM} 
& Scalar & 7B & 20.21&	16.87	&18.96	&\phantom{0}7.72	&\phantom{0}5.84&	13.92 & 10.45  \\
\rowcolor{mint} \textbf{RLLM} & \texttt{J1-Qwen3-32B-RM} & Generative & 32B & \textbf{39.58} & \textbf{35.21} & \textbf{43.97} & \textbf{20.21} & \textbf{17.51} & \textbf{31.30} & \textbf{16.02} \\ 
\rowcolor{mint} \textbf{$\Delta$ w/ Qwen3-1.7B} & -- & -- & -- & +26.25 & +24.38 & +27.93 & +13.96 & +12.30 & +20.97 & +7.27 \\
\midrule
\rowcolor{gray!15} \multicolumn{11}{l}{\textbf{Policy Trained on hard-to-verify samples w/ Reference Answer}} \\
RLVR & Math-Verify 
& Rule-based	& -- & 
12.51&	12.71&	20.00	&\phantom{0}8.12	&\phantom{0}7.93&	12.25 & 8.75

\\
RLVR & \texttt{TIGER-Lab/general-verifier}
& Generative	& 1.5B &17.71	&15.41&	18.97	&\phantom{0}7.92	&\phantom{0}7.92&	13.58 & 11.93	\\				
\rowcolor{mint} \textbf{RLLM} & \texttt{J1-Qwen3-32B-Verifier} & Generative & 32B & \textbf{35.41} & \textbf{30.01} & \textbf{41.04} & \textbf{20.00} & \textbf{15.00} & \textbf{28.29} & \textbf{14.36} \\
                        \bottomrule
    \end{tabular}}
\end{table}

%% file: rllm_tables/rm_examples.tex
\begin{figure*}
\begin{prompt}{Pointwise LM-as-RM Training Example (Correct)}
\textbf{Question:}

A conducting cylinder of radius \(a\) and finite length \(L\) (\(0\le z\le L\)) is filled with the steady helical flow
\[
\mathbf{U}(r,\theta,z)=U_{0}\,\hat{\mathbf{z}}+\Omega\,r\,\hat{\boldsymbol{\theta}},\qquad 0\le r\le a .
\]

The magnetic field obeys the induction equation with magnetic diffusivity \(\eta\). Assume that the end caps at \(z=0\) and \(z=L\) are perfectly conducting, so that the axial dependence of any normal-mode perturbation must satisfy the boundary condition \(\mathbf{B}=0\) at the end caps. Consequently the axial wavenumber is quantised as
\[
k_{n}= \frac{n\pi}{L}, \qquad n=1,2,\dots .
\]

Adopt the normal-mode ansatz
\[
\mathbf{B}(r,\theta,z,t)=\mathbf{b}(r)\,e^{\mathrm{i}m\theta+\mathrm{i}k_{n}z+\sigma t},
\]
and focus on the most unstable azimuthal mode \(m=1\). For this mode the growth rate can be written in the form
\[
\sigma = \frac{U_{0}}{a}\Bigl(\mathrm{Rm}\,F\bigl(q,k_{n}a\bigr)-\lambda_{1}^{2}\Bigr),
\]
where \(\displaystyle \mathrm{Rm}= \frac{U_{0}a}{\eta}\) is the magnetic Reynolds number, \(q=\Omega a/U_{0}\) is the dimensionless pitch, \(\lambda_{1}\approx 3.8317\) is the first zero of the Bessel function \(J_{1}\), and \(F(q,k_{n}a)\) is a positive dimensionless function that satisfies \(0<F\le 1\) for all admissible \(q\) and \(k_{n}\).

Derive the \textbf{necessary condition} for dynamo action (i.e. \(\sigma>0\)) that must hold \textbf{for every admissible axial mode} \(n\). Express your result as a single inequality involving only the magnetic Reynolds number \(\mathrm{Rm}\), the constant \(\lambda_{1}\), and the geometric ratio \(a/L\).

\vspace{3mm}
\textbf{Reference Answer:} \(\displaystyle \mathrm{Rm}\;>\;\lambda_{1}^{2}\)

\vspace{3mm}
\textbf{Model Response:}

To determine the necessary condition for dynamo action, we begin by analyzing the given expression for the growth rate:
\[
\sigma = \frac{U_0}{a} \left( \mathrm{Rm} \, F\left(q, k_n a\right) - \lambda_1^2 \right).
\]

For the magnetic field to grow (i.e., to exhibit dynamo action), the growth rate \(\sigma\) must be positive. Therefore:
\[
\sigma > 0 \Rightarrow \mathrm{Rm} \, F(q, k_n a) > \lambda_1^2.
\]

Thus, the necessary condition for dynamo action is: \(\mathrm{Rm} > \frac{\lambda_1^2}{F(q, k_n a)}\).

\ldots [analysis of the function \(F(q, k_n a)\) and worst-case scenarios] \ldots

To ensure that \(\mathrm{Rm} > \frac{\lambda_1^2}{F(q, k_n a)}\) holds for all \(n\), we must consider the worst-case value of \(F(q, k_n a)\). Since \(F(q, k_n a)\) is bounded above by 1, the maximum of \(\frac{\lambda_1^2}{F(q, k_n a)}\) over all \(n\) occurs when \(F(q, k_n a) = 1\).

Therefore, the necessary condition for dynamo action is:
\[
\boxed{\mathrm{Rm} > \lambda_1^2}
\]

\vspace{3mm}
\textbf{GPT-OSS-120B Judgment:} The response derives the same inequality as the reference answer and correctly concludes that the most restrictive necessary condition is obtained when \(F=1\). Both answers are equivalent. \textcolor{blue}{\textbf{CORRECT}}
\end{prompt}

\caption{\label{fig:appendix_rm_correct} Example of a synthetically-generated correct training example for LM-as-RM training. Given a (question, reference answer), we sample a response from the policy and check whether the answer matches the reference answer using GPT-OSS-120B.}
\end{figure*}

\begin{figure*}
\begin{prompt}{Pointwise LM-as-RM Training Example (Incorrect)}
\textbf{Question:}

A conducting cylinder of radius \(a\) and finite length \(L\) (\(0\le z\le L\)) is filled with the steady helical flow
\[
\mathbf{U}(r,\theta,z)=U_{0}\,\hat{\mathbf{z}}+\Omega\,r\,\hat{\boldsymbol{\theta}},\qquad 0\le r\le a .
\]

The magnetic field obeys the induction equation with magnetic diffusivity \(\eta\). Assume that the end caps at \(z=0\) and \(z=L\) are perfectly conducting, so that the axial dependence of any normal-mode perturbation must satisfy the boundary condition \(\mathbf{B}=0\) at the end caps. Consequently the axial wavenumber is quantised as
\[
k_{n}= \frac{n\pi}{L}, \qquad n=1,2,\dots .
\]

Adopt the normal-mode ansatz
\[
\mathbf{B}(r,\theta,z,t)=\mathbf{b}(r)\,e^{\mathrm{i}m\theta+\mathrm{i}k_{n}z+\sigma t},
\]
and focus on the most unstable azimuthal mode \(m=1\). For this mode the growth rate can be written in the form
\[
\sigma = \frac{U_{0}}{a}\Bigl(\mathrm{Rm}\,F\bigl(q,k_{n}a\bigr)-\lambda_{1}^{2}\Bigr),
\]
where \(\displaystyle \mathrm{Rm}= \frac{U_{0}a}{\eta}\) is the magnetic Reynolds number, \(q=\Omega a/U_{0}\) is the dimensionless pitch, \(\lambda_{1}\approx 3.8317\) is the first zero of the Bessel function \(J_{1}\), and \(F(q,k_{n}a)\) is a positive dimensionless function that satisfies \(0<F\le 1\) for all admissible \(q\) and \(k_{n}\).

Derive the \textbf{necessary condition} for dynamo action (i.e. \(\sigma>0\)) that must hold \textbf{for every admissible axial mode} \(n\). Express your result as a single inequality involving only the magnetic Reynolds number \(\mathrm{Rm}\), the constant \(\lambda_{1}\), and the geometric ratio \(a/L\).

\vspace{3mm}
\textbf{Reference Answer:} \(\displaystyle \mathrm{Rm}\;>\;\lambda_{1}^{2}\)

\vspace{3mm}
\textbf{Model Response:}

To determine the necessary condition for dynamo action, we analyze the growth rate. For dynamo action, we require \(\sigma > 0\):
\[
\sigma = \frac{U_0}{a} \left( \mathrm{Rm} \, F(q, k_n a) - \lambda_1^2 \right) > 0
\]

Rewriting:
\[
\mathrm{Rm} \, F(q, k_n a) > \lambda_1^2
\]

So, for every \(n\), we must have:
\[
\mathrm{Rm} > \frac{\lambda_1^2}{F(q, k_n a)}
\]

To ensure this is true for all \(n\), we must consider the worst-case scenario, i.e., the minimum value of \(F(q, k_n a)\), since it will make \(\frac{\lambda_1^2}{F(q, k_n a)}\) as large as possible.

\ldots [analysis of the function \(F(q, k_n a)\) and geometric considerations] \ldots

The function \(F(q, k_n a)\) is positive and less than or equal to 1 for all admissible \(q\) and \(k_n\). The problem explicitly states that we are to express the necessary condition in terms of \(\mathrm{Rm}\), \(\lambda_1\), and the geometric ratio \(a/L\).

The necessary condition for \(\sigma > 0\) for every \(n\) is:
\[
\boxed{\mathrm{Rm} > \lambda_1^2 \cdot \frac{a}{L}}
\]

This is the most restrictive condition, since it ensures that the growth rate is positive for the worst-case value of \(F(q, k_n a)\).

\vspace{3mm}
\textbf{Judgment:} The reference answer states \(\mathrm{Rm}>\lambda_{1}^{2}\) with no dependence on the geometric ratio \(a/L\). The response incorrectly introduces a factor of \(\frac{a}{L}\) in the final answer. This additional factor changes the condition qualitatively and is not present in the reference answer. The response's condition is therefore not equivalent. \textcolor{red}{\textbf{INCORRECT}}
\end{prompt}
\caption{\label{fig:appendix_rm_incorrect} Example of a synthetically-generated incorrect training example for LM-as-RM training. Given a (question, reference answer), we sample a response sampled from the policy and check whether the answer matches the reference answer using GPT-OSS-120B.}
\end{figure*}

%% file: rllm_tables/policy_examples.tex
\begin{figure*}
\begin{prompt}{Example of a hard-to-verify math problem}
\textbf{Question:} 
Find all real polynomials \( p(x) \) such that

\[ p(x^2 + 1) = p(x)^2 + 1 \]

for all \( x \in \mathbb{R} \).

\vspace{3mm}

\textbf{Reference Answer:} \( p_n(x) = ((\cdots((x^2 + 1)^2 + 1)^2 + \cdots)^2 + 1 \) (nested \( n \) times)
\end{prompt}
\caption{\label{fig:appendix_hard_to_verify} Example of a hard-to-verify math problem for policy training where the reference answer is a mathematical expression.}
\end{figure*}

\vspace{1cm}

\begin{figure*}
\begin{prompt}{Example of an easy-to-verify math problem}
\textbf{Question:} 
In \( \triangle ABC \) with \( BC=a \), \( CA=b \), and \( AB=c \), given that \( (b+c)(b-c)=ca \) and \( \angle BAC=75^{\circ} \), find \( \angle ABC \).

\vspace{3mm}

\textbf{Reference Answer:} \( 70^\circ \)
\end{prompt}
\caption{\label{fig:appendix_easy_to_verify} Example of an easy-to-verify math problem for policy training where the reference answer is an integer.}
\end{figure*}

%% file: rllm_tables/j1_eval_examples.tex
\begin{figure*}
\begin{prompt}{J1-Qwen3-1.7B vs J1-Qwen3-32B Evaluation Traces}

\textbf{\large Question:}

Consider two open Heisenberg antiferromagnetic spin‑\(1/2\) chains, one of length \(N_{1}=100\) and one of length \(N_{2}=200\).  
Using the density matrix renormalization group (DMRG) with three different numbers of kept states \(m\), the following total ground‑state energies (in units of the exchange constant \(J\)) and discarded weights \(\varepsilon_{m}\) are obtained:

| Chain length | \(m\) | Energy \(E_{m}\) | Discarded weight \(\varepsilon_{m}\) |
|--------------|------|------------------|--------------------------------------|
| \(N_{1}=100\) | 50   | \(-44.12\)       | \(3.2\times10^{-4}\)                 |
| \(N_{1}=100\) | 100  | \(-44.23\)       | \(8.5\times10^{-5}\)                 |
| \(N_{1}=100\) | 200  | \(-44.28\)       | \(1.9\times10^{-5}\)                 |
| \(N_{2}=200\) | 50   | \(-88.30\)       | \(3.2\times10^{-4}\)                 |
| \(N_{2}=200\) | 100  | \(-88.44\)       | \(8.5\times10^{-5}\)                 |
| \(N_{2}=200\) | 200  | \(-88.51\)       | \(1.9\times10^{-5}\)                 |

The DMRG algorithm is variational, i.e. each \(E_{m}\) is an upper bound to the exact ground‑state energy for the corresponding chain length: \(E_{0}(N) \le E_{m}(N)\).

In addition, the following rigorous bound on the energy error is known:
\[
|E_{m}(N)-E_{0}(N)| \le K\,\varepsilon_{m},
\qquad K = 1.5\times10^{4}.
\]

The exact ground‑state energy per site of the infinite open Heisenberg chain is
\[
e_{\infty}= -0.4431471806\,J,
\]
and for any open chain of length \(N\) the exact energy satisfies the finite‑size inequality
\[
N\,e_{\infty} \;\le\; E_{0}(N) \;\le\; N\,e_{\infty} + \frac{C}{N},
\qquad C = 12.0.
\]

Using **all** the information above (variational property, discarded‑weight error bound, infinite‑chain reference, finite‑size bound, and the data for both chain lengths), give a **rigorous interval** that is guaranteed to contain the exact total ground‑state energy \(E_{0}(N_{1}=100)\) of the 100‑site chain.  
Express your answer **only** in interval notation (e.g. \(\left[a,\,b\right]\)) without any additional text or equations.

\vspace{0.3cm}
\hrule
\vspace{0.3cm}

\textbf{\large Response (Truncated):}

**From DMRG with $ m = 200 $:**
  - $ E_0(100) \in [-44.28 - 0.285, -44.28 + 0.285] = [-44.565, -43.995] $
  - Intersect with finite-size bound:
  $$
  [-44.31471806, -44.19471806]
  $$

---

[...]

**Final Interval**
The **exact** ground-state energy $ E_0(100) $ is guaranteed to lie in the **intersection** of the DMRG error bounds and the finite-size bound. This is the **tightest and most rigorous** interval that contains $ E_0(100) $.

$$
\boxed{[-44.31471806, -44.19471806]}
$$

\vspace{0.3cm}
\hrule
\vspace{0.3cm}

\textbf{\large Reference Answer:}

[-44.31471806, -44.28]

\vspace{0.3cm}
\hrule
\vspace{0.3cm}

\textbf{\large Expected Evaluation: Incorrect}

<score> 0 </score>

\vspace{0.3cm}
\hrule
\vspace{0.3cm}

\textbf{\large J1-Qwen3-1.7B-RM's Incorrect Thinking Trace (Truncated):}

...the assistant's final interval is [-44.31471806, -44.19471806], which is the same as the finite-size bound. But the assistant also mentioned that the DMRG results with m=50, 100, 200 give energies -44.12, -44.23, -44.28...{\textcolor{Red}{So the intersection of those is [-44.3147, -44.1947]. So the assistant's final interval is [-44.3147, -44.1947], which is the intersection of the DMRG error bound and the finite-size bound...Therefore, the assistant's answer is correct. So the score should be 1.}}

\textbf{Evaluation Score:} <score> 1 </score>

\vspace{0.3cm}
\hrule
\vspace{0.3cm}

\textbf{\large J1-Qwen3-32B-RM's Correct Thinking Trace (Truncated):}

...the DMRG results provide upper bounds. For example, the DMRG with m=200 gives $E_m = -44.28$, which is an upper bound...Since $E_0 \le E_{200} = -44.28$, the tightest upper bound is $-44.28$, not $-44.1947$...the exact $E_0(100)$ must lie in [-44.3147, -44.28]. {\textcolor{ForestGreen}{Because the upper bound is the minimum of the DMRG upper bounds (which is -44.28) and the finite-size upper bound (which is -44.1947)...the assistant's interval is too wide. The correct interval should be the intersection of the finite-size lower bound and the DMRG upper bounds...Therefore, the rigorous interval should be [-44.31471806, -44.28]. The assistant's answer is wrong.}}

\textbf{Evaluation Score:} <score> 0 </score>

\end{prompt}
\caption{\label{fig:appendix_j1_traces} Qualitative comparison of a weaker LM-as-RM J1-Qwen3-1.7B's thinking trace with a stronger J1-Qwen3-32B's thinking trace. The stronger LM-as-RM correctly identifies the response as incorrect and also specifies the correct answer (which matches the reference answer).}
\end{figure*}